\documentclass[nohyperref]{article}

\usepackage{microtype}
\usepackage{graphicx}
\usepackage{subfigure}
\usepackage{booktabs} 
\usepackage{bm}
\usepackage{booktabs, siunitx}
\usepackage[svgnames,table]{xcolor}
\usepackage[tableposition=above]{caption}
\usepackage{booktabs}
\usepackage{enumitem}
\usepackage{hyperref}
\usepackage{multirow}
\usepackage{amsmath}
\usepackage{amssymb}
\usepackage{mathtools}
\usepackage{amsthm}
\usepackage[textsize=tiny]{todonotes}
\usepackage{soul}
\usepackage[capitalize,noabbrev]{cleveref}

\usepackage[accepted]{icml2023}

\theoremstyle{plain}

\theoremstyle{definition}

\theoremstyle{remark}

\icmltitlerunning{Multi-Objective Population Based Training}

\definecolor{bluegray}{rgb}{0.4, 0.6, 0.8}
\definecolor{taupegray}{rgb}{0.55, 0.52, 0.54}
\definecolor{steelblue}{rgb}{0.27, 0.51, 0.71}

\renewcommand{\algorithmiccomment}[1]{\hfill \color{steelblue} $\triangleright$ #1 \color{black}}
\newcommand{\MYCOMMENT}[1]{\hfill \color{taupegray} $\triangleright$ #1 \color{black}}

\DeclareMathOperator*{\argmax}{arg\,max}
\DeclareMathOperator*{\argmin}{arg\,min}

\begin{document}

\setlength{\abovedisplayskip}{6pt}
\setlength{\belowdisplayskip}{6pt}
\setlength{\lineskip}{0pt}

\twocolumn[
\icmltitle{Multi-Objective Population Based Training}

\begin{icmlauthorlist}
\icmlauthor{Arkadiy Dushatskiy}{cwi}
\icmlauthor{Alexander Chebykin}{cwi}
\icmlauthor{Tanja Alderliesten}{lumc}
\icmlauthor{Peter A.N. Bosman}{cwi,delft}

\end{icmlauthorlist}

\icmlaffiliation{cwi}{Centrum Wiskunde \& Informatica, Amsterdam, the Netherlands}
\icmlaffiliation{lumc}{Leiden University Medical Center, Leiden, the Netherlands}
\icmlaffiliation{delft}{Delft University of Technology, Delft, the Netherlands}
\icmlcorrespondingauthor{Arkadiy Dushatskiy}{arkadiy.dushatskiy@cwi.nl}

\icmlsetsymbol{equal}{*}

\icmlkeywords{Machine Learning, ICML}

\vskip 0.3in
]
\printAffiliationsAndNotice{}

\begin{abstract}
Population Based Training (PBT) is an efficient hyperparameter optimization algorithm. PBT is a single-objective algorithm, but many real-world hyperparameter optimization problems involve two or more conflicting objectives. In this work, we therefore introduce a multi-objective version of PBT, \emph{MO-PBT}. 
Our experiments on diverse multi-objective hyperparameter optimization problems (Precision/Recall, Accuracy/Fairness, Accuracy/Adversarial Robustness) show that MO-PBT outperforms random search, single-objective PBT, and the state-of-the-art multi-objective hyperparameter optimization algorithm MO-ASHA.
\end{abstract}

\section{Introduction}

The computational complexity of machine learning tasks has drastically increased in recent years. This has been caused by larger models (especially, deep neural networks \cite{dosovitskiy2020image, kaplan2020scaling}) and larger available datasets \cite{thomee2016yfcc100m, byeon2022coyo}. At the same time, the problem of tuning model hyperparameters remains crucial for achieving maximal performance \cite{kadra2021well, zhang2021importance, liu2022convnet}. Thus, there is a growing demand for efficient algorithms to do hyperparameter tuning. Moreover, in real-world problems, there might be more than one objective that a user is interested in. An example of such a scenario which recently received a lot of attention from the machine learning community is finding a trade-off between the 
 predictive accuracy of a classifier and its fairness \cite{schmucker2020multi, chuang2021fair}. When different objectives are conflicting and the target trade-off is not known a priori, usually no single best model (or hyperparameter setting) exists. Thus, many models with different trade-offs between the objectives should be presented to the user. Finding hyperparameters that result in models with the best trade-offs is a multi-objective optimization problem.

One of the most efficient approaches to single-objective Hyperparameter Optimization (HPO) is Population Based Training (PBT) \cite{jaderberg2017population}. PBT has two features which ensure its efficiency. Firstly, it is a highly parallelizable, asynchronous algorithm, which means that the available hardware can be effectively utilized. Secondly, in contrast to standard optimization techniques which usually train models from scratch in order to estimate the performance of a particular hyperparameter setting, PBT optimizes hyperparameters during model training. In this work, we propose to expand Population Based Training \cite{jaderberg2017population} to Multi-Objective HPO (MO-HPO). The population of models used in PBT should be especially well suited for solving Multi-Objective (MO) problems, as maintaining a population is naturally helpful for finding a good trade-off front of solutions, which is known from the Evolutionary Algorithms (EAs) literature~\cite{deb2001multi, morales2022survey}. EAs such as NSGA-II \cite{deb2002fast} have been used for efficiently solving MO optimization problems, including Neural Architecture Search \cite{lu2019nsga}.

\begin{figure*}[h!]
    \centering
    \includegraphics[width=\textwidth]{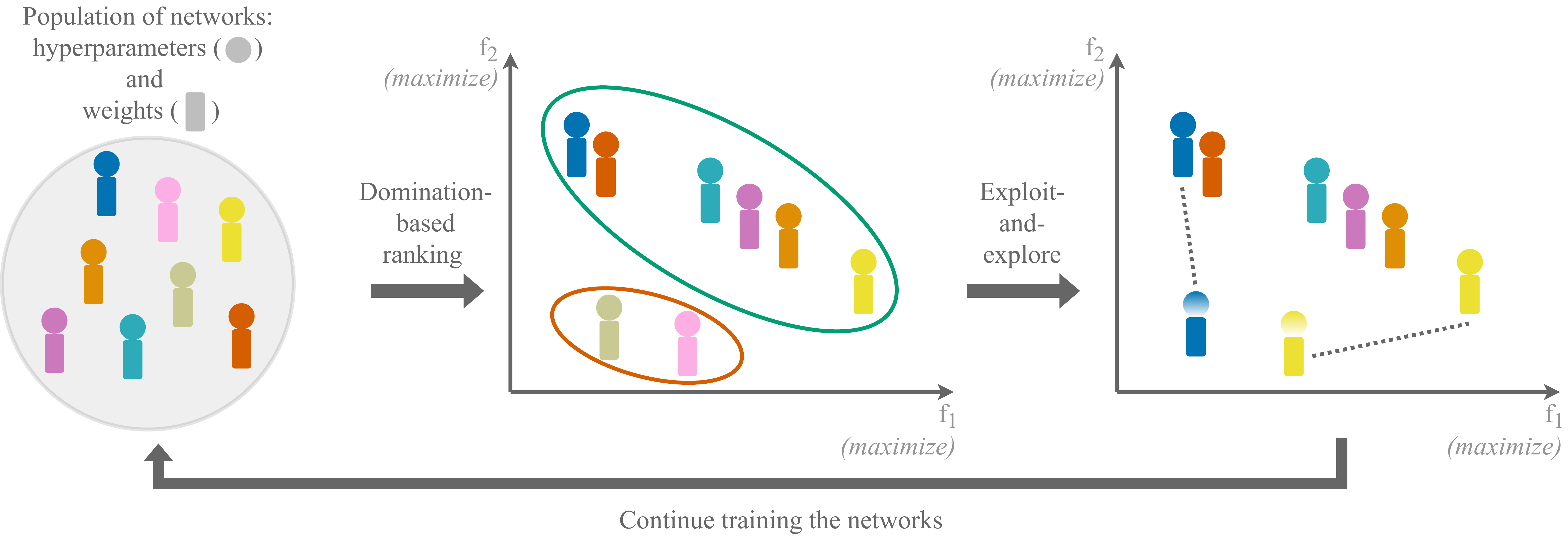}
    \caption{The scheme of the proposed MO-PBT applied to a bi-objective maximization task. After the networks (weights and  hyperparameters) in a population have been trained for several epochs, they are ranked using a \emph{domination-based} procedure. Here, each solution inside the smaller, dark-orange oval is dominated by at least one solution inside the larger, green oval (and therefore is considered to be worse). The inferior solutions are replaced with copies of the superior ones (copying is depicted with dotted lines), with  hyperparameters perturbed (depicted by adjusted colors). Then the networks are trained for several more epochs, and the loop continues.}
    \label{fig:pbt}
\end{figure*}

The main contributions of our work are the following:
\begin{enumerate}
    \item We expand Population Based Training to MO-HPO scenarios. The overview of the proposed Multi-Objective PBT (\emph{MO-PBT}) algorithm is demonstrated in Figure~\ref{fig:pbt}.
    
    \item We demonstrate that using single-objective PBT for MO-HPO by transforming it into a single-objective problem (via a scalarization technique or simply optimizing one of the objectives) is an inferior approach to MO-PBT which uses an MO technique of domination-based selection \cite{deb2002fast}.
    
    \item On a set of diverse MO-HPO problems, we demonstrate that MO-PBT outperforms the state-of-the-art efficient, parallelizable hyperparameter optimization algorithm, Multi-Objective Asynchronous Successive Halving (MO-ASHA) \cite{schmucker2021multi}. 

\end{enumerate}

\newpage
Experiments are performed on three different types of MO-HPO problems: precision/recall of a model, the predictive performance of a classifier/its fairness, and accuracy/adversarial robustness.

\section{Related work}
\subsection {Multi-objective hyperparameter optimization applications}
There are different scenarios in which a user might be interested in having the option to choose among models with different trade-offs between two (or even more) objectives. The classical example in machine learning is choosing a trade-off between precision and recall of a classifier. The importance of each metric might change depending on the application requirements. Another example is choosing a trade-off between the predictive quality of a classifier and its fairness. It was shown in \cite{chuang2021fair} that these objectives are conflicting. In \cite{zhang2019theoretically}, it was demonstrated that the classifier accuracy and its robustness to adversarial attacks are conflicting, and therefore, finding a trade-off between them is another interesting MO-HPO application.

\subsection {Efficient multi-objective optimization algorithms}

A popular class of algorithms for MO-HPO is the Bayesian Optimization (BO)  algorithms. Some of them work by reducing an MO optimization problem into a single-objective one by using scalarization techniques \cite{knowles2006parego, zhang2020random, paria2020flexible, zhang2009expensive}. An alternative approach to MO optimization with BO is based on expected hypervolume improvement calculation \cite{emmerich2011hypervolume}. 
While BO algorithms are sequential by  nature, recently they were extended to the batch-wise calculation of an objective function for solving MO problems \cite{daulton2020differentiable, daulton2021parallel}. However, the considered batch sizes were moderate (up to 32 solutions), so parallelization capabilities remain limited.

One of the drawbacks of typical BO algorithms is the equal allocation of resources to all evaluated solutions. In contrast to this, it was proposed to greedily stop underperforming model evaluations to save computational resources in Successive Halving \cite{jamieson2016non} and its extension Hyperband (that proposes a more complex resource allocation scheme) \cite{li2017hyperband}. 
Then, an asynchronous version of Hyperband called Asynchronous Successive Halving Algorithm (ASHA) was proposed, which was shown to achieve a substantial wall-clock time speed-up \cite{li2020system}. Hyperband was extended to MO problems by using random scalarizations in \cite{schmucker2020multi, guerrero2021bag}. 
Finally, ASHA was extended to MO problems in \cite{schmucker2021multi}. Different approaches to adapting ASHA to MO optimization were compared in \cite{schmucker2021multi} and it was concluded that the techniques utilizing the geometry of the Pareto front, in other words, domination-based selection such as in NSGA-II \cite{deb2002fast}, outperform  scalarization-based techniques.

It was proposed to integrate BO algorithms into Hyperband: \cite{falkner2018bohb} replaces random sampling of new candidate solutions by using a Bayesian sampler (TPE \cite{bergstra2011algorithms}) for more efficient search space exploration. MO-BOHB extends this idea to a multi-objective TPE sampler (MOTPE) \cite{ozaki2020multiobjective}. 

\subsection{Population Based Training}
A general formulation of PBT was proposed in \cite{jaderberg2017population}. It was shown to be an efficient way to jointly optimize hyperparameters and model weights of agents in reinforcement learning tasks, Generative Adversarial Networks, and Transformer networks applied to the machine translation task. Later it was shown that it can be also used to efficiently optimize data augmentation parameters for standard image classification datasets such as CIFAR-10/100 \cite{ho2019population} and 3D object detection \cite{cheng2020improving}. In \cite{liang2021regularized} it was proposed to incorporate an exploration component in the evaluation procedure of the solutions and add a crossover operator to recombine hyperparameter vectors. In \cite{dalibard2021faster} a more complex training scheme with multiple populations was proposed in order to improve the original PBT on problems where the greedy nature of the algorithm might lead to suboptimal results. Other modifications of PBT aim at improving its efficiency by integrating BO techniques \cite{parker2020provably, pmlr-v188-wan22a}. However, we would like to emphasize that all existing PBT modifications are single-objective and are not well-suited to solve multi-objective problems.

In this work, we follow the original design of PBT, which is simpler than the later proposed alternatives and was shown to work well on a diverse set of problems \cite{jaderberg2017population, ho2019population, cheng2020improving}. However, our approach is general and can potentially be used with any PBT modification.

\section{Preliminaries}

\subsection{Multi-objective optimization} \label{subsec:mo_optimization}
MO optimization problems are characterized by the presence of multiple conflicting objectives. Thus, solving the optimization problem entails finding the best possible trade-offs between the objectives. An MO optimization problem (without loss of generality, we consider maximization) with $K$ objectives can be formulated as follows:
 $$\max_{x\in X}{f}(x) = \max_{x\in X} {(f_1(x), f_2(x), \dots ,f_K(x))},$$ where $X \subseteq \mathcal{S}$ is a search space of solutions considered feasible ($\mathcal{S}$ is a search space of all solutions). 
 It is said that a solution $x'$ dominates a solution $x$ ($x' \succ x$) if $\forall i~ f_i(x') \ge f_i(x)$ and $\exists i~s.t.~f_i(x') > f_i(x)$.

The Pareto set $\mathcal{P}_s$ of $f$ is a set of all non-dominated solutions, i.e. $\mathcal{P}_s = \{x \in X | \nexists~x': x' \succ x\}$ while the Pareto front $\mathcal{P}_f$ is a set of objective values of solutions in $\mathcal{P}_s$: $\mathcal{P}_f = \{(f_1(x), f_2(x), \dots ,f_K(x)) | x \in \mathcal{P}_s \}$. While the Pareto front is often not known, the considered tangible goal of MO optimization algorithms is to obtain a good approximation of it. A popular measure of approximation quality is the dominated hypervolume \cite{zitzler1999evolutionary}. The hypervolume of a finite set of solutions $S$ is calculated as follows: $HV_r(S)=\lambda_K({z \in \mathbb{R}^K: \exists y \in S, r \prec z \prec f(y)})$, where $r \in \mathbb{R}^K$ is a chosen reference point and $\lambda_K$ is a Lebesgue measure. Intuitively, the hypervolume represents the volume (the area in the bi-objective case) between the reference point and the non-dominated trade-off front of solutions (see Appendix~\ref{sec:appendix_metrics},~Figure~\ref{fig:metrics_calc} for visualization).

 \subsection{Scalarization techniques} \label{subsec:scalarization}
 Scalarization is a commonly used technique for MO optimization, which transforms a multi-objective problem into a single-objective one: $\max_{x \in X} V(f(x), w)$, where $V$ is a scalarization function and $w$ is a scalarizing weight vector. 
Following MO optimization literature \cite{karl2022multi}, we use ParEGO scalarization function  \cite{knowles2006parego} (also called augmented Chebyshev scalarization \cite{steuer1983interactive}): $V_{ParEGO}=\rho V_{WS} + V_{Chebyshev}$, where  $V_{WS}$ is Weighted Sum scalarization: $V_{WS}(f(x), w))=\sum_i{w_i f_i(x)}$ and $V_{Chebyshev}$ is the Chebyshev scalarization: $V_{Chebyshev}(f(x), w))=\min_i({w_i f_i(x))}$, $\rho$ is set to $0.05$ in the original ParEGO implementation. Following MO-ASHA \cite{schmucker2021multi}, we also use Golovin scalarization \cite{zhang2020random}: $V_{Golovin}(f(x), w))=\min_i(\max(0, f_i(x)/w_i))^K$ ($K$ is the number of objectives).

 \section{Multi-Objective Population Based Training} \label{sec:MOPBT}

We start with a short summary of PBT and then describe our extension of it to the MO setting.

The goal of PBT is to optimize an objective function $f$. PBT has a population of $N$ solutions $\mathcal{P}=\{p_i\}_{i=1}^N$, where each individual $p_i$ comprises a tuple of model weights and hyperparameters:  $(\theta_i, \mathcal{H}_i)$. The main working principle of PBT is to optimize weights and hyperparameters in an interleaved fashion, which is achieved via two key operators: \emph{exploit} and \emph{explore}. The \emph{exploit} operator replaces a bad solution with a copy of a good one (both weights and hyperparameters are copied). The solution quality is determined by a ranking procedure. The \emph{explore} operator creates a new solution by, e.g., perturbing the hyperparameters of the existing one. 
Between exploit-and-explore steps, the weights of the models are trained as usual, e.g., using gradient descent.

How solutions are ranked needs to be changed when going from single- to multi-objective optimization. In the single-objective scenario, the population members can be ranked according to the optimization objective value, but with multiple objectives, ranking becomes less trivial. 

The first approach we consider is using a scalarization technique, i.e., mapping an objective vector into a scalar. It is then used for ranking solutions, just as in the single-objective case. Secondly, we consider domination-based ranking, as used for example in NSGA-II \cite{deb2002fast}. The main component of such an approach is the \emph{non-dominated sort} of solutions. The idea of the non-dominated sort is to partition a population of solutions $\mathcal{P}$ into non-dominated fronts of solutions, i.e., $\mathcal{P}=F^1 \cup F^2,\dots, \cup F^R; F^i \cap F^j = \emptyset ~\forall i,j$ such that: 
\begin{enumerate}[itemsep=-1pt, topsep=-1pt]
    \item All solutions in each front are non-dominated by each other: $\forall k: \forall v_1,v_2 \in F^k~ v_1 \nsucc v_2\text{~and~} v_2 \nsucc v_1$
    \item In the  $\mathit{k}^{th}$ front ($k > 1$), all solutions are dominated by a solution from a front with a smaller index: $\forall k, 2\le k\le R: \forall v_1 \in F^k~\exists v_2 \in F^m, m < k: v_2 \succ v_1$
\end{enumerate}
In the sorting procedure, all solutions from $F^1$ are ranked higher than 
 the solutions from $F^2$, the ones from $F^2$ are ranked higher than the solutions from $F^3$, etc. Within each front, the solutions are ranked according to an additional ranking criterion. In the original NSGA-II algorithm, the \emph{crowding distance} criterion was used. However, in \cite{schmucker2021multi} it was shown that the greedy scattered subset selection \cite{bosman2003balance} (called $\epsilon-$network in \cite{schmucker2021multi, salinas2021multi}) ranking performs better when integrated into the MO-ASHA algorithm (compared to MO-ASHA with the crowding distance). We also experimentally found that MO-PBT with the greedy scattered subset selection performs slightly better than MO-PBT with the crowding distance, as shown in Appendix~\ref{sec:appendix_ranking}. 
 
 The main idea behind the greedy scattered subset selection is to rank higher the solutions that are further away from the others. Specifically, the next solution is iteratively chosen in a greedy way such that it has the largest Euclidean distance (in the \emph{objective space}) to the closest already ranked solution. The visualization of this ranking procedure is shown in Figure~\ref{fig:NDS}, its pseudocode is listed in Appendix~\ref{sec:appendix_codes},~Algorithm~\ref{alg:epsnetwork}. 

 The \emph{exploit} and \emph{explore} operators of MO-PBT are described in Section~\ref{pbt_operators}.

\begin{figure}[ht]
     \centering
     \includegraphics[width=0.35\textwidth]{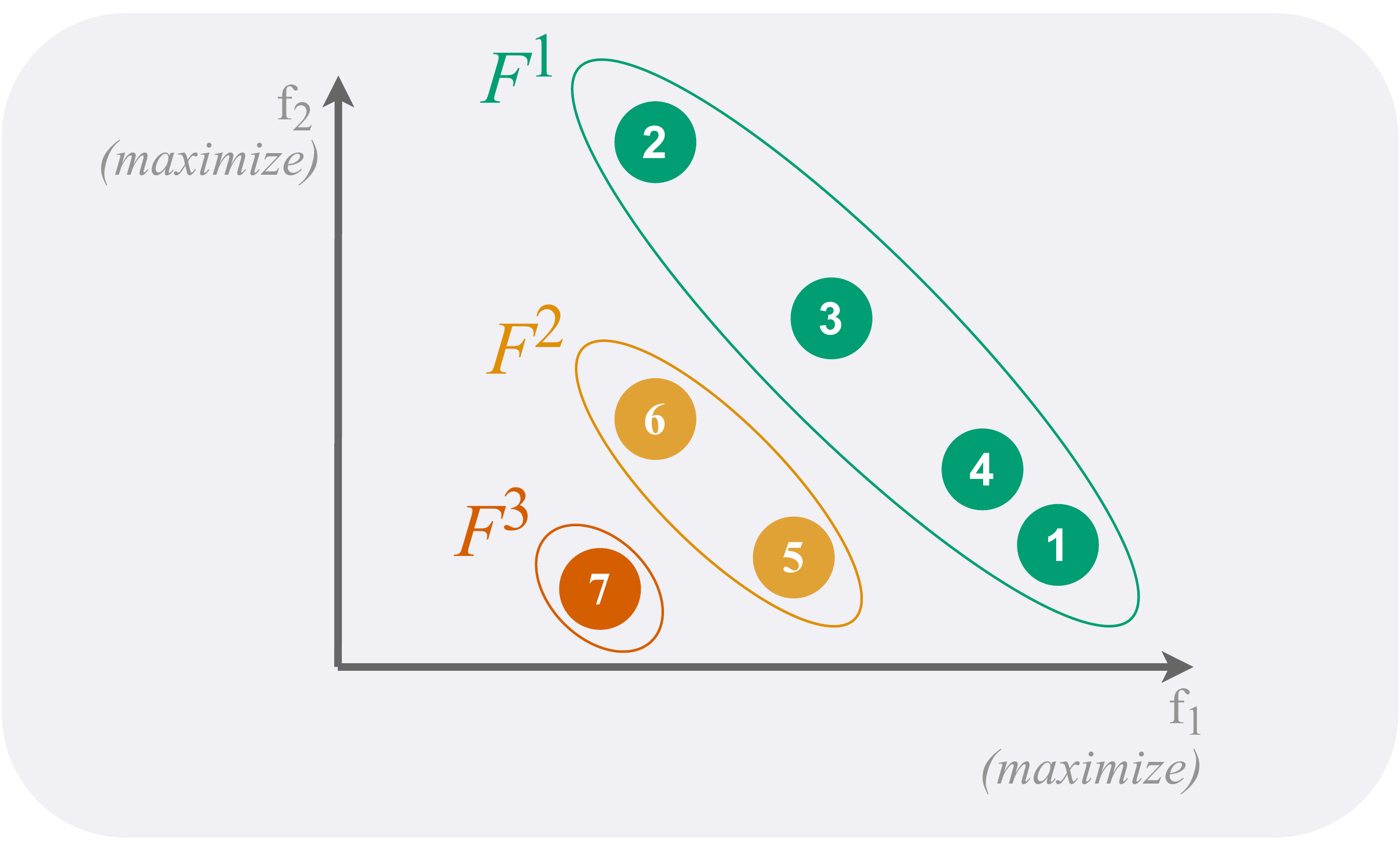}
     \caption{The ranking procedure of solutions (shown with circles) in MO-PBT. Numbers inside circles show the assigned rank (smaller is better). First, solutions are sorted using non-dominated sort. Here, it partitions the solutions into the non-dominated front $F^1$ (green), the second front $F^2$ (light orange), and the third front $F^3$ (dark orange). The solutions in the first front are considered first. The solution with the largest $f_1$ value is ranked first. Then other solutions from $F^1$ are ranked one-by-one such that the solution which is the furthest away from the already ranked ones is picked next. This distance-based ranking is continued in the second (third, etc.) fronts.}
     \label{fig:NDS}
 \end{figure}

\section{Multi-objective hyperparameter optimization tasks}
\subsection{\emph{Precision/Recall} in classification} \label{subsec:task_prrec}
Balancing between precision and recall of a model is a classical trade-off problem in machine learning \cite{karl2022multi, levesque2012multi}. 
In this work, we use modern FT-Transformer (Feature Tokenizer Transformer) neural networks \cite{gorishniy2021revisiting} and three diverse binary classification datasets: Adult \cite{Dua2019}, Higgs \cite{baldi2014searching}, and Click prediction \cite{OpenML2013}.
We optimize the regularization parameters: weight decay and dropout, and additionally the class weights in the cross-entropy-loss function (which is a natural way to balance between class-wise performances, and therefore precision and recall).
The training procedure for FT-Transformer is adopted from \cite{gorishniy2021revisiting} (but without early stopping).
The early stopping is not included because it is not well-suited for a standard PBT setup (also used here), where a predetermined number of exploit-and-explore steps (and therefore training epochs) is performed.

\subsection{Model \emph{Accuracy/Fairness}} \label{subsec:task_tabfairness}
The fairness of a model is understood as its ability to predict the target attribute, e.g., income, without a bias on the sensitive attribute, e.g., gender or race. In our experiments, we consider the standard setup of model fairness in binary classification, where labels $Y \in \{0,1\}$, sensitive attributes $A \in \{0, 1\}$, and model predictions are $\hat{Y}$.
Different fairness metrics have been proposed \cite{garg2020fairness}. Two of the most popular ones are  Statistical Parity (SP) and Equalized Odds (EO). SP requires the independence of predictions $\hat{Y}$ on the sensitive attribute $A$: P$(\hat{Y}|A = 0) = P(\hat{Y} |A = 1)$. EO requires conditional independence of $\hat{Y}$ and $A$ with respect to $Y$: $P(\hat{Y} |A = 1, Y = y) = P(\hat{Y} |A = 0, Y=y)$ for $y \in \{0, 1\}$.

Following \cite{madras2018learning, schmucker2021multi, chuang2021fair}, we optimize the relaxed versions of SP and EO called Difference in Statistical Parity (DSP) and Difference in Equalized Odds (DEO).
$$DSP(f) = |E_{x\sim P_0} f(x)-E_{x\sim P_1} f(x)|$$
$$DEO(f) = \sum_{y \in \{0,1\}}|E_{x\sim P_0^y} f(x)-E_{x\sim P_1^y} f(x)|,$$
where $P_a=P(\cdot|A=a)$ and $P_a^y=P(\cdot|A=a,Y=y)$.

Following \cite{chuang2021fair}, the loss during training can be composed of standard Cross-Entropy (CE) and weighted DSP (gap regularization): 
$$\mathcal{L}_{fairness}(f(x),y)=CE(f(x),y)+\lambda DSP(f),$$ 
where $x$ is a training sample, $y$ is the target, and $\lambda$ is a trade-off parameter.

We consider the Adult dataset with gender as the sensitive attribute and income as the target. We use the same setup as in \ref{subsec:task_prrec} (FT-Transformer neural networks, optimizing regularization), but instead of a class weighting parameter in the cross-entropy loss, we use the $\mathcal{L}_{fairness}$ loss and optimize the $\lambda$ parameter.
Also, we use the CelebA dataset \cite{liu2015faceattributes} with gender as the sensitive attribute and Attractiveness as the binary classification target. The training setup is the same as in Section~\ref{subsec:task_advtraining}, but with the $\mathcal{L}_{fairness}$ loss.

\subsection{\emph{Accuracy/Adversarial robustness}} \label{subsec:task_advtraining}
It has been shown that standard model accuracy and its adversarial robustness (accuracy on samples generated by an adversarial attack) are conflicting objectives 
\cite{zhang2019theoretically}.
In this task, we use the TRADES loss \cite{zhang2019theoretically}:
\begin{multline*}
\mathcal{L}_{TRADES}(f(x),y)=CE(f(x),y)\\+\max_{x' \in \mathcal{B}(x,\epsilon)}\lambda CE(f(x),f(x')),
\end{multline*}
where $x$ is a training sample, $x'$ is a generated adversarial sample in the $\epsilon-$neighborhood of $x$, and $y$ is the target.  The parameter $\lambda$ affects the trade-off between accuracy and adversarial robustness.
We use the same adversarial attack and TRADES loss parameters as in \cite{zhang2019theoretically}.
We search for data augmentation parameters: parameters of the RandAugment augmentation strategy \cite{cubuk2020randaugment} and Cutout \cite{devries2017improved} (probability and size). 
Experiments are performed for CIFAR-10/100 datasets using the WideResNet-28-2 \cite{zagoruyko2016wide} and the training setup from \cite{zhang2019theoretically}.

\subsection{Search spaces}
In this work, we perform search in discretized search spaces. Such an approach was successfully used, for instance, for augmentations search \cite{ho2019population, cubuk2020randaugment}. For all described optimization tasks, search spaces of hyperparameters are specified in Appendix \ref{sec:appendix_searchspaces}.

\section{Experimental setup} \label{sec:experiments}

\subsection{PBT operators} \label{pbt_operators}
Here we describe the operators of MO-PBT following the notation from \cite{jaderberg2017population}.

\textbf{Exploit} We use the simple truncation selection operator used in the original PBT \cite{jaderberg2017population} algorithm. After the population is sorted according to some criterion (non-dominated sort followed by the greedy scattered subset selection in the case of MO-PBT),  each of the bottom $\tau$\% of solutions in the population is randomly replaced by a solution from the top $\tau$\% (we use the default value of $\tau$ is 25). The pseudocode of the used \emph{exploit} operator is listed in Appendix~\ref{sec:appendix_codes}, Algorithm~\ref{alg:mopbt_exploit}.

\textbf{Explore} We use the \emph{explore} operator previously used in Population Based Augmentations \cite{ho2019population}. It assumes that the encoding of hyperparameter values in a search space is ordinal. The key idea of the operator is locality: the new value of a hyperparameter is chosen from the vicinity of the current value. The pseudocode of the used \emph{explore} operator is listed in Appendix~\ref{sec:appendix_codes}, Algorithm~\ref{alg:mopbt_explore}.

\textbf{Ready} In all considered tasks, we perform the exploit-and-explore procedure every 2 epochs of training.

We use a population of size 32 in our main experiments and in Section~\ref{sec:scalability} study how the performance scales with increasing population size.
Note that we do not specifically tune \emph{exploit} and \emph{explore} operators of MO-PBT, but in Appendix~\ref{sec:appendix_ablation} we analyze how their design impacts the performance and conclude that the considered design options perform similarly.
 
\subsection{Hypervolume as the performance metric}  \label{subsec:performance_metric}
We use the hypervolume, a commonly used metric in MO optimization \cite{riquelme2015performance} (see Section \ref{subsec:mo_optimization}). We calculate the reference point $r=(r_1,\dots,r_K)$ with the following approach, which is used, for instance, in \cite{knowles2006parego, ishibuchi2011many}. First, all non-dominated fronts are collected from all evaluation points of all algorithms and all performed runs and stored in a set $\mathcal{F}$. Then, the reference point $r$ is calculated as $r_i=\min_{x \in \mathcal{F}}{f(x_i)}-\rho(\max_{x \in \mathcal{F}}{f(x_i)}-\min_{x \in \mathcal{F}}{f(x_i)}), \text{for~} i=1,\dots,K$, where $\rho$ is typically set to a small value, here we use $\rho=0.1$. 
This strategy selects the reference point that is guaranteed to be worse than all points on all fronts. This ensures that all non-dominated points are considered in the hypervolume calculation and are not discarded. Furthermore, the reference point is shifted with respect to the range of values of each objective in order to prevent its positioning too far away from the fronts.

In the experimental evaluation, we use the following common metric (used, e.g., in \cite{schmucker2021multi, daulton2021parallel}. 
First, we obtain an approximation of the Pareto front by collecting all evaluated solutions from all runs of all algorithms and selecting the non-dominated subset $\mathcal{P}^*$ of them. The approximation of the optimal hypervolume is then calculated as $HV^*=Hypervolume(\mathcal{P}^*)$. The reported performance metric of an algorithm run $r$ at timestamp $t$ is the logarithmic difference of the hypervolume of a non-dominated set of solutions obtained by this timestamp and the ideal hypervolume: $log_{10}(HV^*-HV^t_r)$. Finally, this metric is averaged over multiple runs.

Datasets are split into train/validation/test subsets before experiments. In our main results, we report the above-described hypervolume metric on the \emph{validation} subset to evaluate the search performance of the algorithms. Additionally, we provide results on the test subsets in Appendix~\ref{sec:appendix_generalization}.

\subsection{Baselines}
\subsubsection{Random search} First, we consider a trivial search baseline --- random search: for each hyperparameter, a random value is sampled at the beginning of model training.

\subsubsection{Single-objective PBT} We use modifications of PBT that convert an MO problem into a SO one. First, we use one of the objectives as the fitness function of PBT. Comparing against this baseline can show that the considered MO problems are challenging, and optimizing just one objective is inferior to using MO techniques. Secondly, we implement different scalarization functions in PBT. The first technique we use is random scalarization as, for instance, in the ParEGO algorithm \cite{knowles2006parego}: at each invocation of the evaluation procedure, the scalarization vector is sampled randomly. Here we use the ParEGO scalarization function (as defined in Section~\ref{subsec:scalarization}) as it was originally proposed to use for random scalarizations in \cite{knowles2006parego}. Secondly, we use the maximum scalarization technique proposed in \cite{schmucker2020multi}: the objective value is calculated as $\max_{w\in W, ||w||=1} V(f(x),w)$, where W is a set of randomly sampled unit vectors and V is a scalarization function. Following \cite{schmucker2021multi}, we use $|W|=100$ and the Golovin scalarization, which was demonstrated to outperform other scalarization functions.

\subsubsection{MO-ASHA variants} We consider MO-ASHA with greedy scattered subset selection ranking ($\epsilon$-network), which was shown to perform better than alternative MO-ASHA variants in \cite{schmucker2021multi}. Secondly,  to compare MO-PBT against a strong BO baseline that is well parallelizable we adapt the MO-BOHB \cite{guerrero2021bag} approach to MO-ASHA. 
We refer to this MO-ASHA modification as BO-MO-ASHA.

\subsection{Evaluation setup} \label{subsec:evaluation-setup}
The main design principle of our evaluation of algorithms is to compare the achieved performance with respect to elapsed wall-clock time instead of the performed number of training epochs. We choose this approach because in practice we are more interested in the achieved performance by a specific time point rather than a specific epoch. We do not set a time limit for all PBT variants and random search but rather allow them to fully finish the training cycle of all solutions in the population. For MO-ASHA, we allocate the time budget equal to the run time of the slowest PBT run.
We ran each algorithm 10 times on tabular datasets (Adult, Higgs, Click prediction), and 5 times on image ones (CIFAR-10/100, CelebA).
When plotting performance over time, we plot mean performance, with the area between the worst and the best runs shaded.

Further experimental setup details are provided in Appendix,~\ref{sec:appendix_implementation}. The code is available at \url{https://github.com/ArkadiyD/MO-PBT}. 

\section{Results}

\subsection{Overall performance}

\begin{figure}[h!]
    \centering
    \includegraphics[width=4cm]{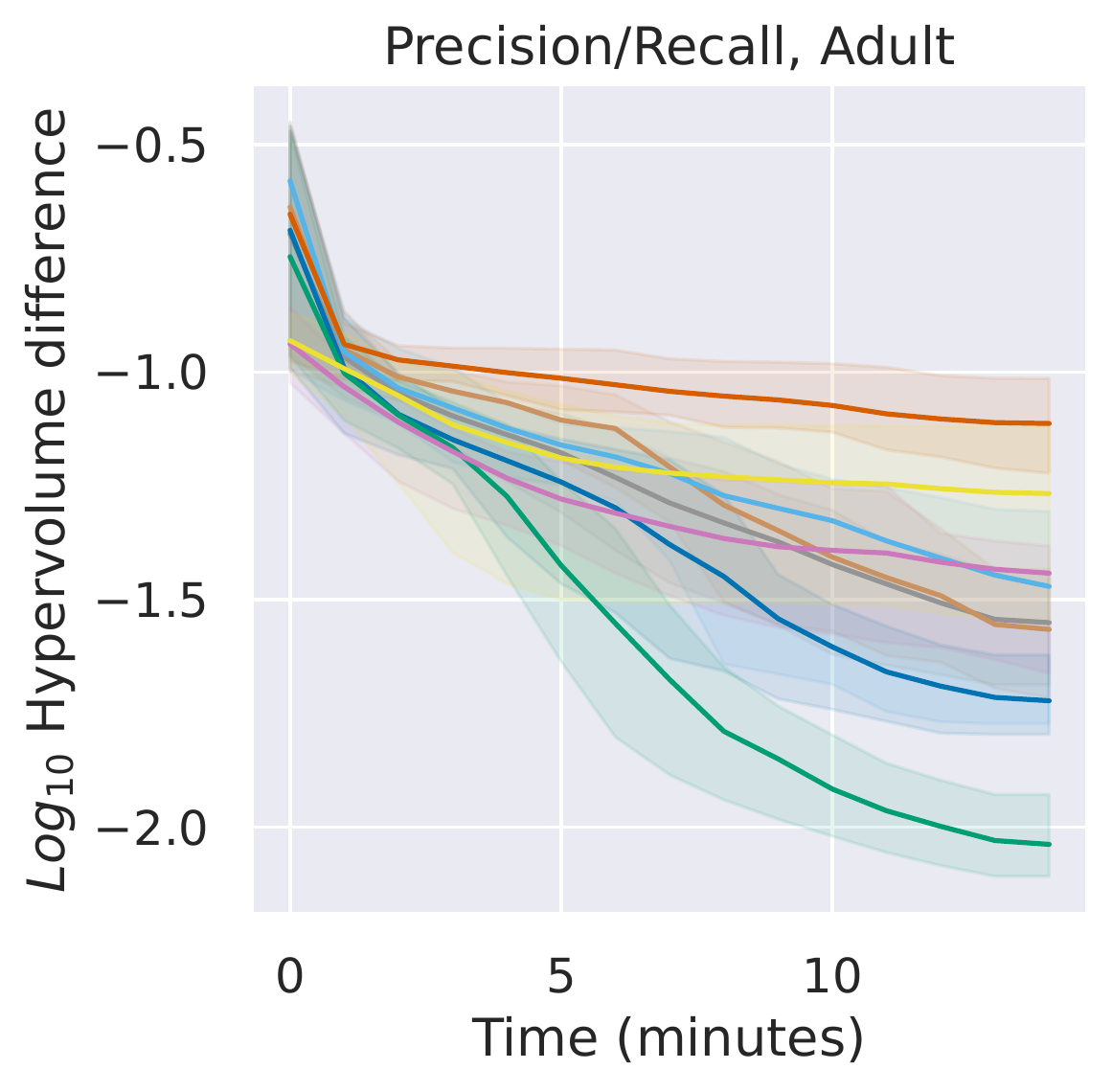}~   \includegraphics[width=4cm]{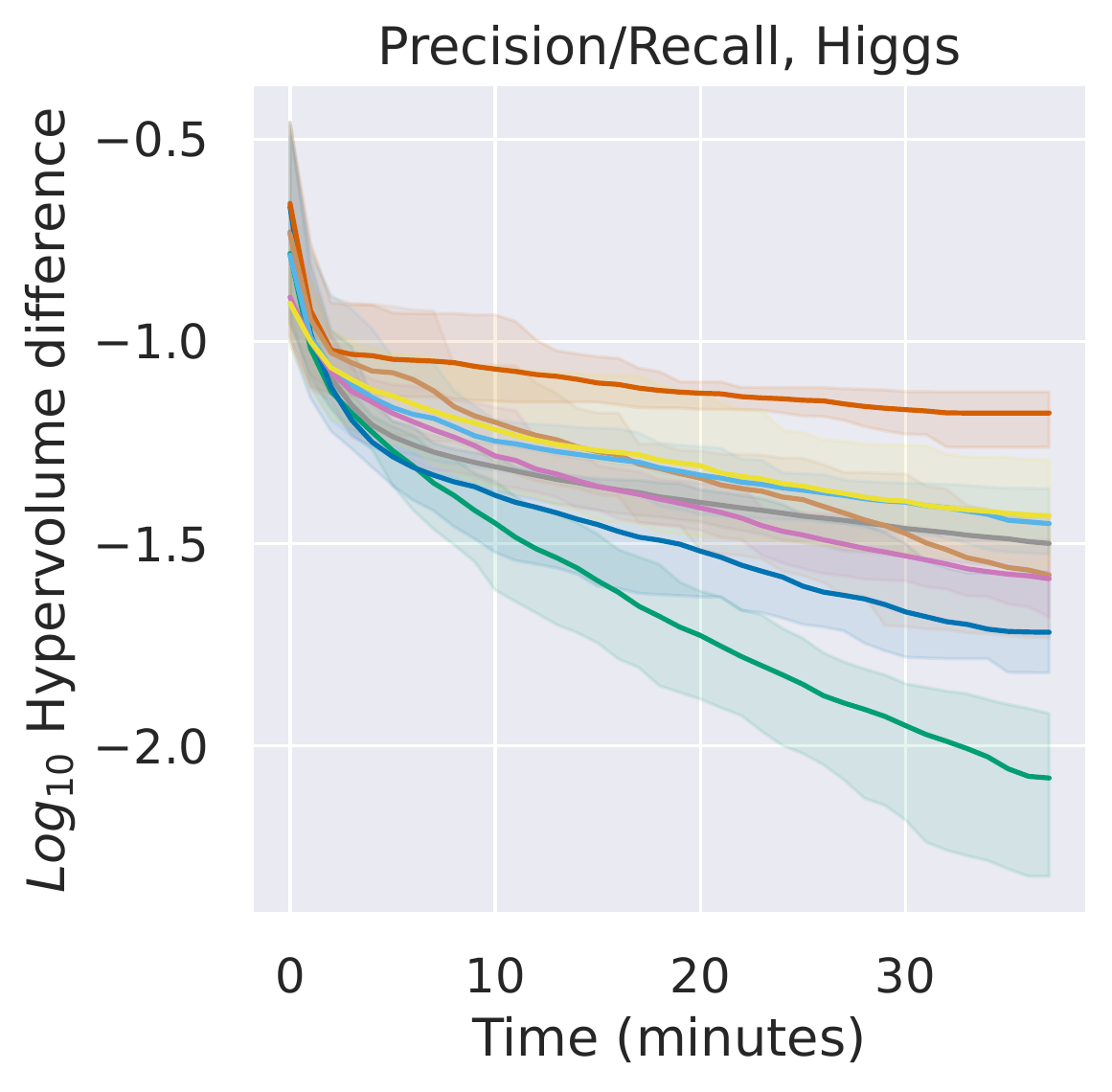}
  
    \includegraphics[width=4cm]{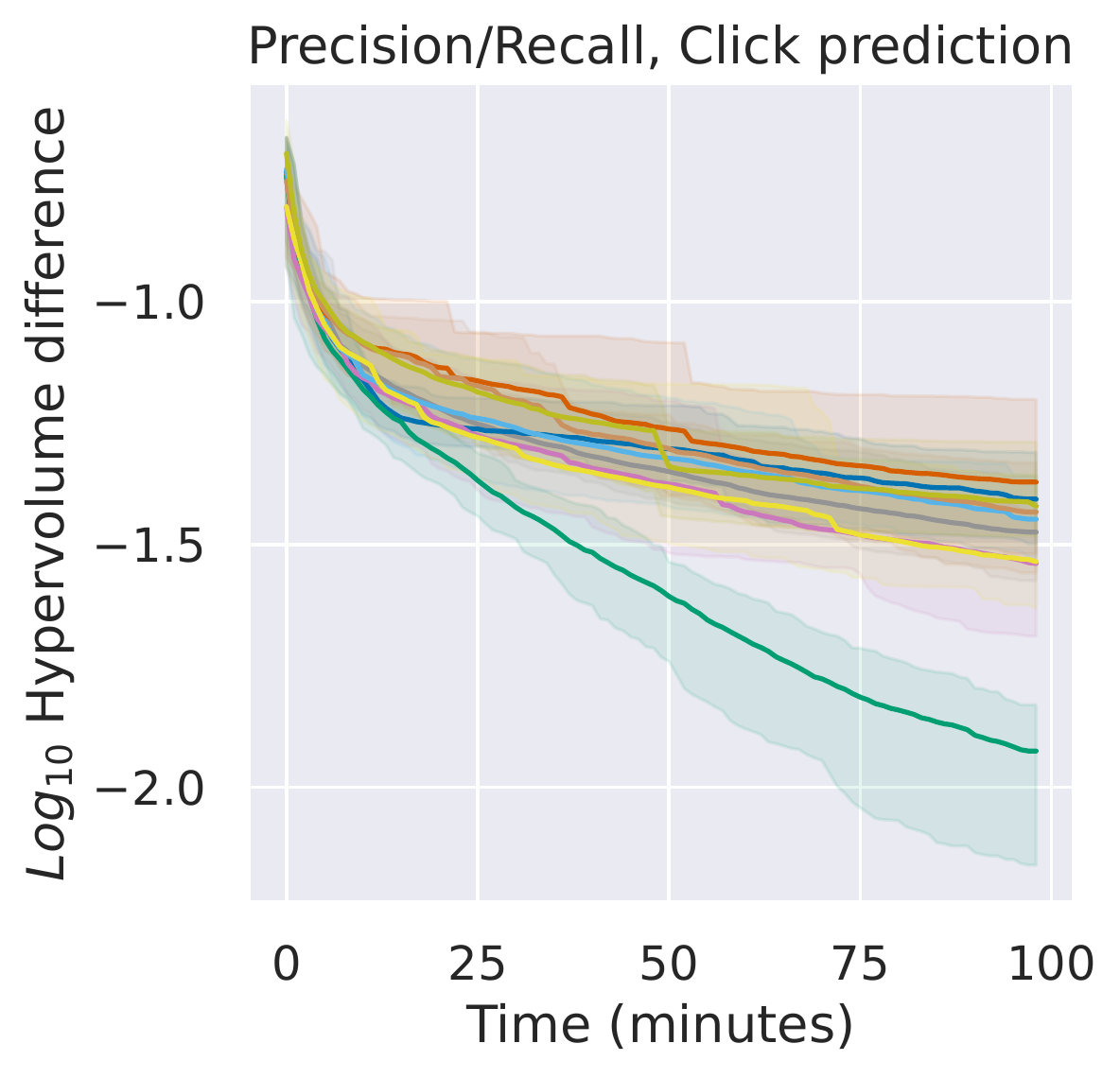}
    \includegraphics[width=0.45\textwidth]{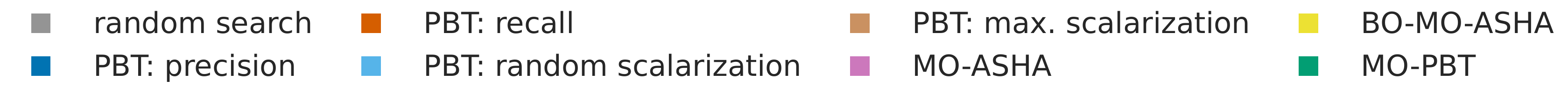}

    \caption{Optimization results on the Precision/Recall task.}
    \label{fig:results1}
    \vspace{-1pt}
\end{figure}

\begin{figure}[h!]
    \centering
    \includegraphics[width=4cm]{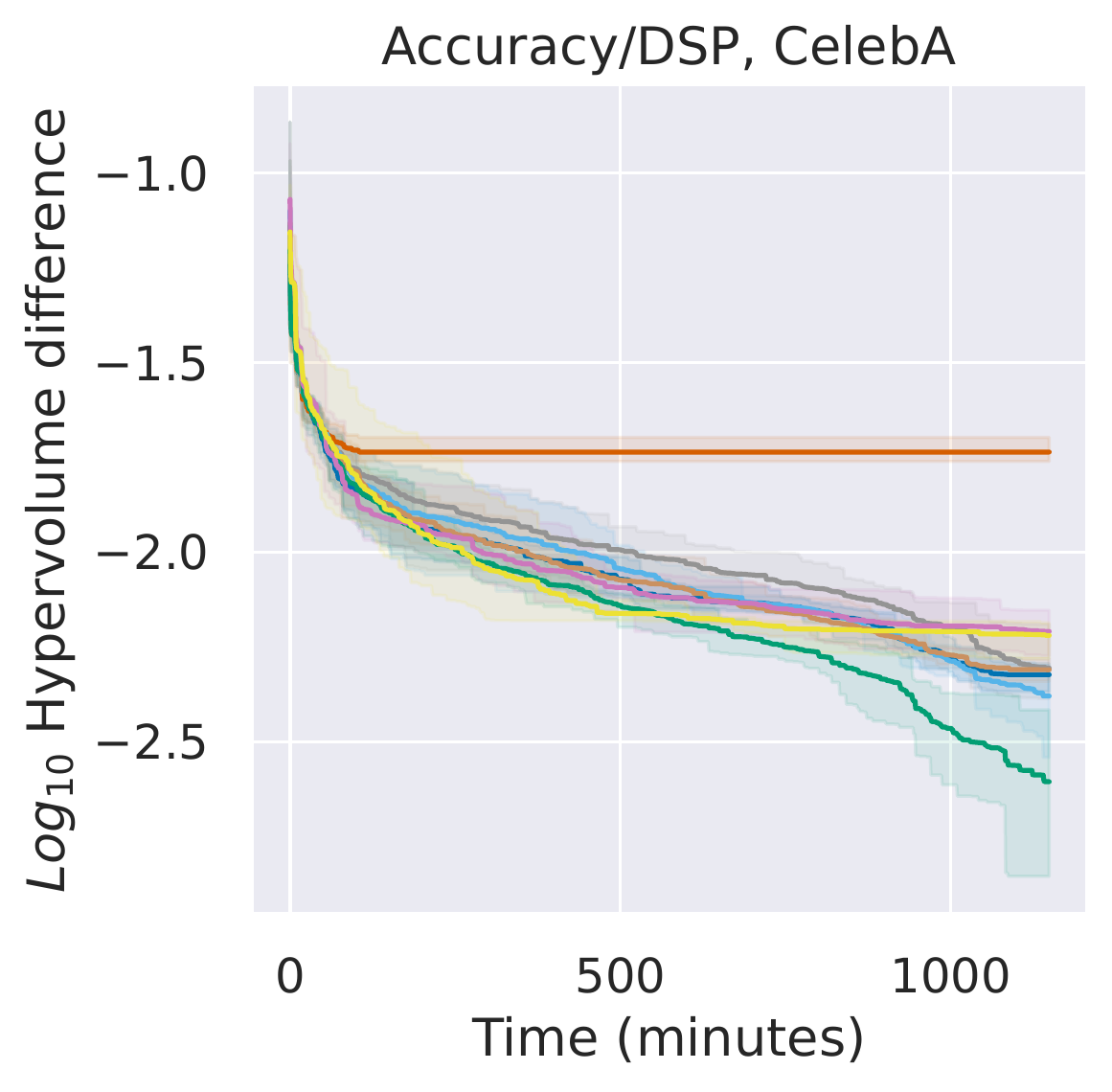}
    \includegraphics[width=4cm]{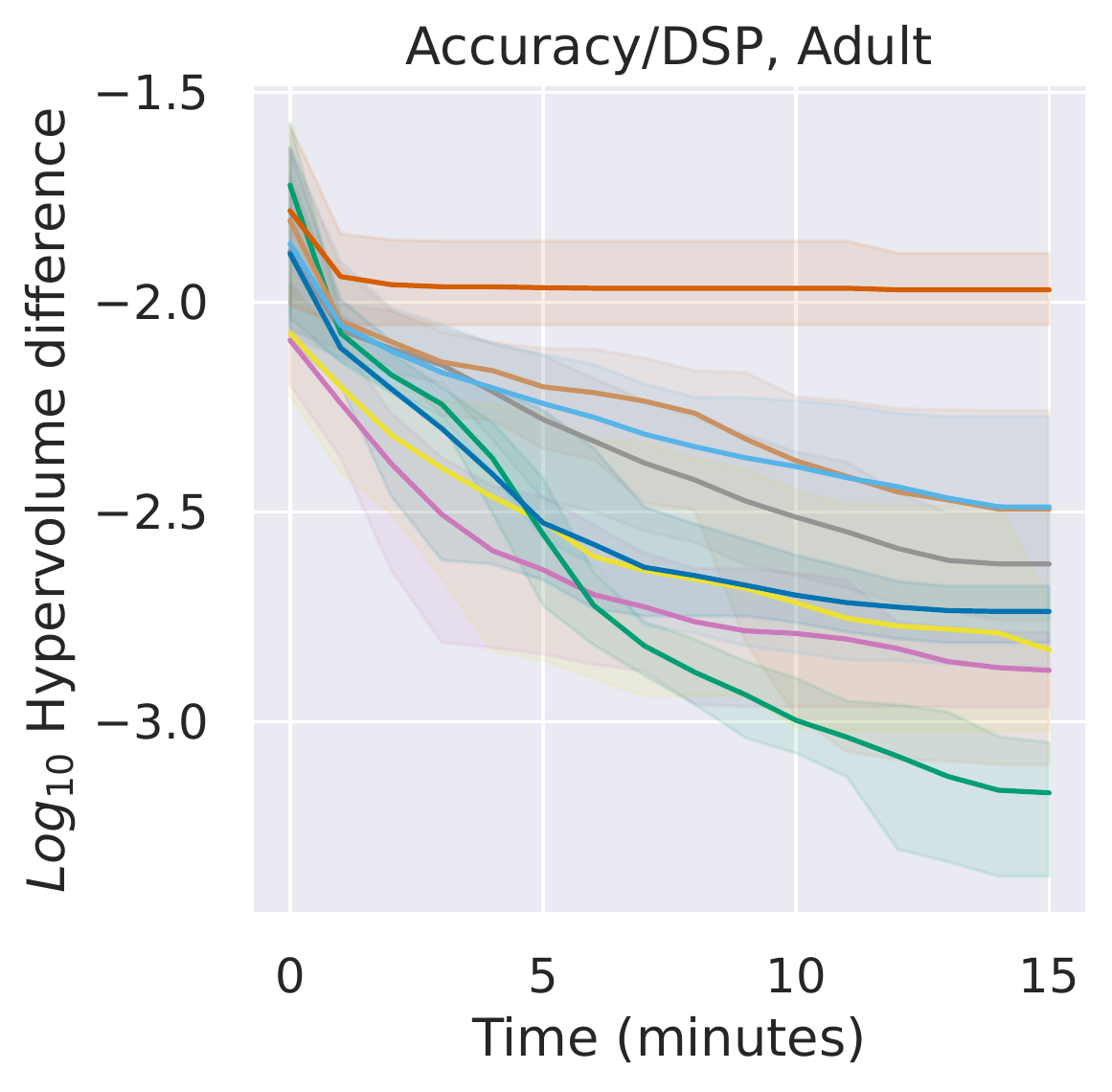}    
\vspace{0.1cm}
    \includegraphics[width=4cm]{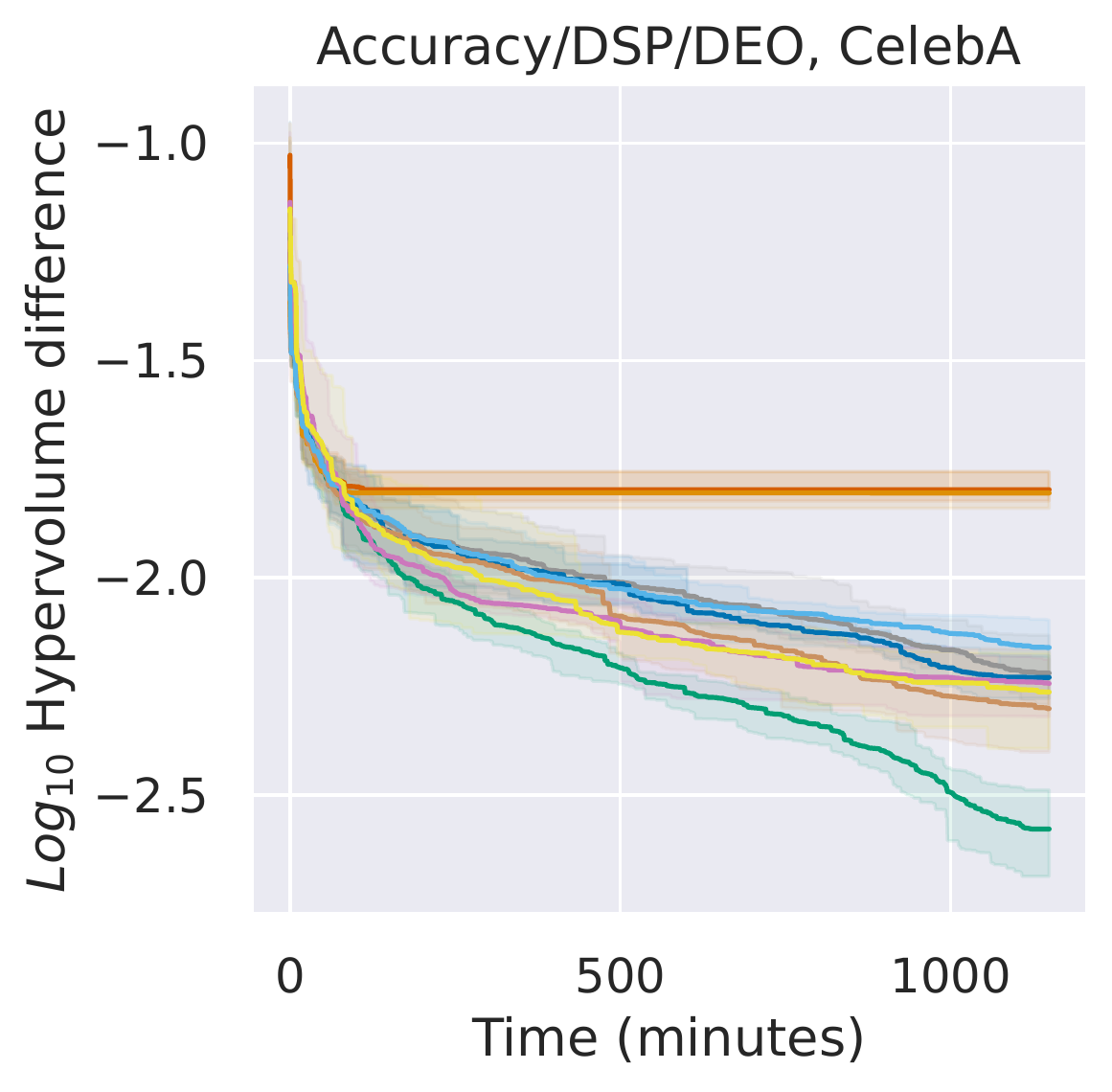}
    \includegraphics[width=4cm]{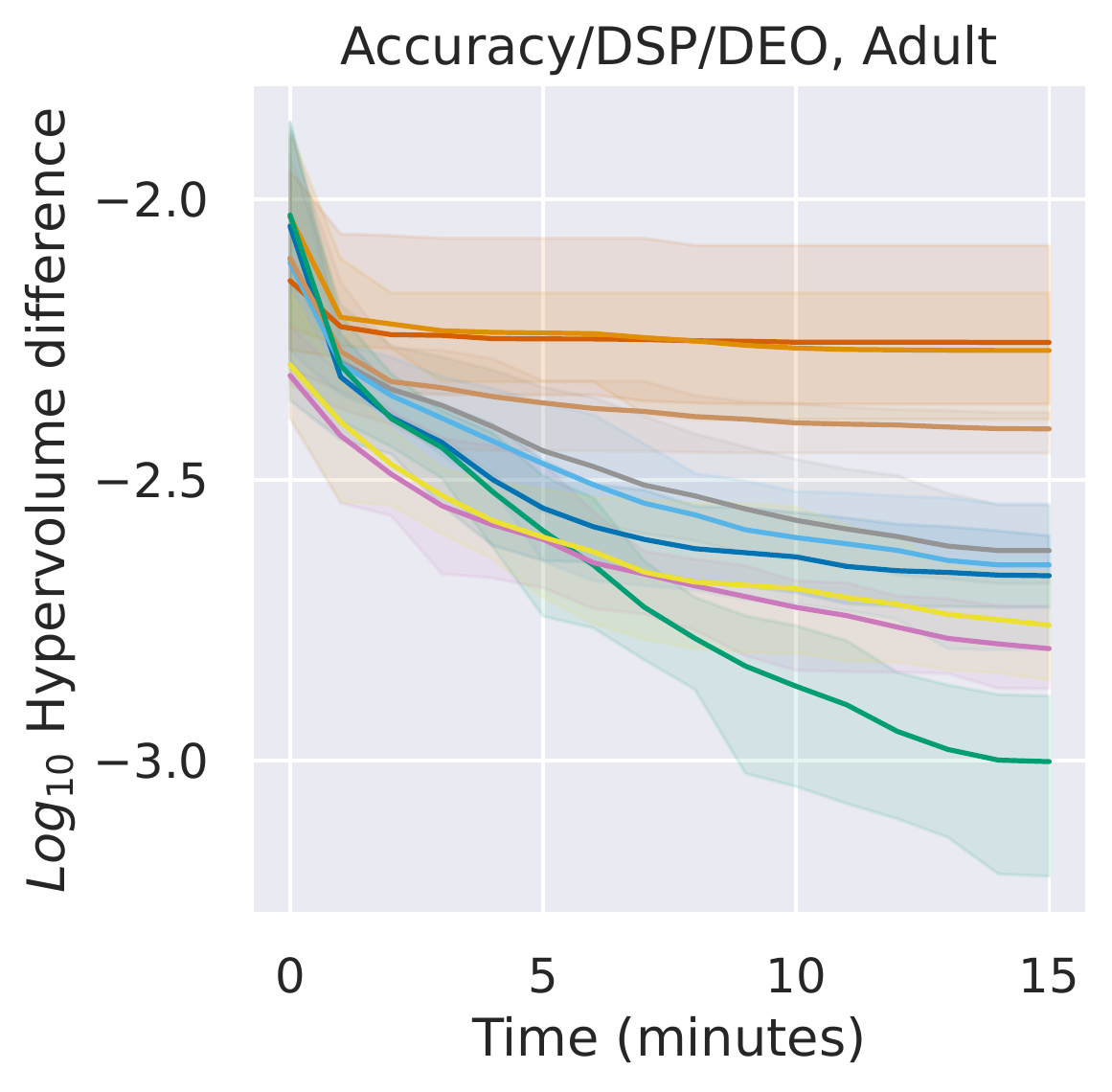}
    \includegraphics[width=0.45\textwidth]{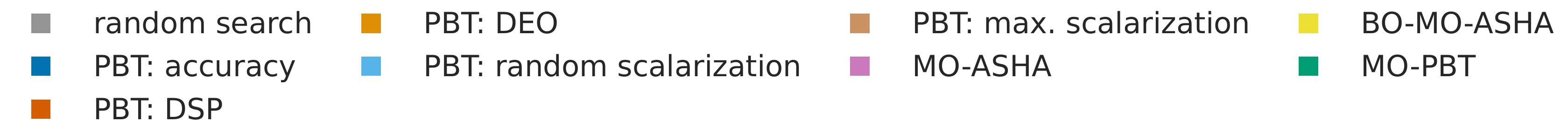}
   \caption{Optimization results on the Accuracy/Fairness task.}
    \label{fig:results2}
    \vspace{-10pt}
\end{figure}

\begin{figure}[h!]
    \centering
    \includegraphics[width=4cm]{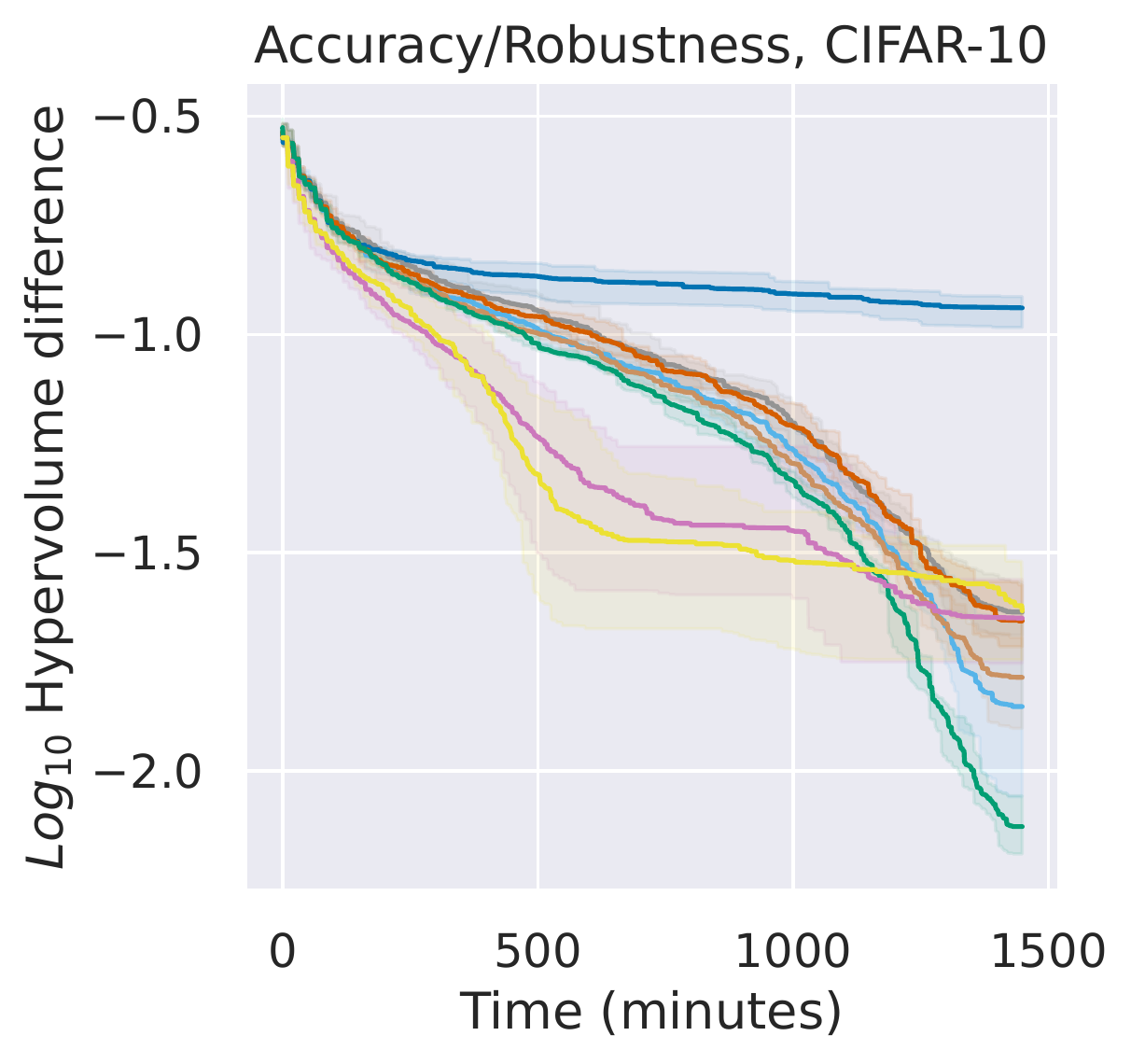}
    \includegraphics[width=4cm]{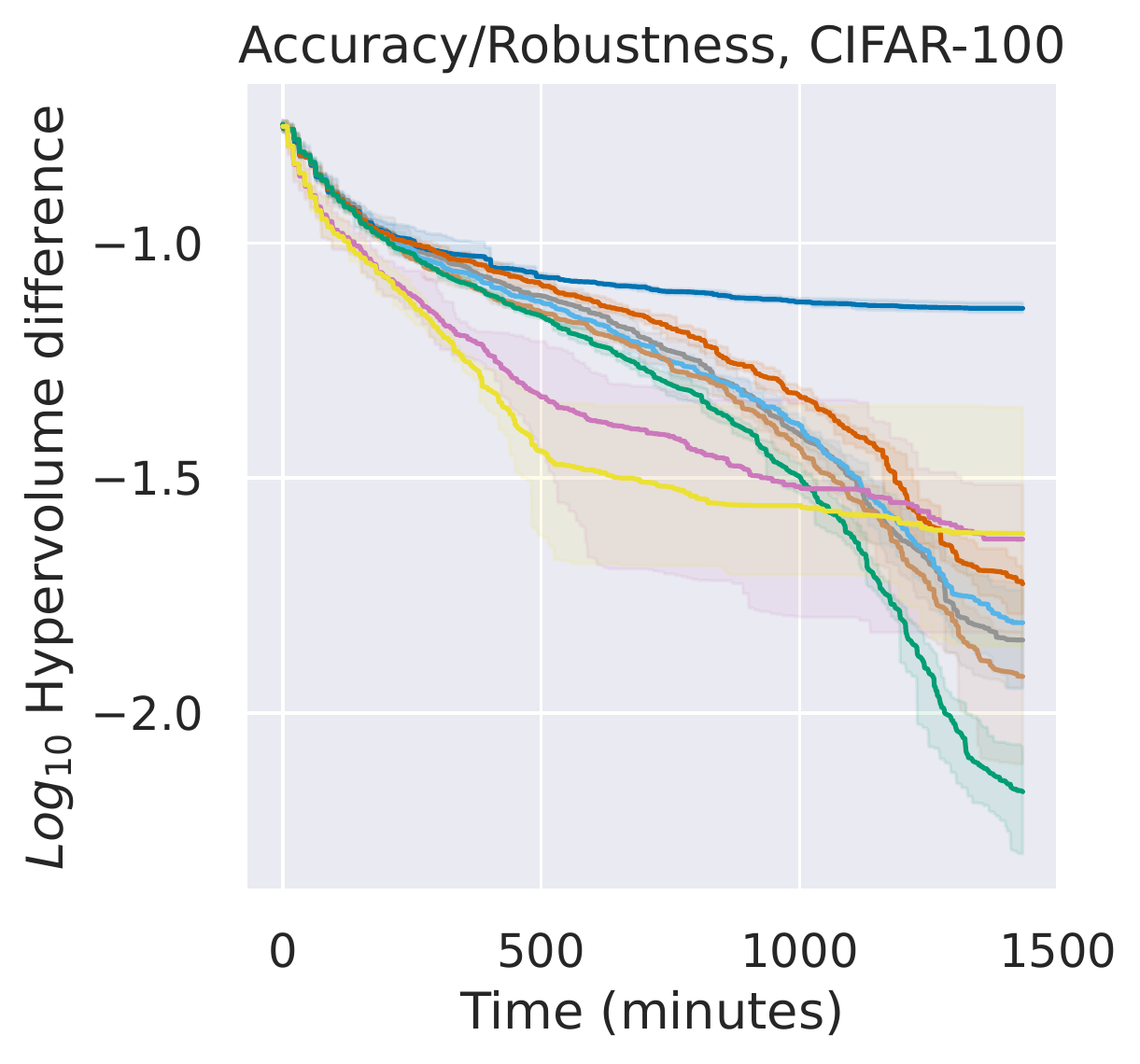}    
    \includegraphics[width=0.45\textwidth]{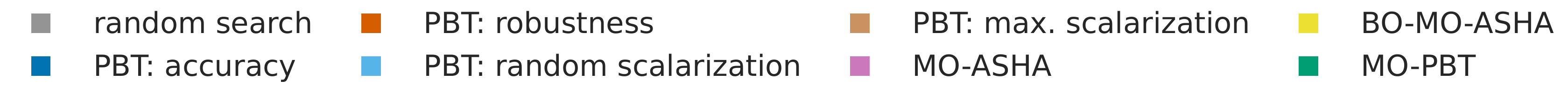}
    \caption{Optimization results on the Accuracy/Robustness task.}
    \label{fig:results3}
    \vspace{-10pt}
\end{figure}

Results of hypervolume-based performance evaluation (as described in \ref{subsec:performance_metric}) are shown in Figures~\ref{fig:results1},\ref{fig:results2},\ref{fig:results3}. On every considered task, MO-PBT outperforms baselines (the standard deviations of the hypervolume are provided in Appendix~\ref{sec:appendix_tables}, Table~\ref{tab:hv_results}). Noteworthy, on the three-objective problems MO-PBT is also the best-performing algorithm. The consistently good performance of MO-PBT on the considered diverse tasks empirically demonstrates its generality.

On Accuracy/Fairness tasks, optimizing the fairness objective with SO PBT leads to obtaining mostly inaccurate models, and, therefore, poor hypervolume values. Similarly, on the Accuracy/Robustness task, if only accuracy is optimized, the results achieved for the robustness objective are poor. PBT with scalarization techniques performs, in general, better than single-objective PBT.

Comparing MO-ASHA variants, we cannot conclude that BO-MO-ASHA performs better than MO-ASHA.
We note that on CIFAR-10/100 Accuracy/Robustness tasks (Figure~\ref{fig:results3}), MO-ASHA and BO-MO-ASHA
perform better than MO-PBT in the beginning of the search as they train networks in a different order compared to MO-PBT: some selected networks are fully trained earlier in time than in MO-PBT, where all the networks are trained simultaneously. However, as soon as the population is trained for more epochs, MO-PBT catches up and at the end of the search substantially outperforms MO-ASHA. We note that this behavior occurs only because the number of parallel workers in our experiments is smaller than the population size.

\subsection{Analysis of the obtained trade-off fronts}
\begin{figure}[h!]
    \centering
    \includegraphics[width=4cm]{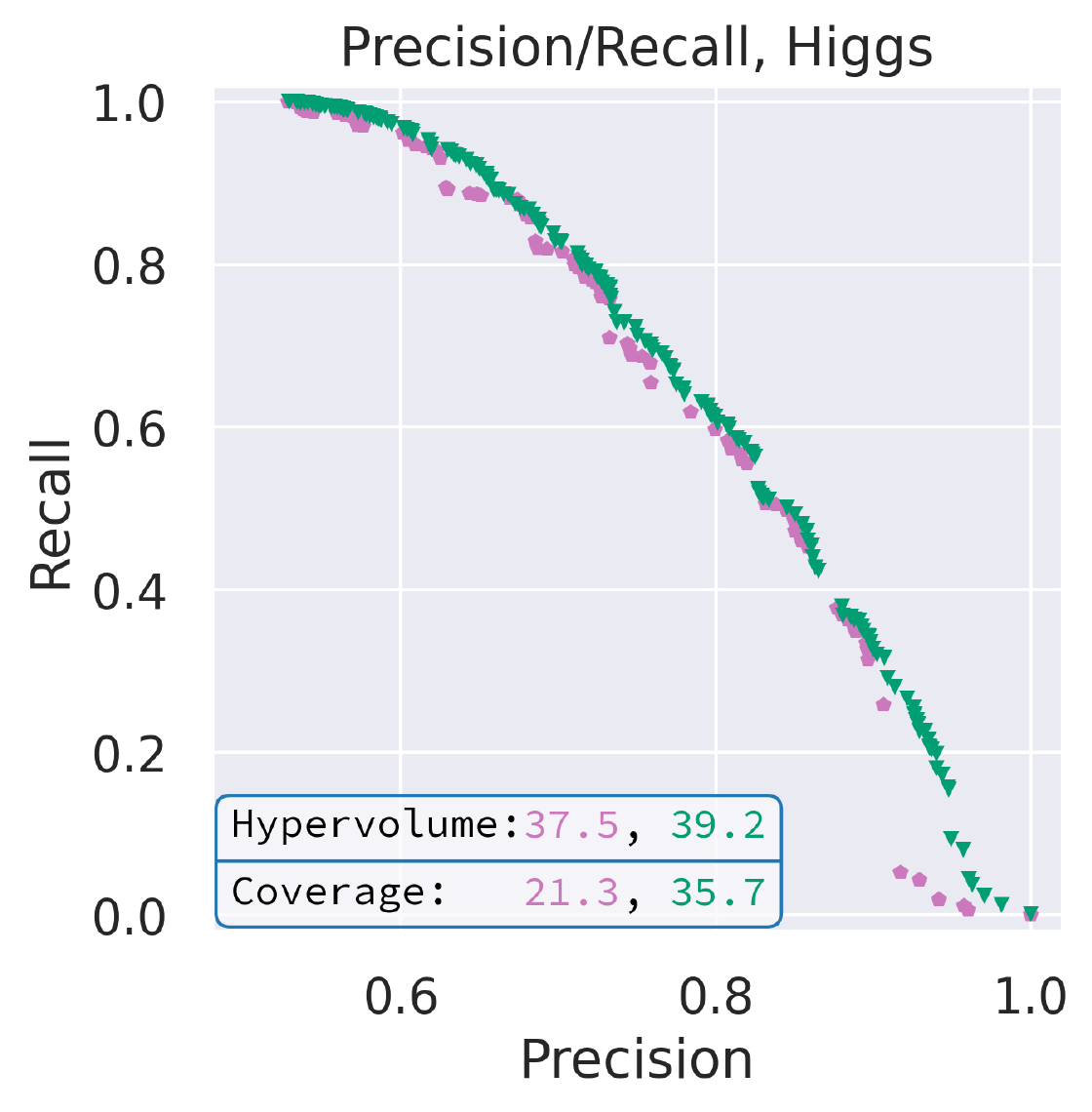}
    \includegraphics[width=4cm]{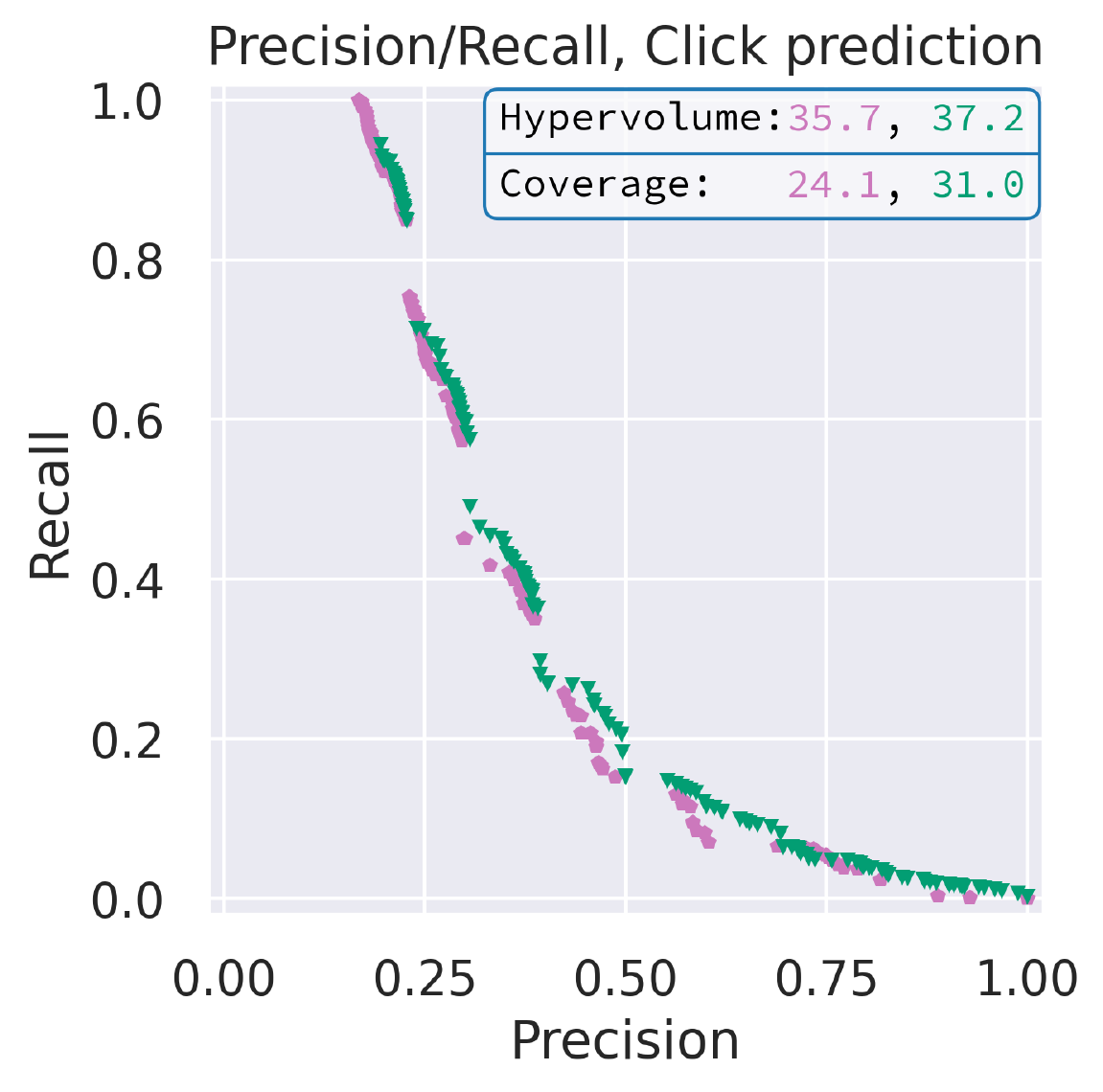}
    \includegraphics[width=4cm]{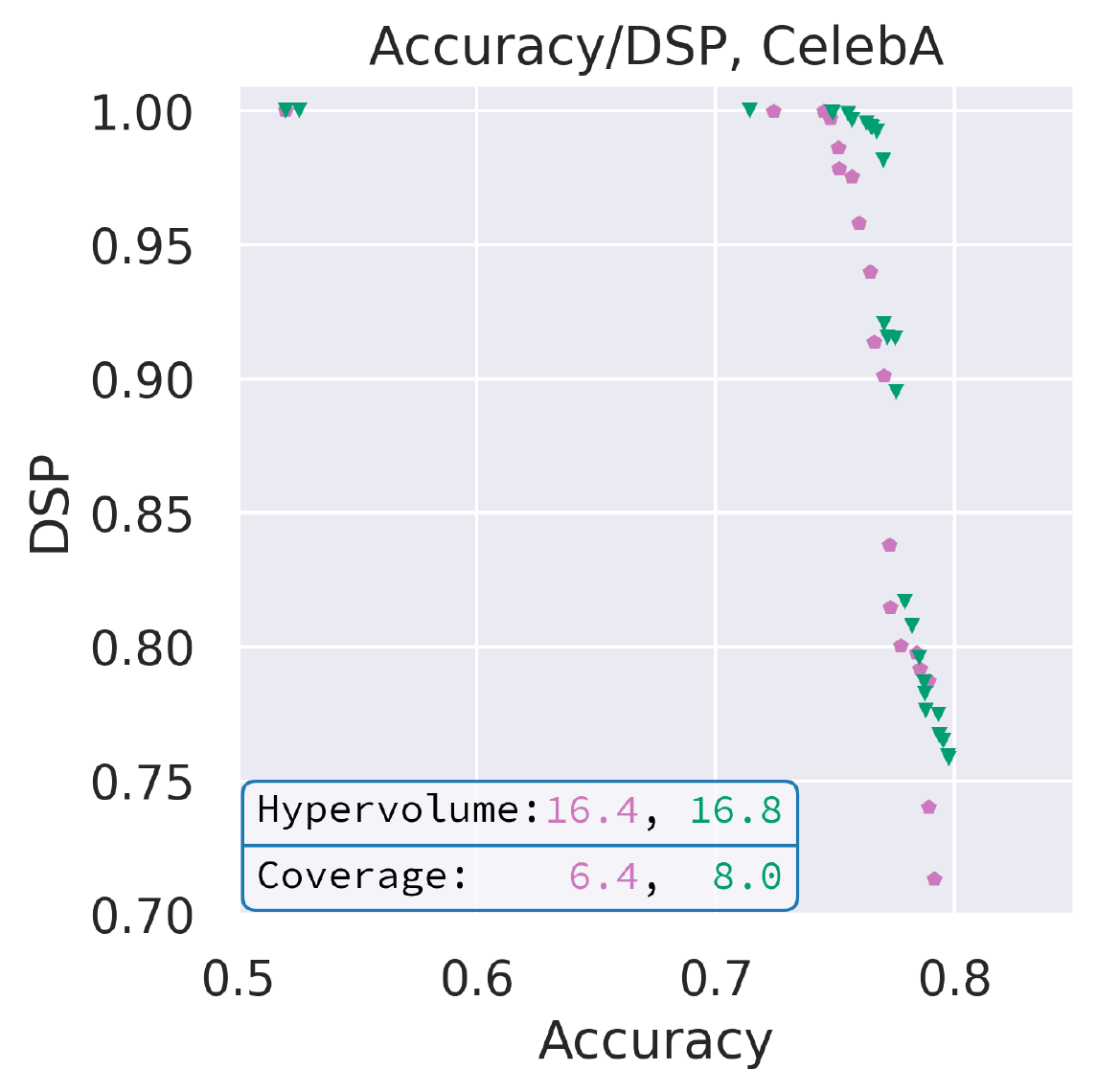}
    \includegraphics[width=4cm]{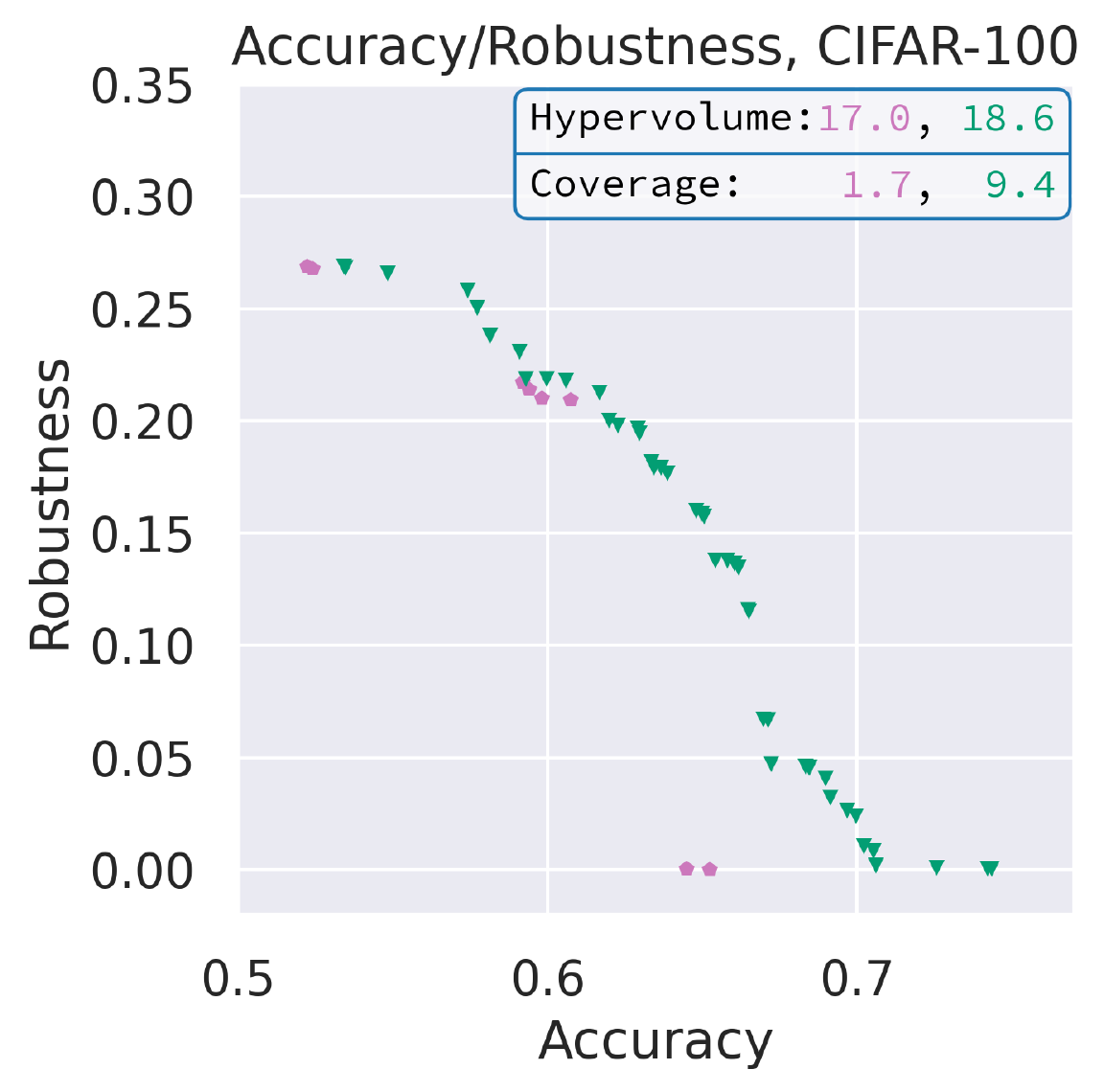}
    \includegraphics[width=0.25\textwidth]{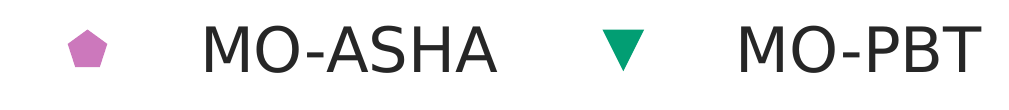}
    
    \caption{Comparison of the non-dominated fronts obtained by different algorithms. For each algorithm, the run with the median hypervolume value is shown. For each front, the values of the hypervolume and coverage metrics (multiplied by 100) are reported in the corresponding color.
    }
    \label{fig:fronts1}
\vspace{-10pt}
\end{figure}

The comparison of non-dominated fronts of solutions obtained by different algorithms is shown in Figure~\ref{fig:fronts1}. Quantitative results of the front diversity evaluation (using the coverage metric introduced in \cite{scriven2009dynamic} and described in Appendix~\ref{sec:appendix_metrics}) are shown in Appendix~\ref{sec:appendix_tables}, Table~\ref{tab:coverage_results}. 

We observe that on the Precision/Recall and the Accuracy/Robustness tasks, MO-PBT achieves substantially better coverage of the trade-off front compared to other algorithms. On Accuracy/DSP tasks, MO-ASHA on average has slightly better coverage than MO-PBT (though the variance of the results is large and in some runs MO-PBT has better coverage). Nevertheless, it should be noted that the quality of the most points on the trade-off fronts obtained by MO-ASHA is worse in terms of domination (can be seen on Figure~\ref{fig:fronts1}).

These results demonstrate that MO-PBT can not only find solutions closer to the reference front than MO-ASHA (which is reflected in the better hypervolume performance), but also produce more diverse fronts along the entire trade-off curve. For practical usage, this means that more options for trade-offs between objectives are available for the user to choose from.

\subsection{Where do different algorithms focus their search?}
\begin{figure*}[ht!]
    \centering
    \includegraphics[width=0.9\textwidth]{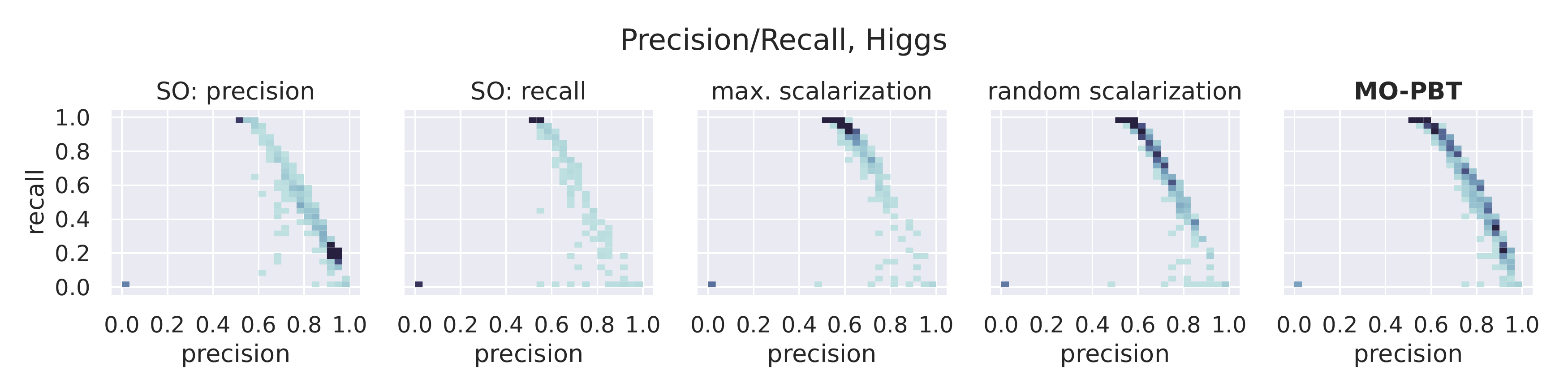}%
    \vspace{-0.1cm}
    \includegraphics[width=0.9\textwidth]{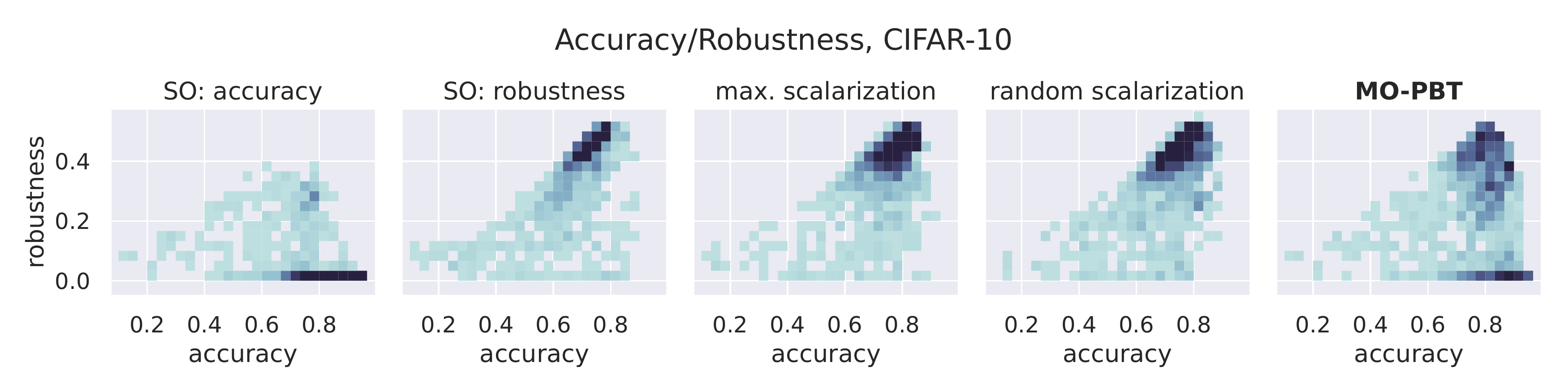}%
    
    \caption{2D histograms of solutions (in the objective space) collected during one run of each algorithm. Darker color denotes more solutions in the corresponding bin (for visualization purposes, bins with more solutions than the 95th percentile of the bin counts have the darkest color on the plot). For each algorithm, solutions obtained during the run with the median hypervolume value are plotted. 
    SO denotes PBT applied to optimizing one of the objectives.}
    \label{fig:densities1}
\end{figure*}

We analyze how algorithms differ in their search behavior by plotting all solutions collected during the search in the objective space and highlighting areas where more solutions are concentrated. These plots are shown in Figure~\ref{fig:densities1}. They show a clear difference between approaches which turn an MO problem into an SO one (scalarization and optimizing one objective) and MO-PBT. MO-PBT obtains solutions scattered more uniformly along the entire trade-off trajectory between two objectives, in contrast to concentrating on one area of it. 

\subsection{Scalability} \label{sec:scalability}
\subsubsection{Population size}
\begin{figure}[h!]
    \centering
    \includegraphics[width=4cm]{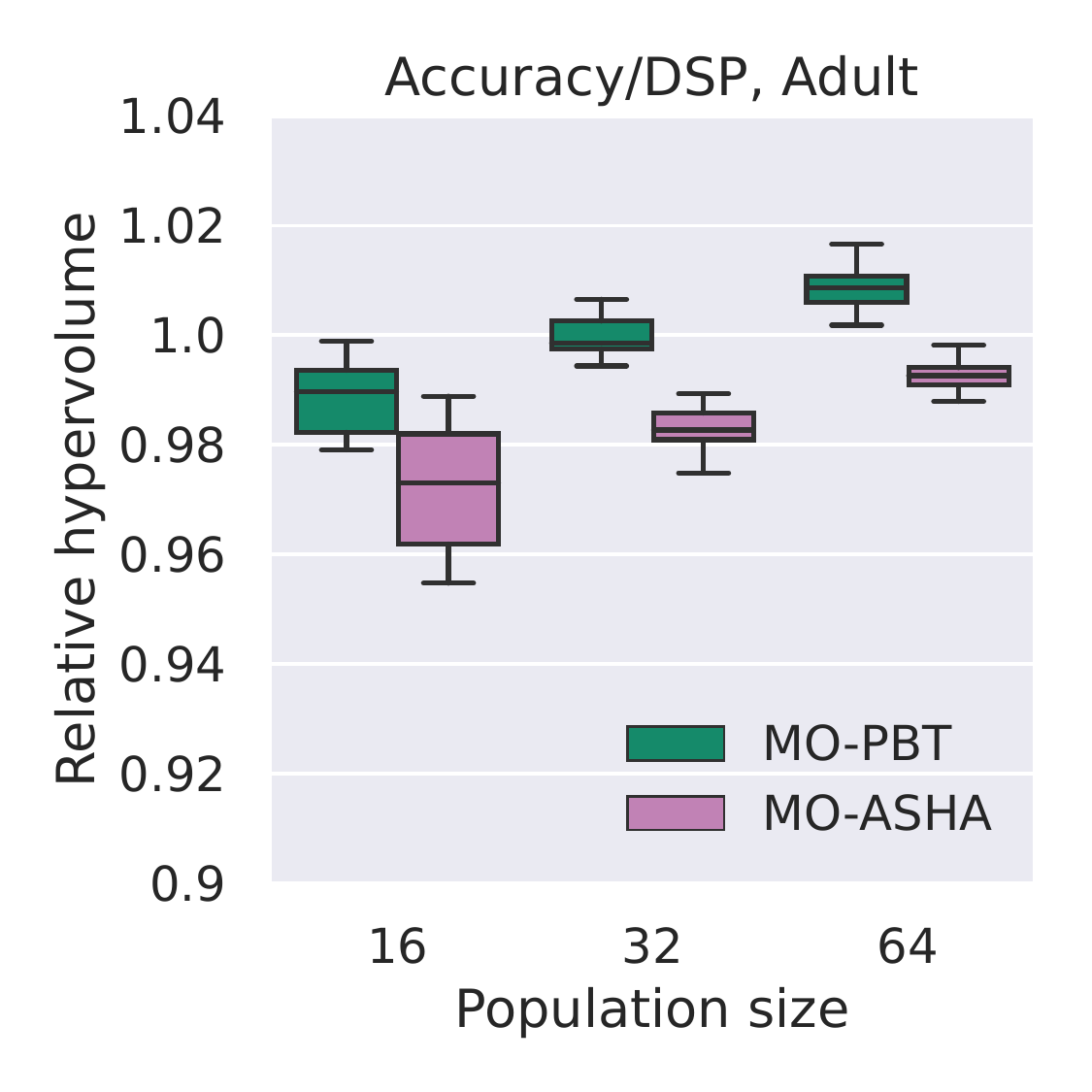}
    \includegraphics[width=4cm]{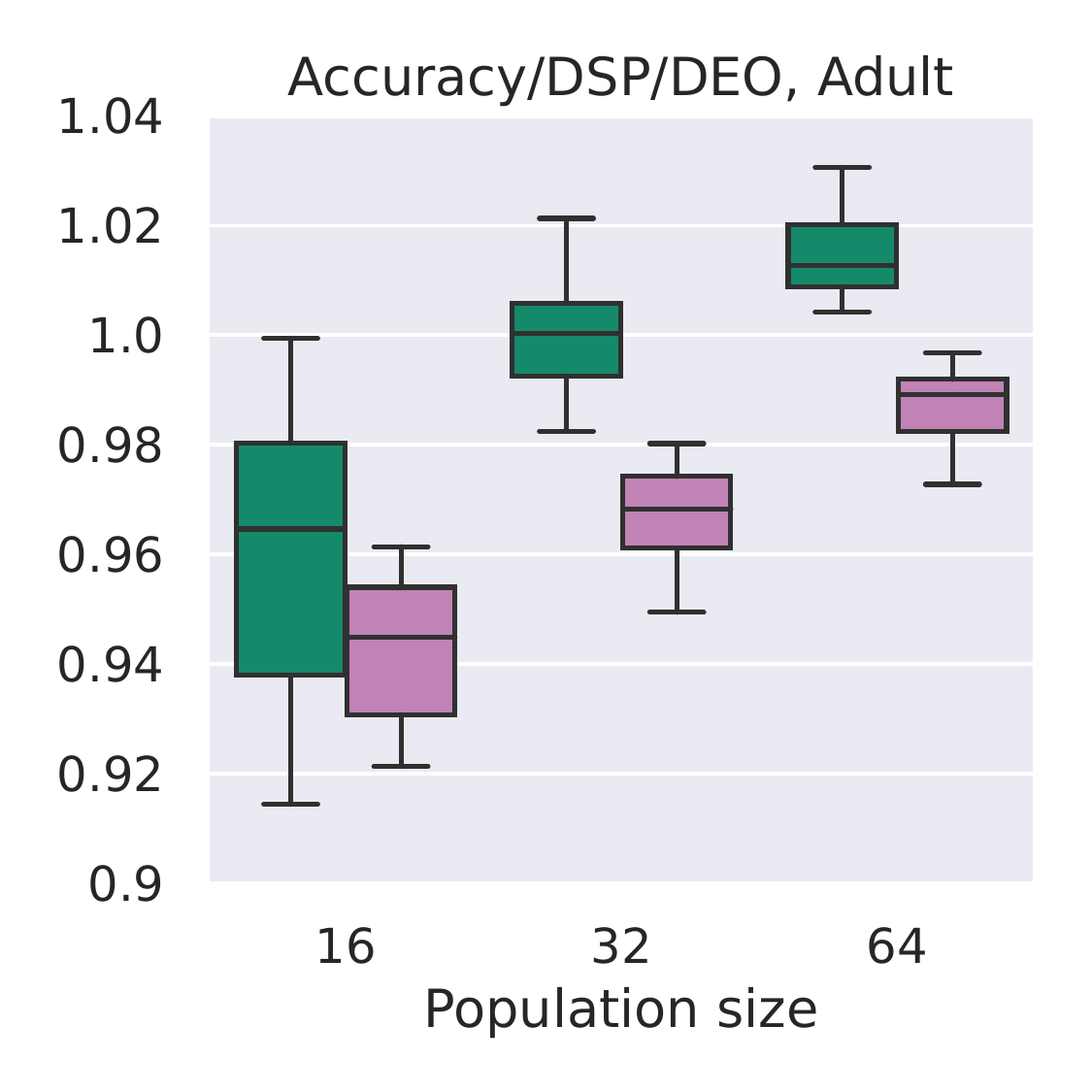}
    
    \caption{Comparison of MO-PBT with increasing population size and MO-ASHA. The time budget for MO-ASHA was adjusted accordingly. The hypervolume is normalized by the average hypervolume performance of MO-PBT with population 32.}
    \label{fig:ablation_population}
\end{figure}

We investigate whether MO-PBT benefits from increasing the population size.  The scaling experimental results are shown in Figure~\ref{fig:ablation_population}. We find that for the population sizes we considered (16, 32, 64), performance keeps improving. For reference, we also ran MO-ASHA with correspondingly increased time budgets and found that it scales similarly. We note that the performance gains when the population size is increased (from 16 to 32 and from 32 to 64) are stronger on the task with 3 objectives. This is expected behavior, as the algorithm needs more solutions to scatter along 3D approximation fronts compared to 2D in the bi-objective case.

\begin{figure}[h!]
    \centering
    \includegraphics[width=4cm]{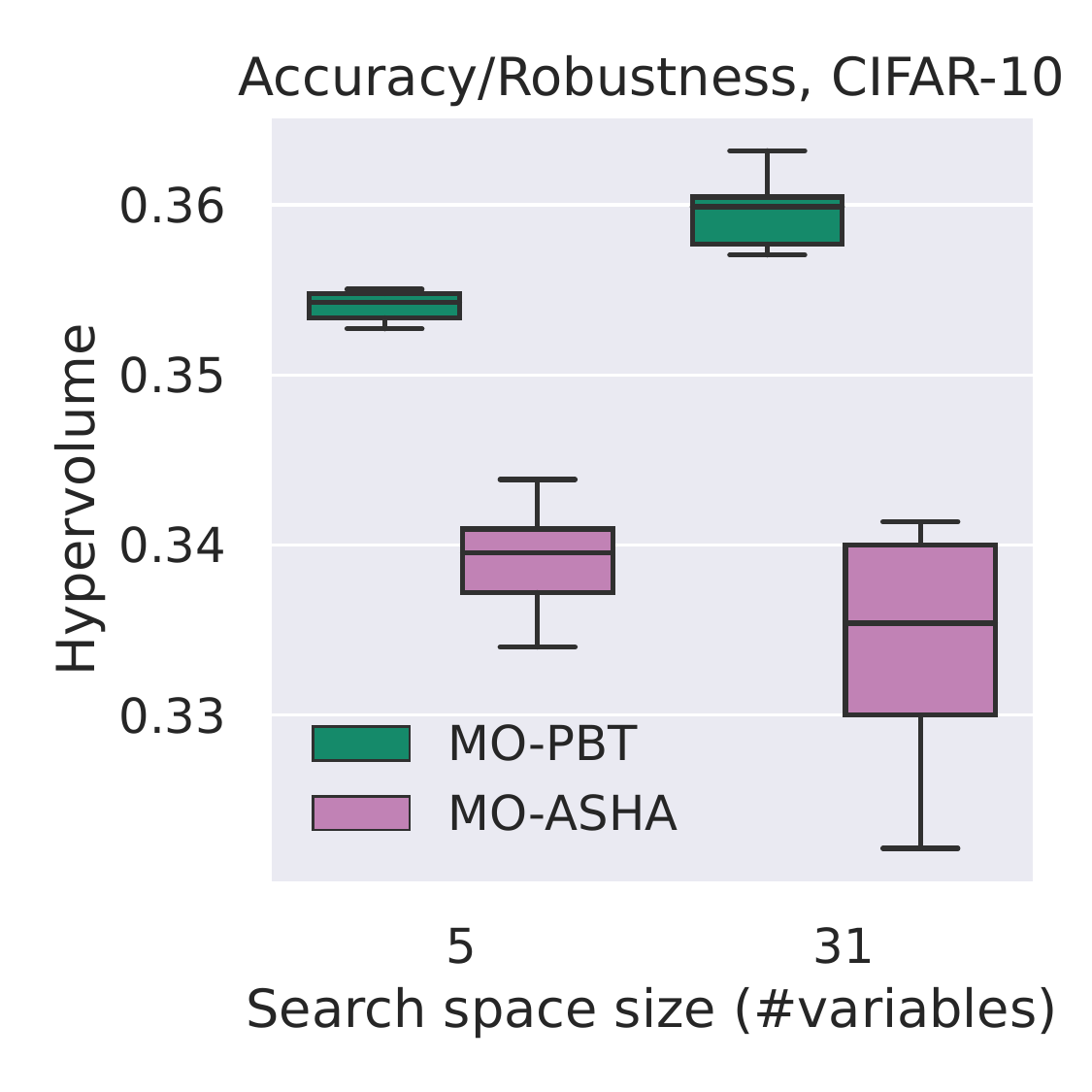}    
    \caption{Comparison of MO-PBT and MO-ASHA applied to the search spaces of different sizes.}
    \label{fig:ablation_searchspace}
\end{figure}

\subsubsection{Search space size}
In our main experiments on image datasets, we search for the two parameters  (number of augmentations and their magnitude) of the RandAugment augmentation policy, which was shown to be effective \cite{cubuk2020randaugment}. The whole search space has in that case 5 variables. To additionally study whether MO-PBT is capable of performing search efficiently in larger search spaces, we construct a substantially larger search space (comprising 31 variables) by replacing the RandAugment policy with an augmentation policy similar to the one used in \cite{ho2019population}: the magnitude and probability of each augmentation can be adjusted separately; additionally, the number of applied augmentations is searchable too. The results are shown in Figure~\ref{fig:ablation_searchspace}. We observe that in a larger search space, MO-PBT does not lose its efficiency and has even a slightly better performance.

\subsection{Further experiments to demonstrate the effectiveness of MO-PBT}
In addition to our main experiments, we compare MO-PBT to the algorithms which are not (fully) parallel. In Appendix~\ref{sec:appendix_qnehvi} we demonstrate that MO-PBT outperforms state-of-the-art BO algorithm for MO optimization: Parallel Noisy Expected Hypervolume Improvement (qNEHVI) \cite{daulton2021parallel}. Furthermore, we conduct experiments to ensure that MO-PBT is an efficient optimization algorithm even in the scenario when it is executed sequentially. In Appendix~\ref{sec:appendix_search_eff}, MO-PBT is shown to outperform common MO baselines NSGA-II \cite{deb2002fast} and ParEGO \cite{knowles2006parego} in the sequential setup.

\section{Discussion and limitations}
We have proposed MO-PBT and compared it with various baselines, including a prominent parallelizable algorithm for MO-HPO, MO-ASHA, reaching the conclusion that MO-PBT performs better. We note however that an advantage of MO-ASHA (and similar algorithms such as Hyperband) is its ability to perform not only HPO, but also architecture search, and, furthermore, joint optimization of the architecture and  hyperparameters. In PBT and MO-PBT all architectures are assumed to be identical in the population, therefore architecture search cannot be performed (without additional modifications). 

We note that quantifying the results of MO algorithms is, in, general, challenging. Many metrics have been proposed \cite{audet2021performance} and each has its own pros and cons. While hypervolume remains, arguably, the most commonly used metric, its downside is the dependence on a  user-selected reference point. Thus, while we ensured that the proposed MO-PBT outperformed alternative algorithms in terms of hypervolume, we also visually analyzed the obtained non-dominated fronts of solutions and quantified the results using a coverage metric \cite{scriven2009dynamic}. This analysis also demonstrated good performance of MO-PBT in terms of solutions diversity and density (they are well spread across different areas of the objective space).

In principle, MO-PBT (as well as the original PBT) can operate with any type of search space as long as an \emph{explore} operator is defined (moreover, the search spaces can be defined separately for each hyperparameter). 
One of the benefits of the discretized search space used in this work is (in contrast to a real-valued search space), its ability to explicitly set some values: e.g., zero value of $\lambda$ in the $\mathcal{L}_{fairness}$ turns this loss into the standard cross-entropy. Thus, more interpretable hyperparameter search results can be obtained. However, a real-valued search space can, potentially, enable performing a more fine-grained search which in some cases might be more important than the interpretability of results. 

For the main experiments of this work, we used MO-PBT with population size of 32. We additionally observed that its performance scales with increasing population size. However, population size remains a hyperparameter of MO-PBT that needs to be set by the user. Adjusting it automatically (for example, as done in EA literature \cite{harik1999parameter}) could be an interesting direction for future work.

\color{black}

\section{Conclusion}
We introduced a multi-objective version of Population Based Training: MO-PBT. We considered diverse multi-objective hyperparameter optimization tasks and  found that a multi-objective approach to ranking solutions, non-dominated sort, outperforms more simple ones such as scalarization techniques. This was demonstrated by not only better hypervolume performance, but also a better tradeoff front coverage by MO-PBT. MO-PBT was shown to outperform MO-ASHA variants (standard and Bayesian optimization based), single-objective PBT, and random search.

\section*{Acknowledgements}
The work in this paper is supported by: the Dutch Research Council (NWO) through project OCENW.GROOT.2019.015 "Optimization for and with Machine Learning (OPTIMAL)"; and project DAEDALUS funded via the Open Technology Programme of the NWO, project number 18373; part of the funding is provided by Elekta and ORTEC LogiqCare.
\clearpage

\bibliography{references}
\bibliographystyle{icml2023}

\clearpage
\appendix
\onecolumn

\section{Ablation studies} \label{sec:appendix_ablation}

\subsection {\emph{Explore} operator}
\begin{figure}[h!]
    \centering
    \includegraphics[width=0.35\textwidth]{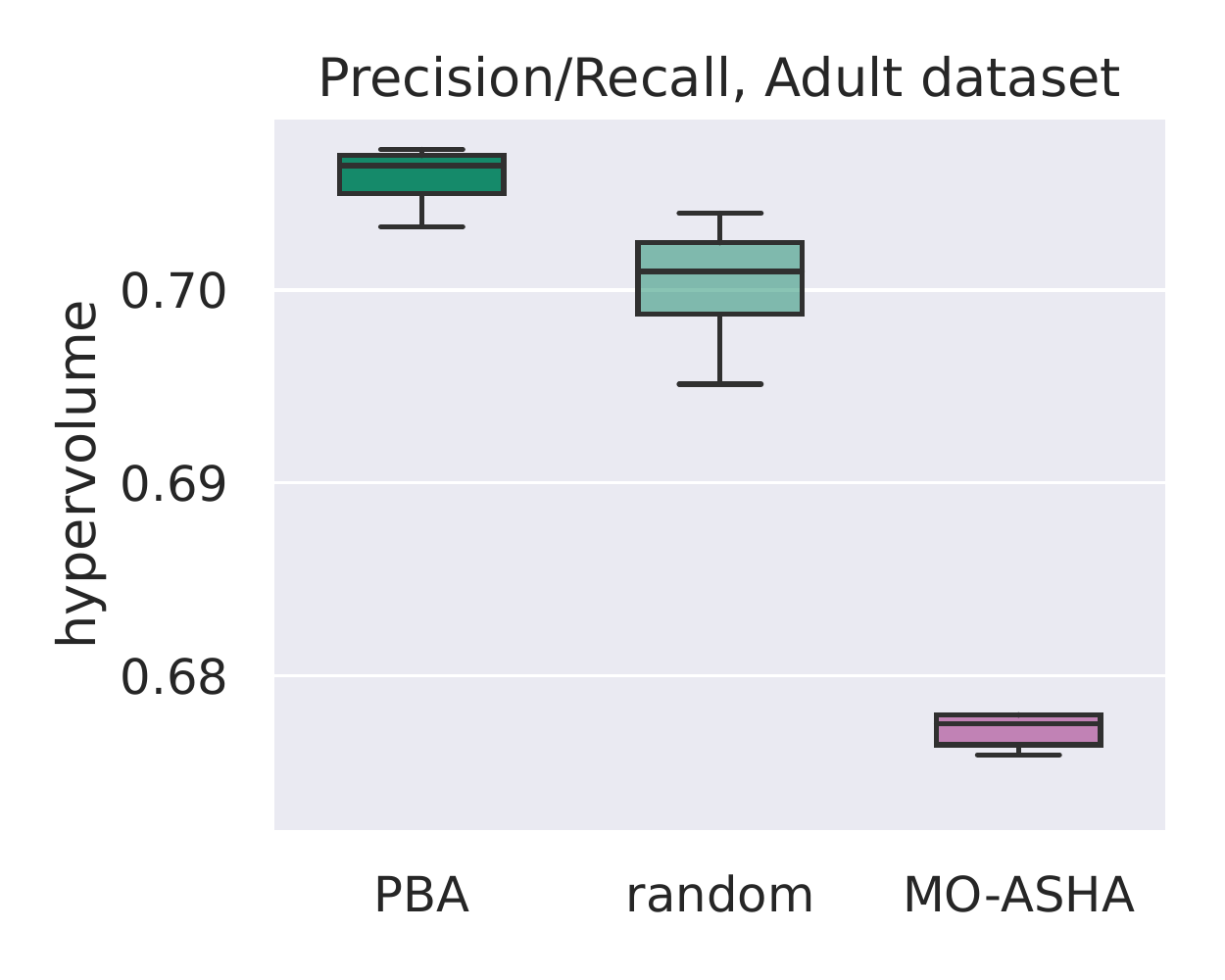}
    \hspace{1cm}
    \includegraphics[width=0.35\textwidth]{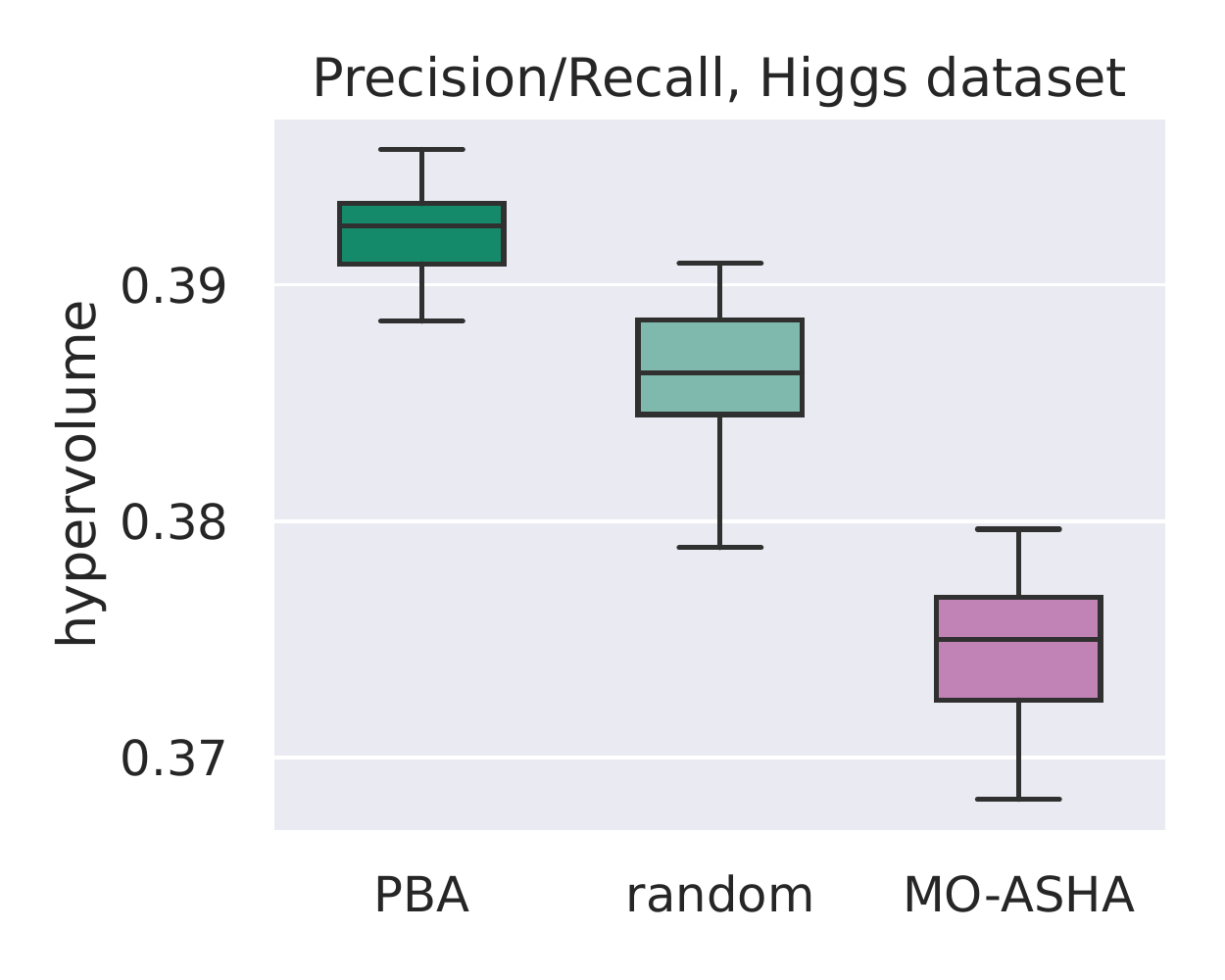}
    \caption{Comparison of MO-PBT with different \emph{explore} operators (mutations) and MO-ASHA. PBA denotes the mutation used in PBA \cite{ho2019population} (local mutation), random denotes simple and random mutation.}
    \label{fig:ablation_explore}
\end{figure}

We compare the used local \emph{explore} operator from PBA \cite{ho2019population} with random mutation from classic EAs: for each of $P$ variables, the new value is randomly sampled from the corresponding search domain $\mathcal{H^P}$ with probability $\frac{1}{P}$. As demonstrated in Figure~\ref{fig:ablation_explore}, the local mutation performs slightly better.
%\vspace{-0.5cm}

\subsection {\emph{Exploit} operator}
\begin{figure}[h!]
    \centering
    \includegraphics[width=0.35\textwidth]{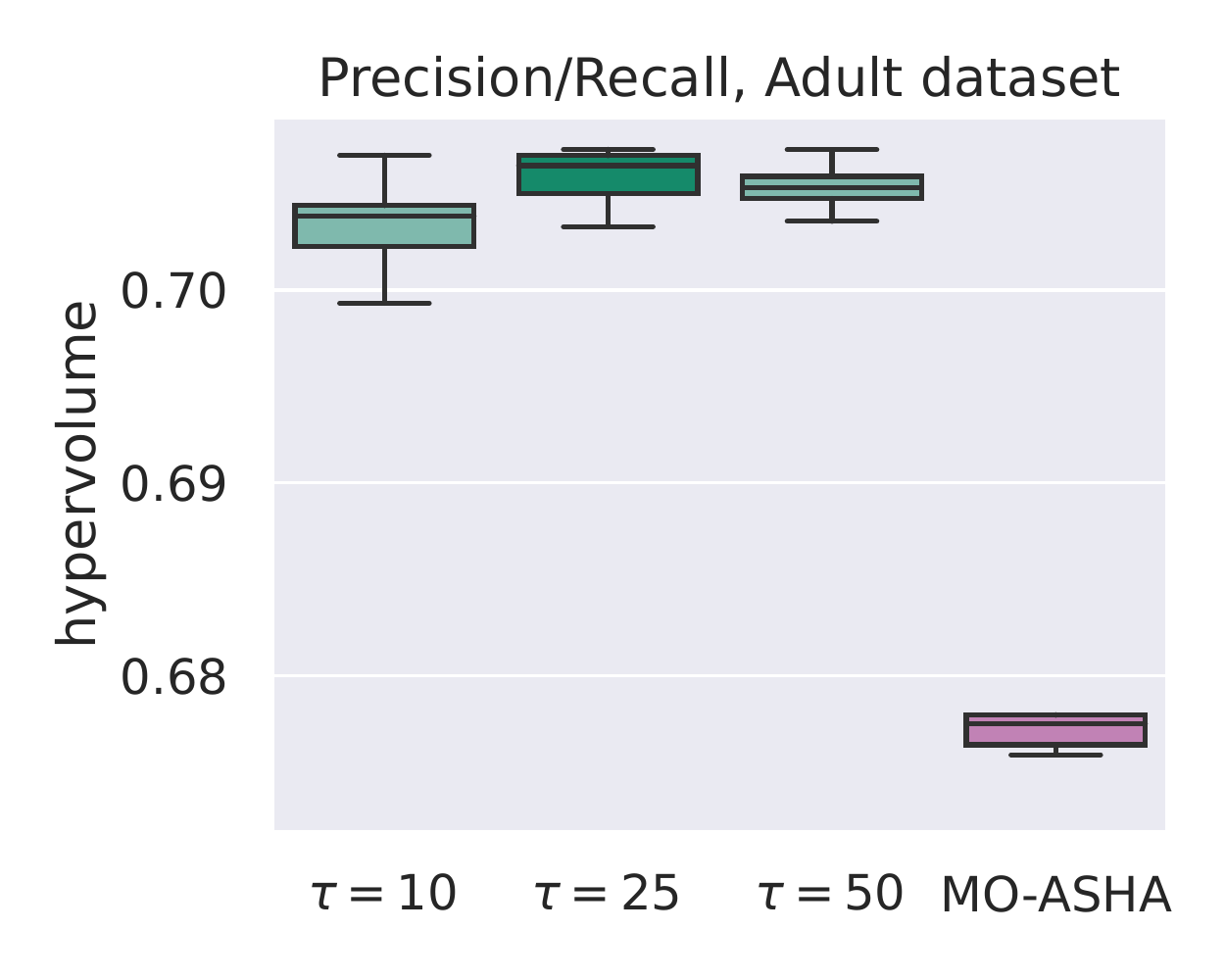}
    \hspace{1cm}
    \includegraphics[width=0.35\textwidth]{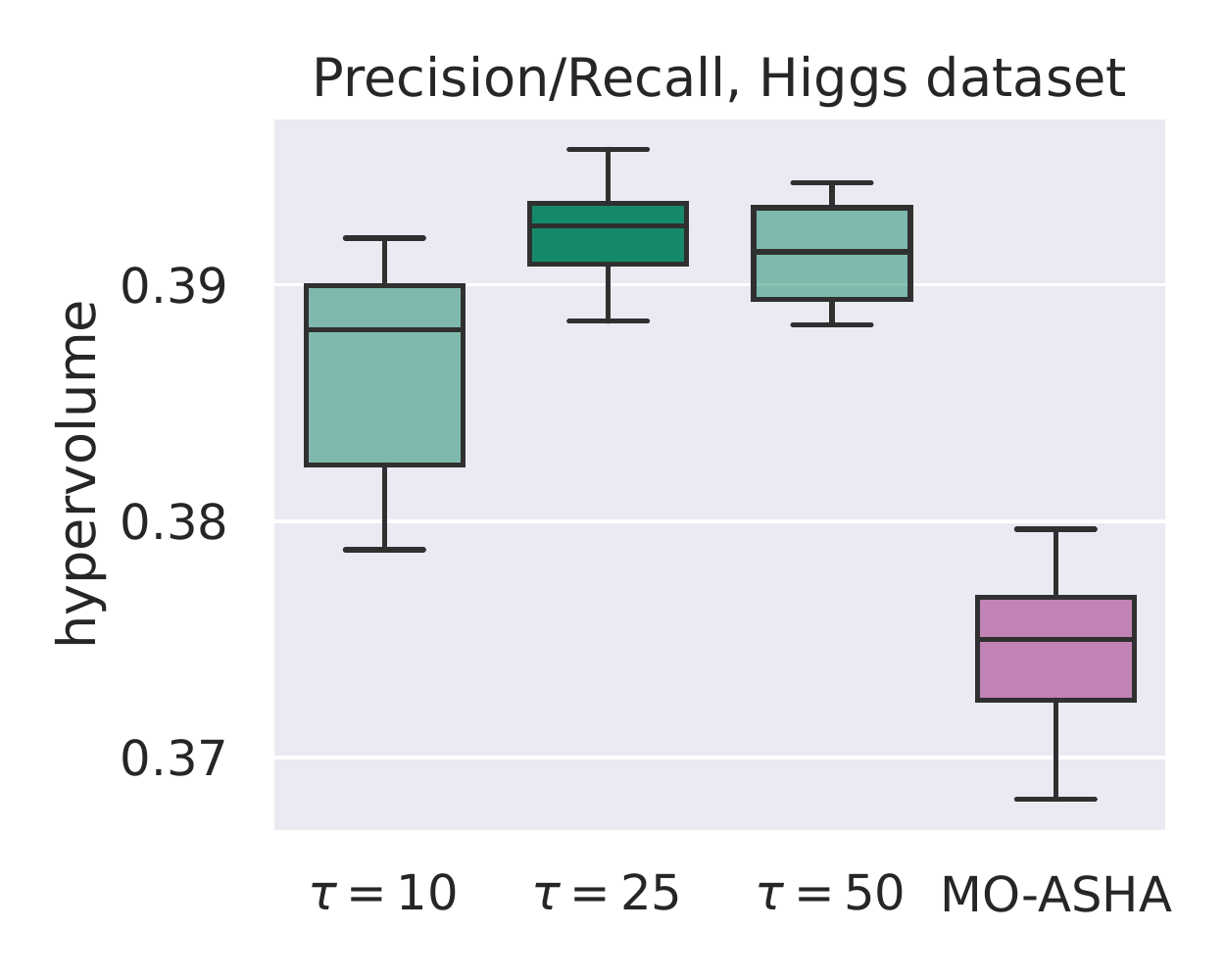}
    \caption{Comparison of MO-PBT with different truncation selection values in the \emph{exploit} operator (parameter $\tau$ in Algorithm~\ref{alg:mopbt_exploit}) and MO-ASHA.}
    \label{fig:ablation_exploit}
\end{figure}

In \cite{jaderberg2017population} it was shown that using truncation selection as the \emph{exploit} operator works best out of considered options. Here, we study whether the truncation selection parameter  ($\tau$ in Algorithm~\ref{alg:mopbt_exploit}, larger value means more solutions are replaced) has a significant impact on performance. As demonstrated in Figure~\ref{fig:ablation_exploit}, the value of $\tau=25$ used in \cite{jaderberg2017population} performs slightly better than values $10$ and $50$.

%\vspace{-0.5cm}
\vspace{3cm}

\subsection {Ranking criterion} \label{sec:appendix_ranking}
We study whether greedy scattered subset selection used for solutions ranking (together with non-dominated sort as described in Section~\ref{sec:MOPBT}) performs better than the crowding distance (the default ranking criterion in NSGA-II).

\begin{figure}[h!]
    \centering
    \includegraphics[width=0.35\textwidth]{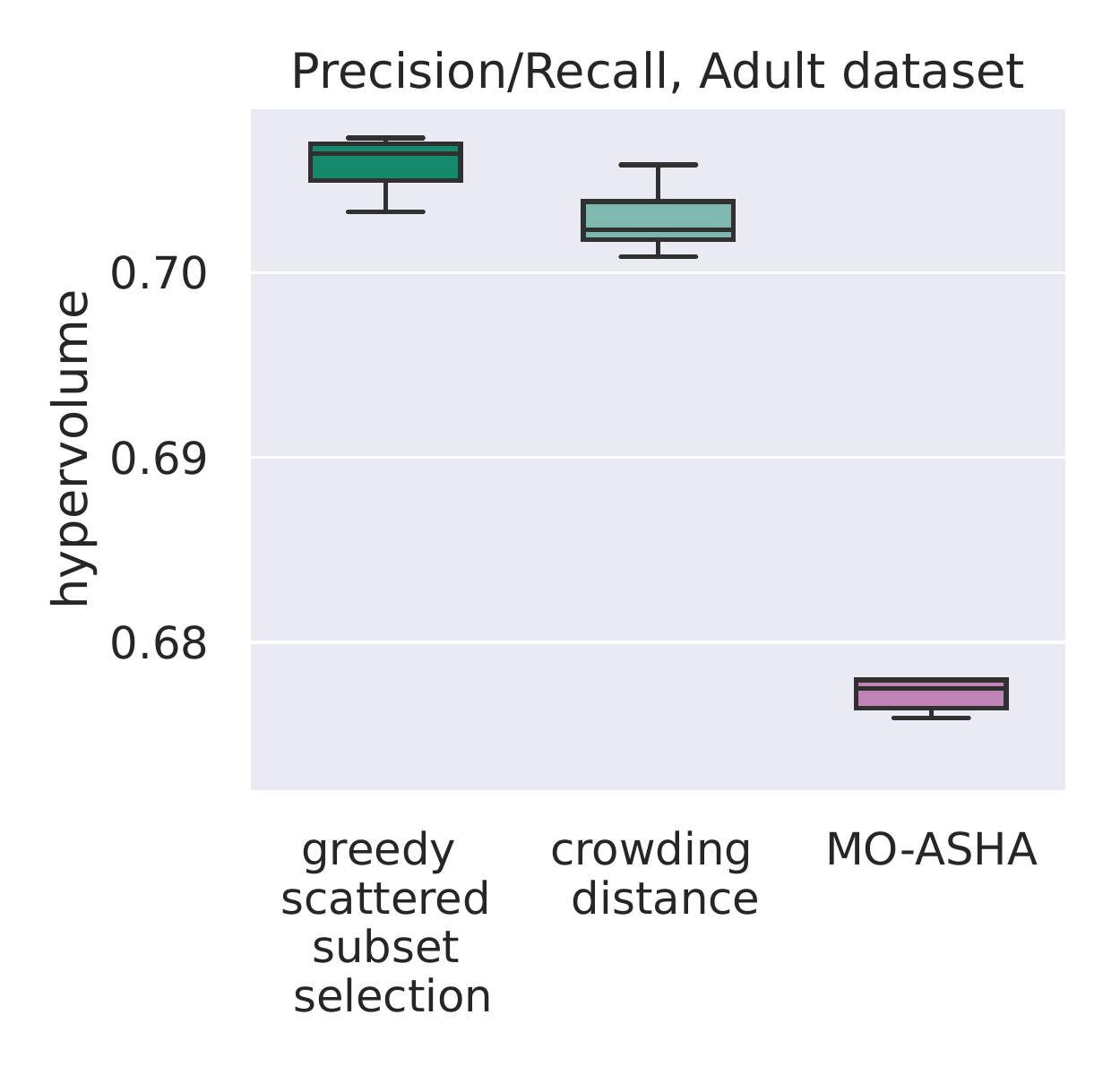}
    \hspace{1cm}
    \includegraphics[width=0.35\textwidth]{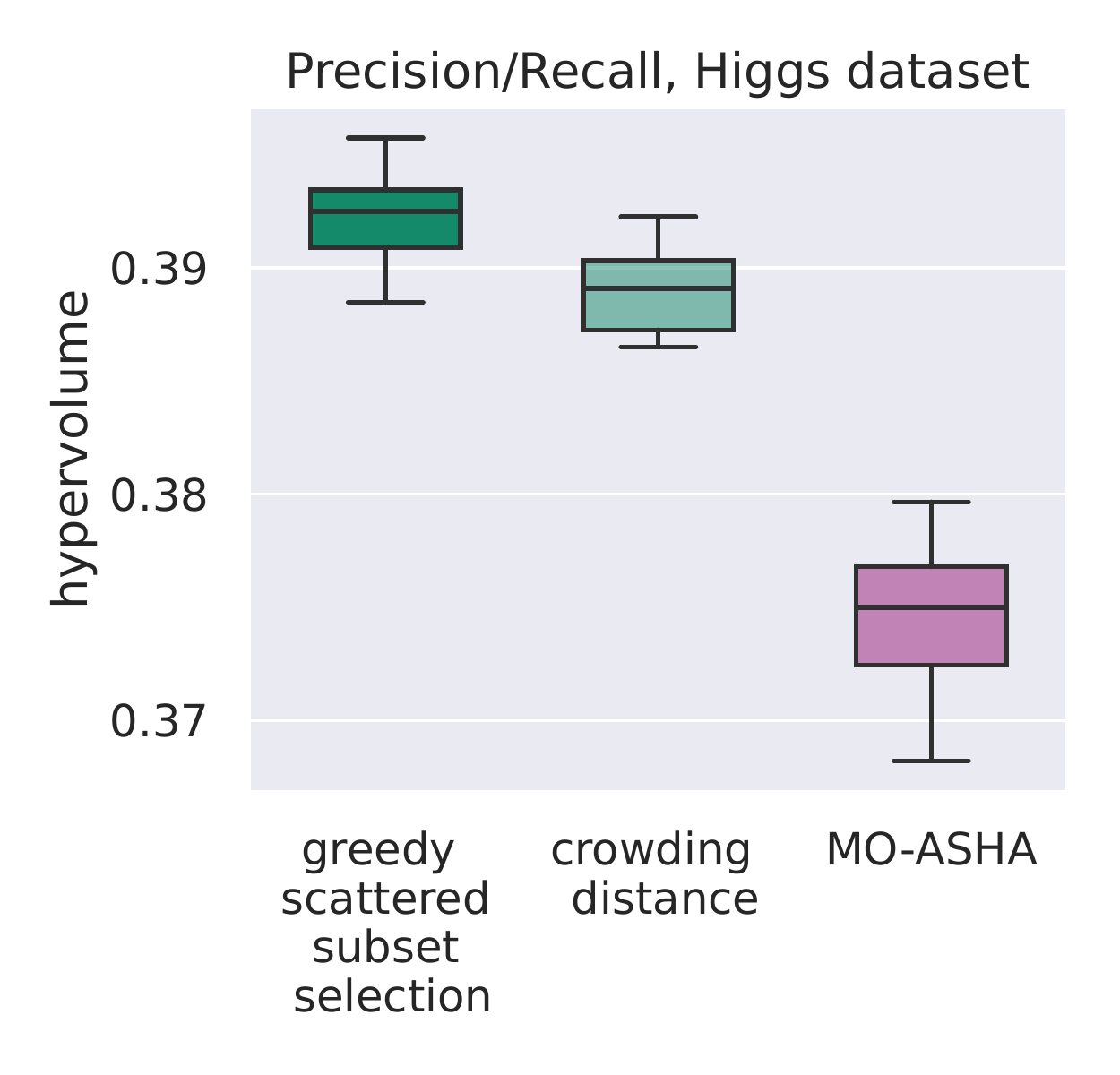}
    \caption{Comparison of MO-PBT with different criteria of ranking solutions and MO-ASHA.}
    \label{fig:ablation_ranking}
\end{figure}

\subsection {Ablation studies conclusions} 
We can conclude that MO-PBT demonstrates robustness to its operators' design: while the default designs of \emph{explore} and \emph{exploit} operators also used in PBT for augmentations policy search \cite{jaderberg2017population} perform slightly better than considered alternatives, MO-PBT outperforms MO-ASHA for all considered operators. A similar result is observed for the ranking criterion of the solutions: MO-PBT with the greedy scattered subset selection performs better than MO-PBT with the crowding distance, but the difference is not substantial.

\section{Implementation and experimental details} \label{sec:appendix_implementation}

We implemented all algorithms using Ray Tune library \cite{liaw2018tune}. Network training was performed using PyTorch \cite{NEURIPS2019_9015}. We used machines with 3 Nvidia A5000 GPUs and trained 4 networks on each GPU simultaneously, i.e., 12 networks could be trained in parallel. The total utilized number of GPU hours to reproduce all of our experiments (8 algorithms and 10 (5) seeds per algorithm on tabular (image) datasets) is 900 GPU hours for the experiments on tabular datasets and 10,000 GPU hours for the experiments on image datasets. One run of MO-PBT took less than 2 wall-clock hours on tabular datasets and less than 25 wall-clock hours on image datasets.

The training procedure for FT-Transformer (in Precision/Recall tasks and Accuracy/Fairness on the Adult dataset) is adapted from \cite{gorishniy2021revisiting}: AdamW \cite{loshchilov2017decoupled} with learning rate $10^{-5}$ (no learning rate scheduler is used) but without early stopping. Batch size is set to 512. The training is performed for 100 epochs. On the image datasets, we use standard  for WideResNet  (used, for instance, in \cite{cubuk2020randaugment}) cosine learning rate schedule with an initial learning rate 0.1 for SGD with momentum value of 0.9, and batch size 128. The training is performed for 100 epochs.
\clearpage
\section{Generalization results} \label{sec:appendix_generalization}

\begin{figure*}[h!]
    \centering
    \includegraphics[width=4cm]{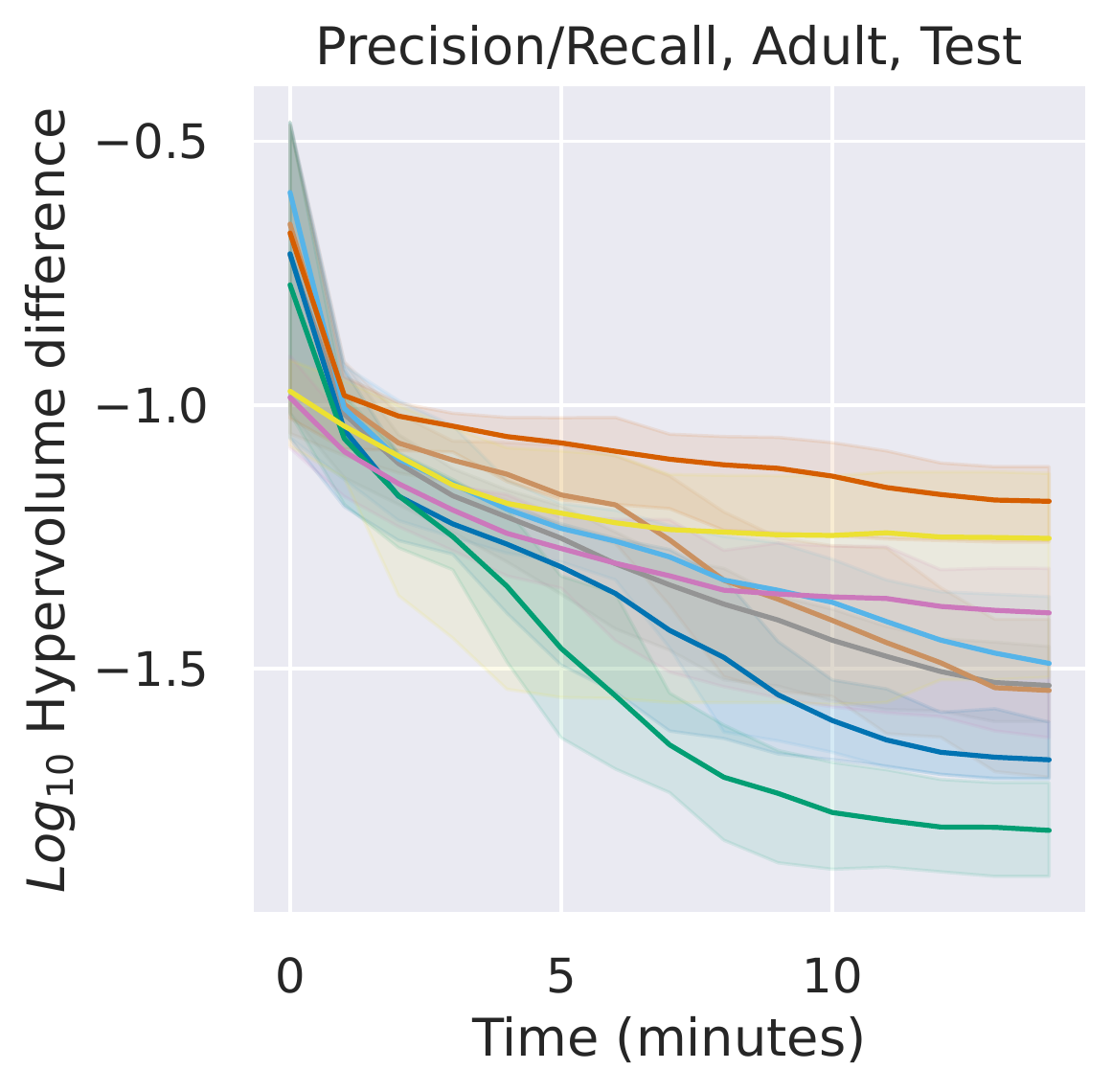}
    \includegraphics[width=4cm]{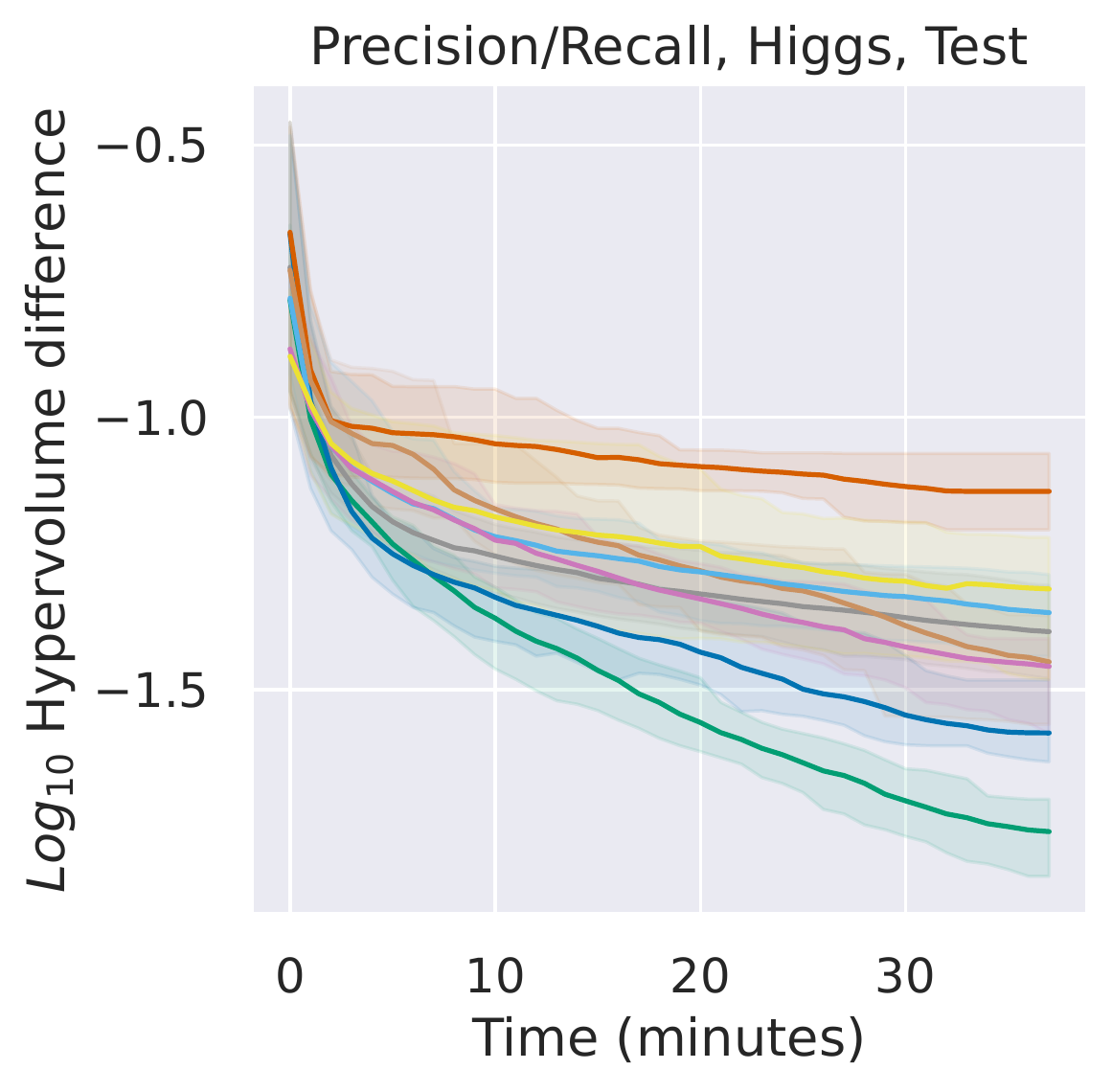}
    \includegraphics[width=4.2cm]{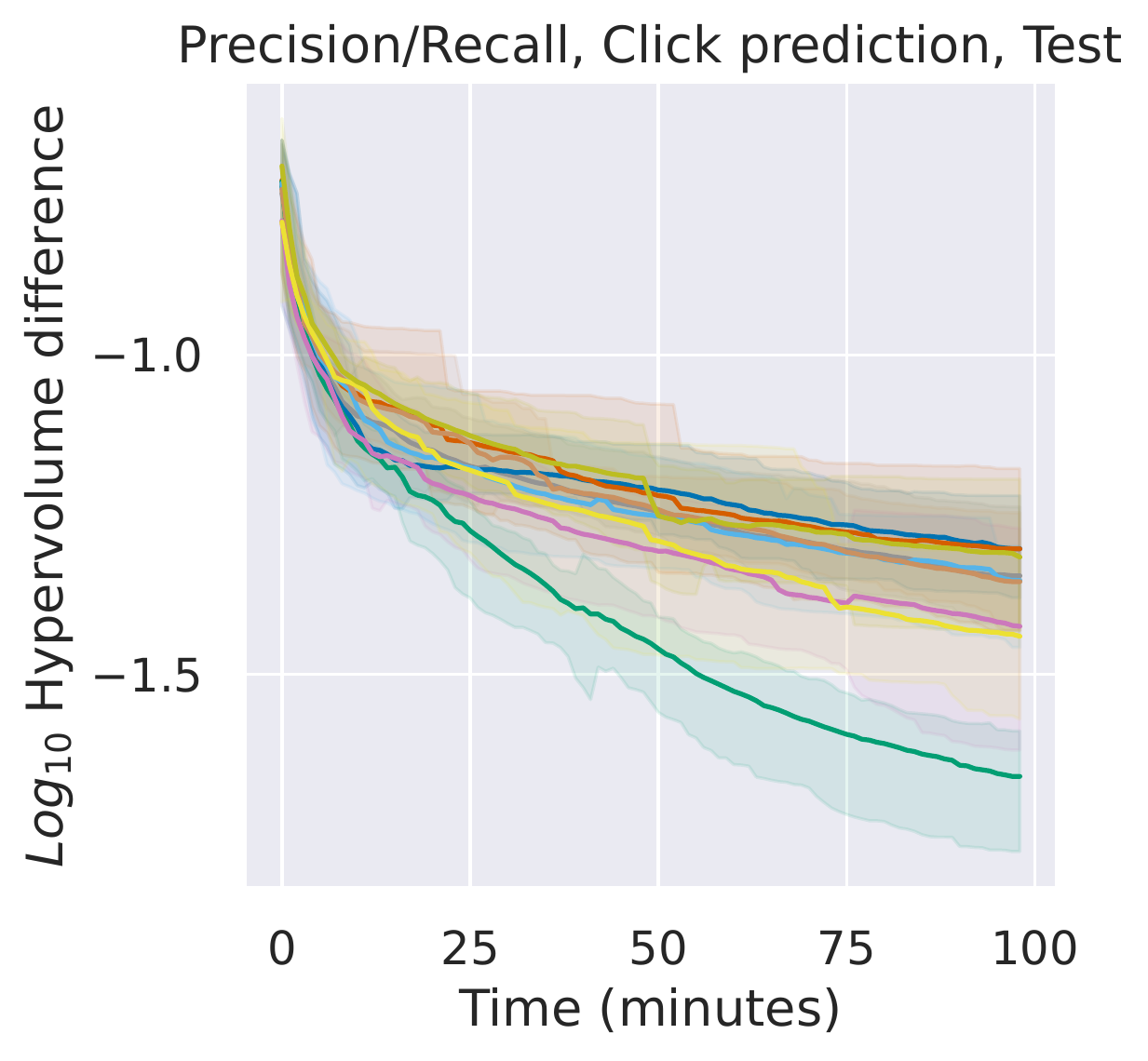}
 \includegraphics[width=0.65\textwidth]
    {images_pdfs/legend1.pdf}
    \vspace{0.5cm}
    
    \includegraphics[width=4cm]{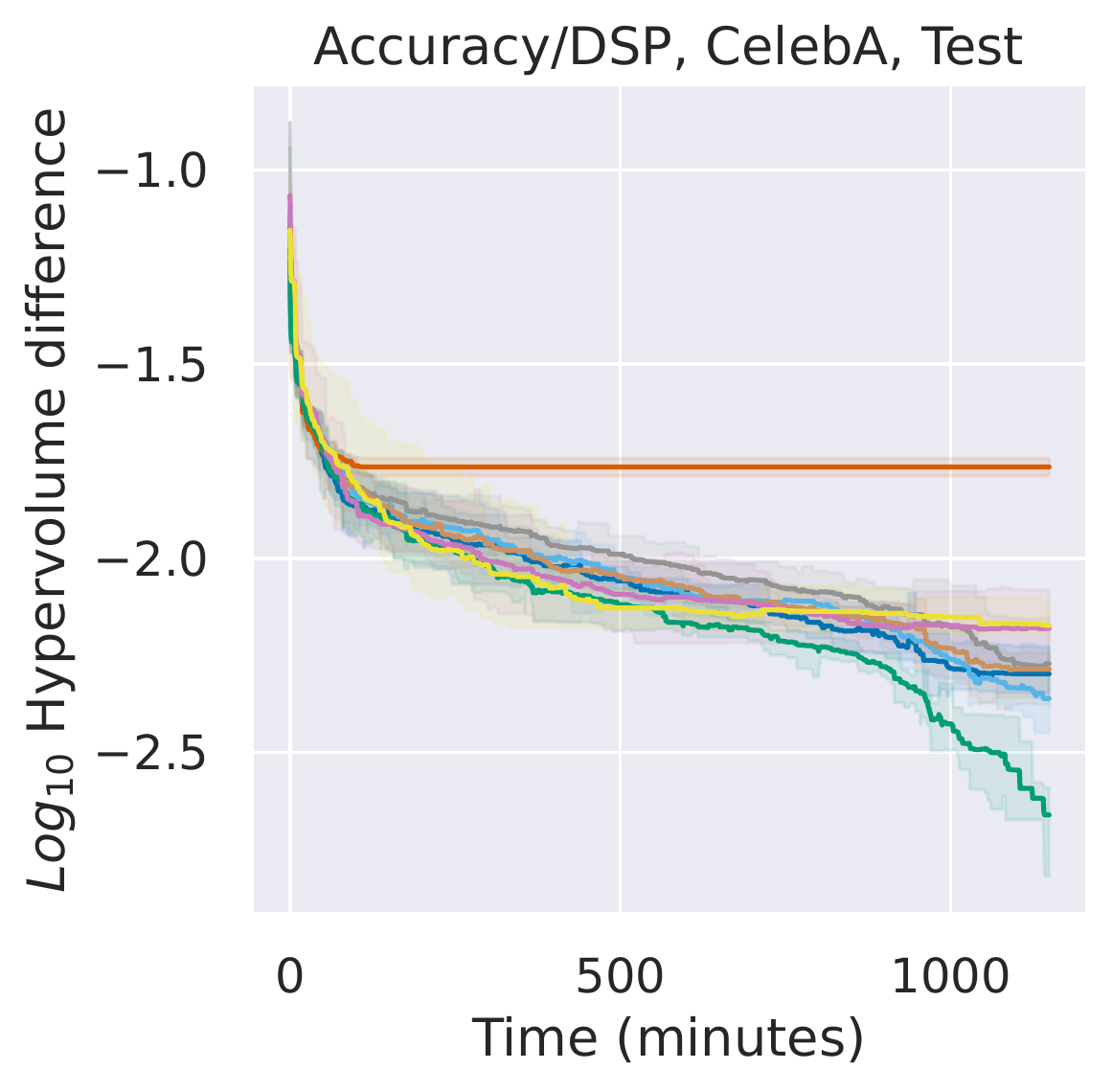}
    \includegraphics[width=4cm]{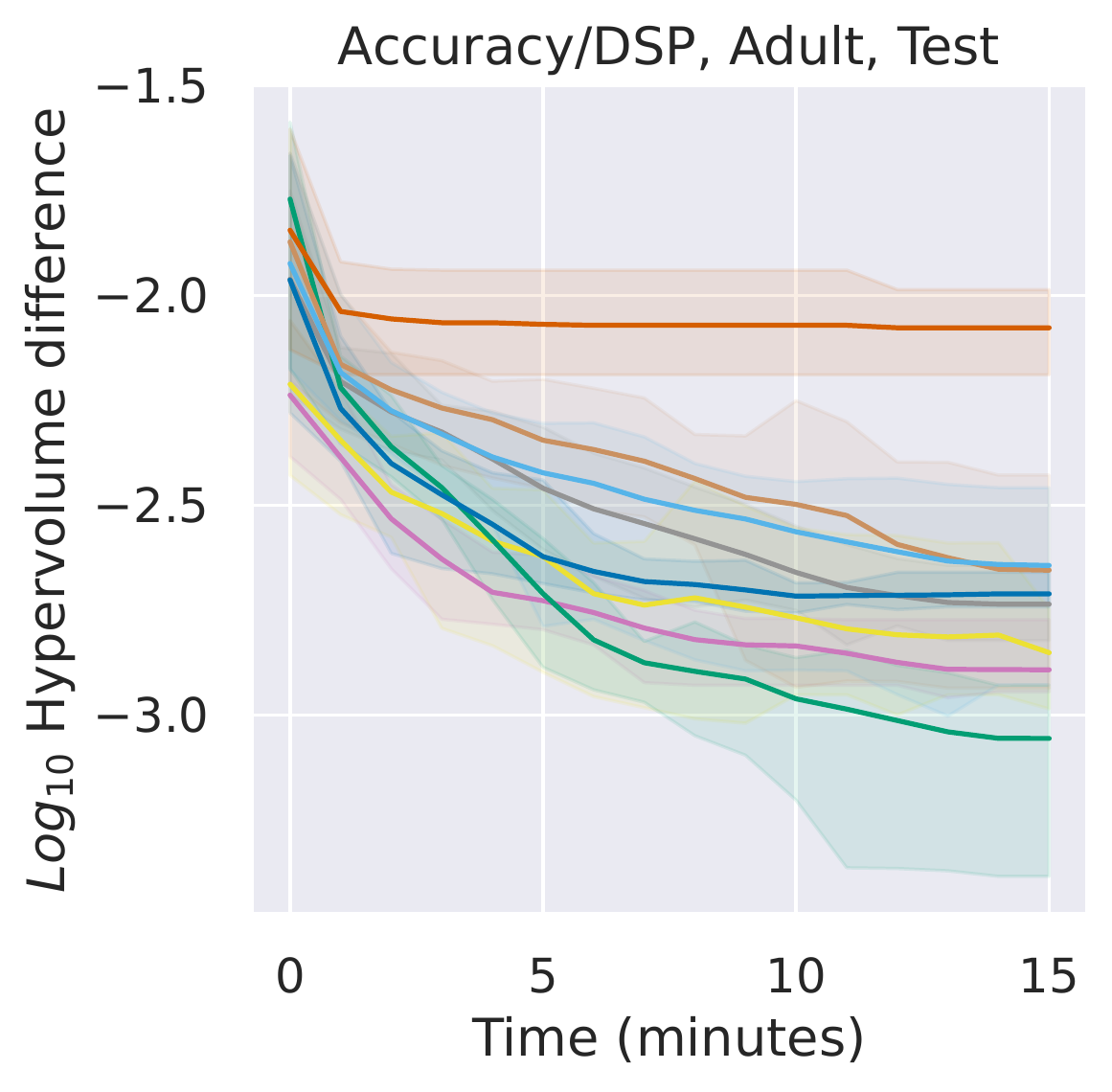}
\includegraphics[width=4cm]{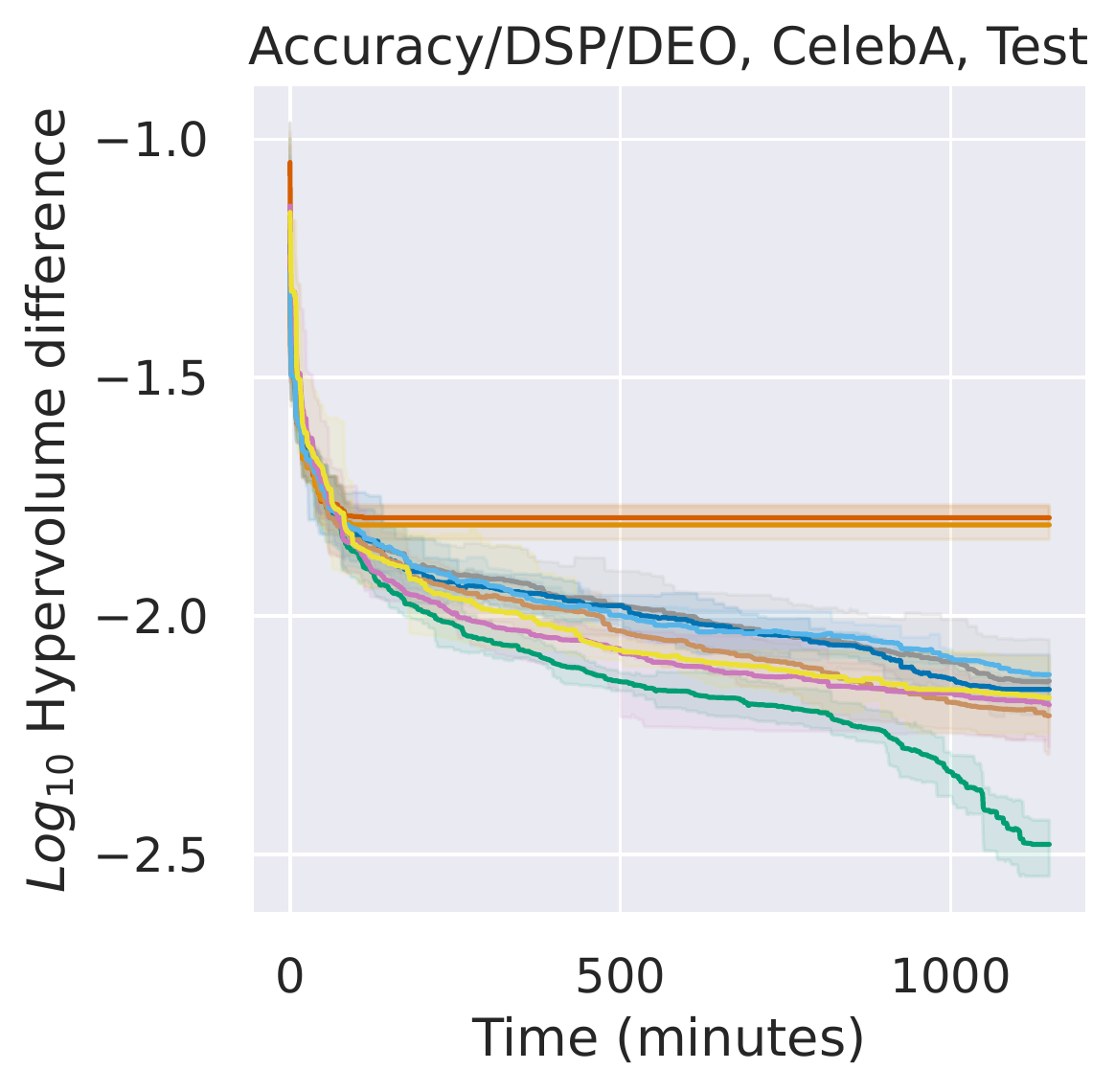}
    \includegraphics[width=4cm]{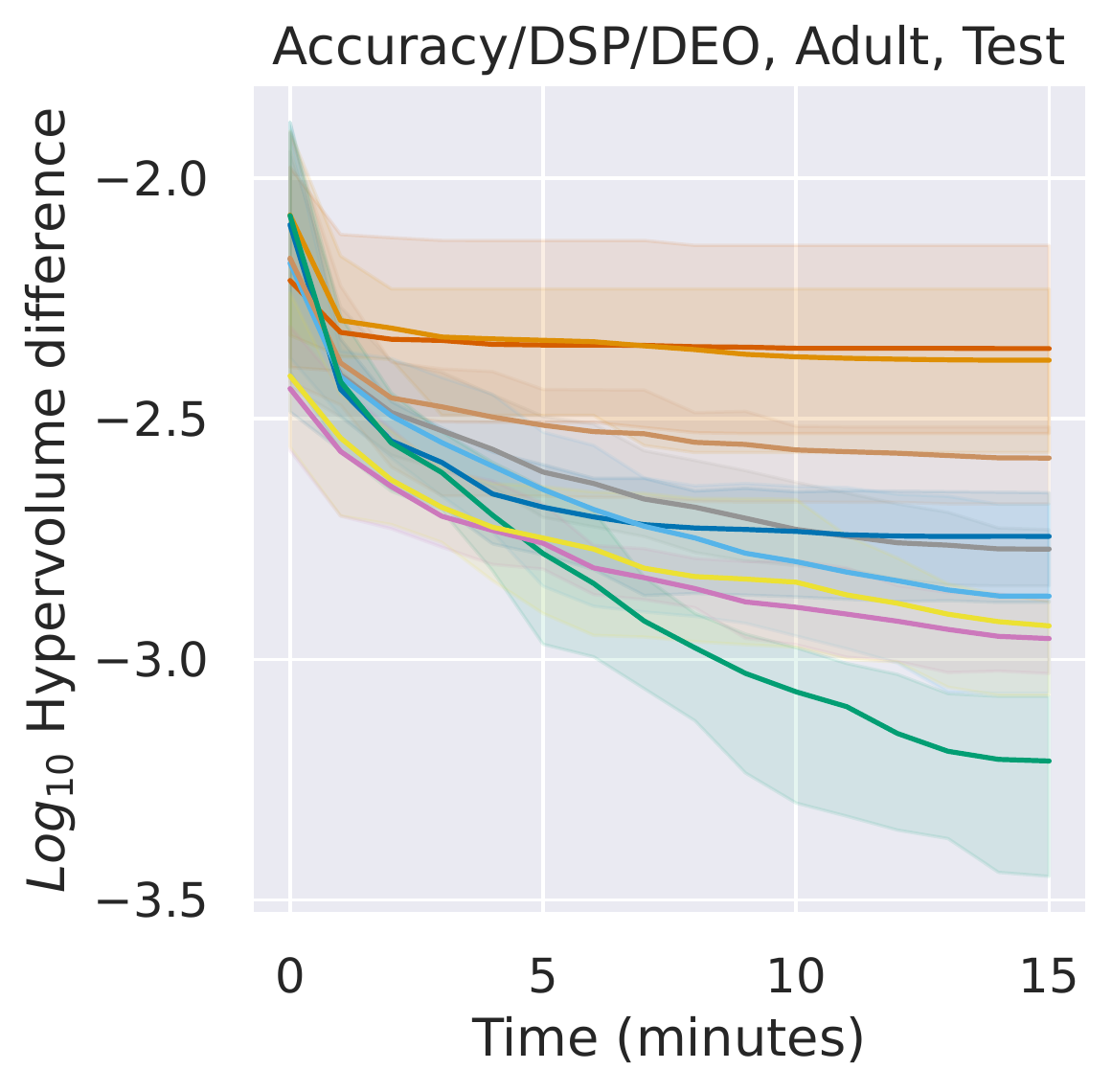}

    \includegraphics[width=0.65\textwidth]
    {images_pdfs/legend4.pdf}
    \vspace{0.5cm}
    
    \includegraphics[width=4cm]{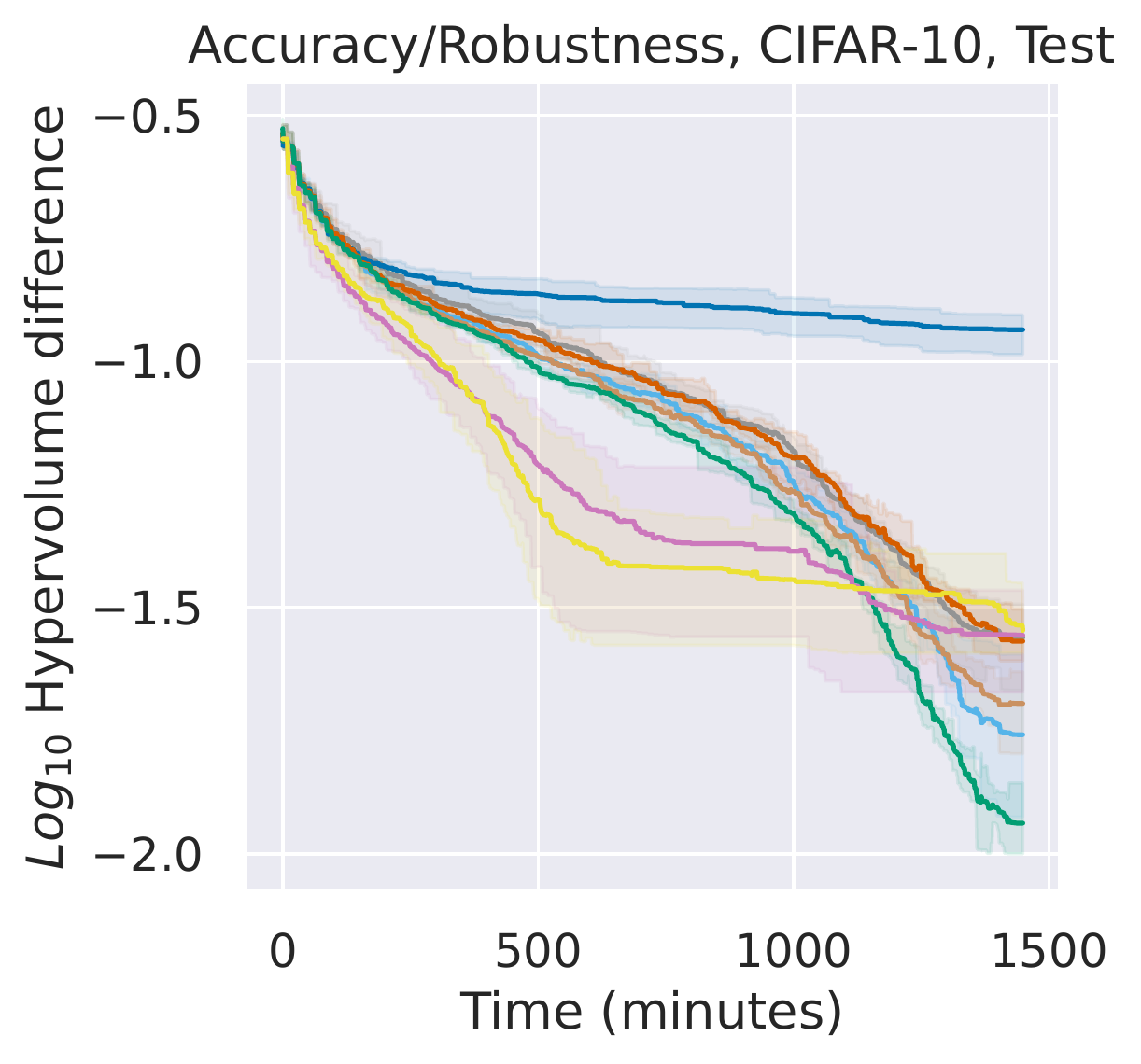}
    \includegraphics[width=4cm]{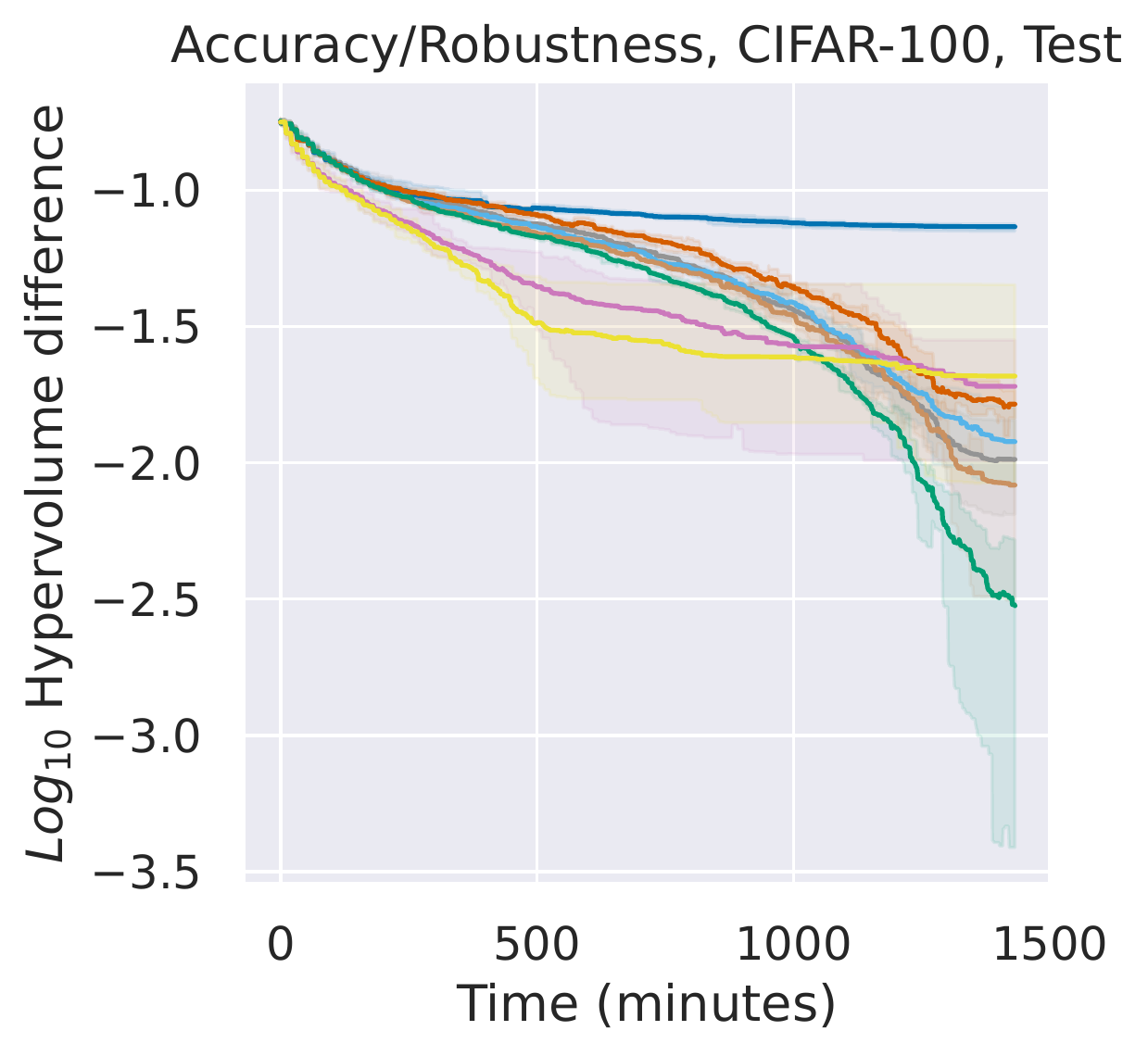}
    \includegraphics[width=0.65\textwidth]{images_pdfs/legend3.pdf}
    \caption{Generalization results (on the \emph{test} data subset) on all tasks: Precision/Recall (top row), Accuracy/Fairness (middle row), Accuracy/Robustness(bottom row).}
    \label{fig:results_test}
\end{figure*}

In this section, we inspect how the performance of the algorithms transfers from the validation set to the test set. The solutions on the trade-off front were determined based on the validation metrics. Then, these selected models are evaluated on the test set. Some of them may perform worse than expected, and not be a part of the trade-off front anymore because they are dominated by other solutions on it. The hypervolume of the remaining solutions is computed and visualized in  Figure~\ref{fig:results_test} and listed in Table~\ref{tab:hv_results_test}.

MO-PBT outperforms the baselines, although on some tasks (Precision/Recall, Accuracy/DSP on the Adult dataset) the difference in final performance becomes smaller. 

\clearpage

\section{Comparison to parallel BO algorithms} \label{sec:appendix_qnehvi}

\begin{figure}[h!]
    \centering
    \includegraphics[width=0.2\textwidth]{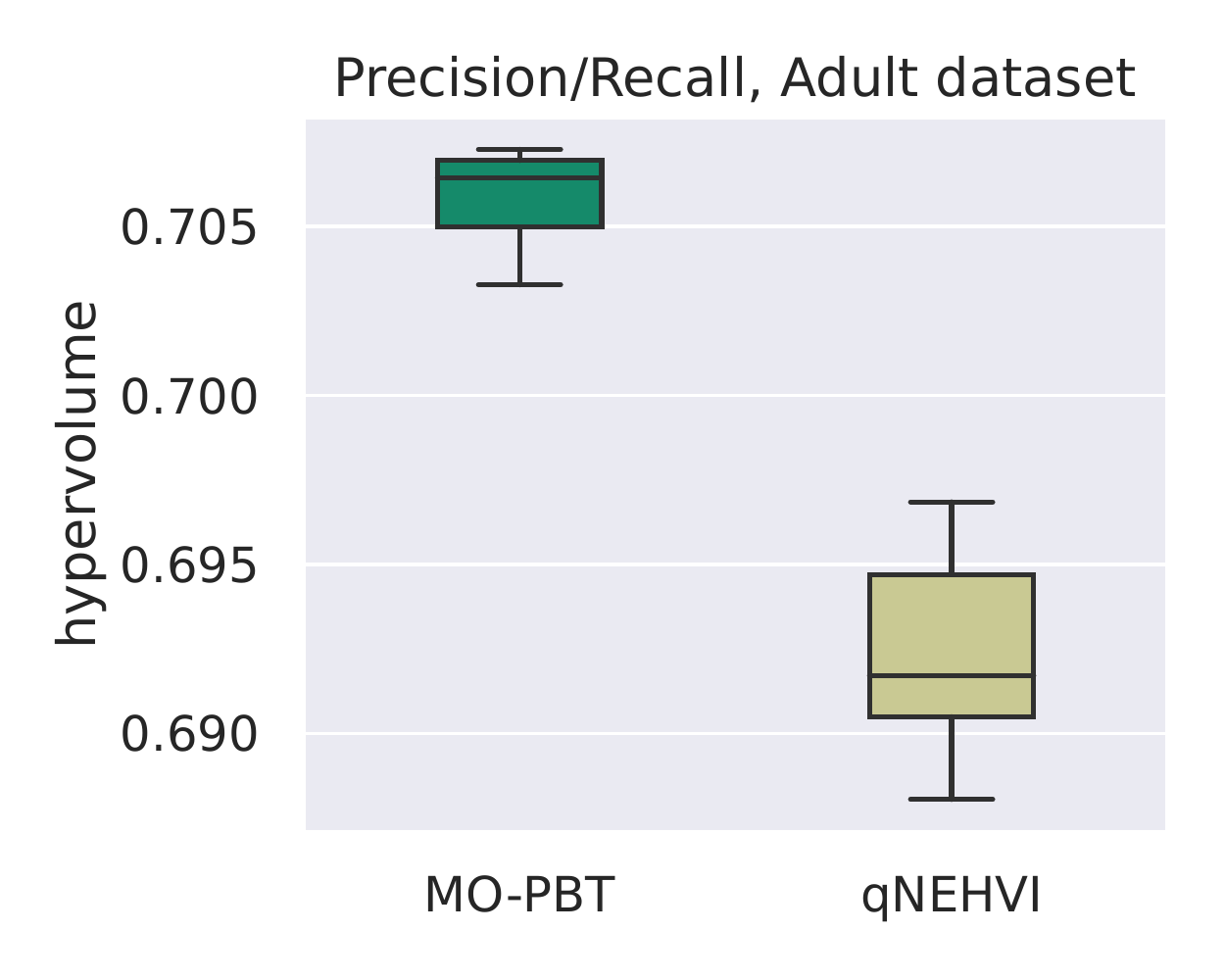}%
    \hspace{0.1cm}
    \includegraphics[width=0.2\textwidth]{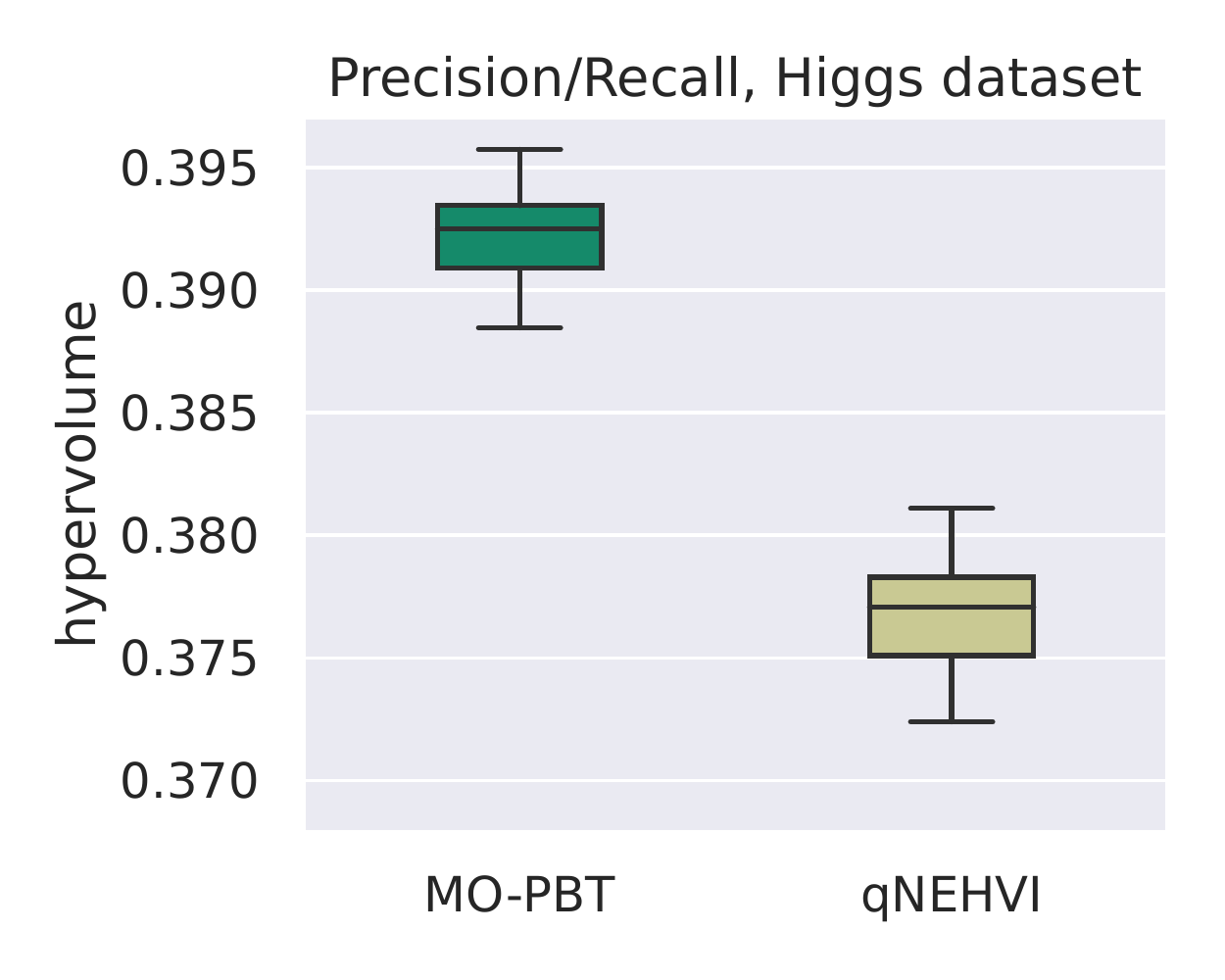}%
    \hspace{0.1cm}
    \includegraphics[width=0.2\textwidth]{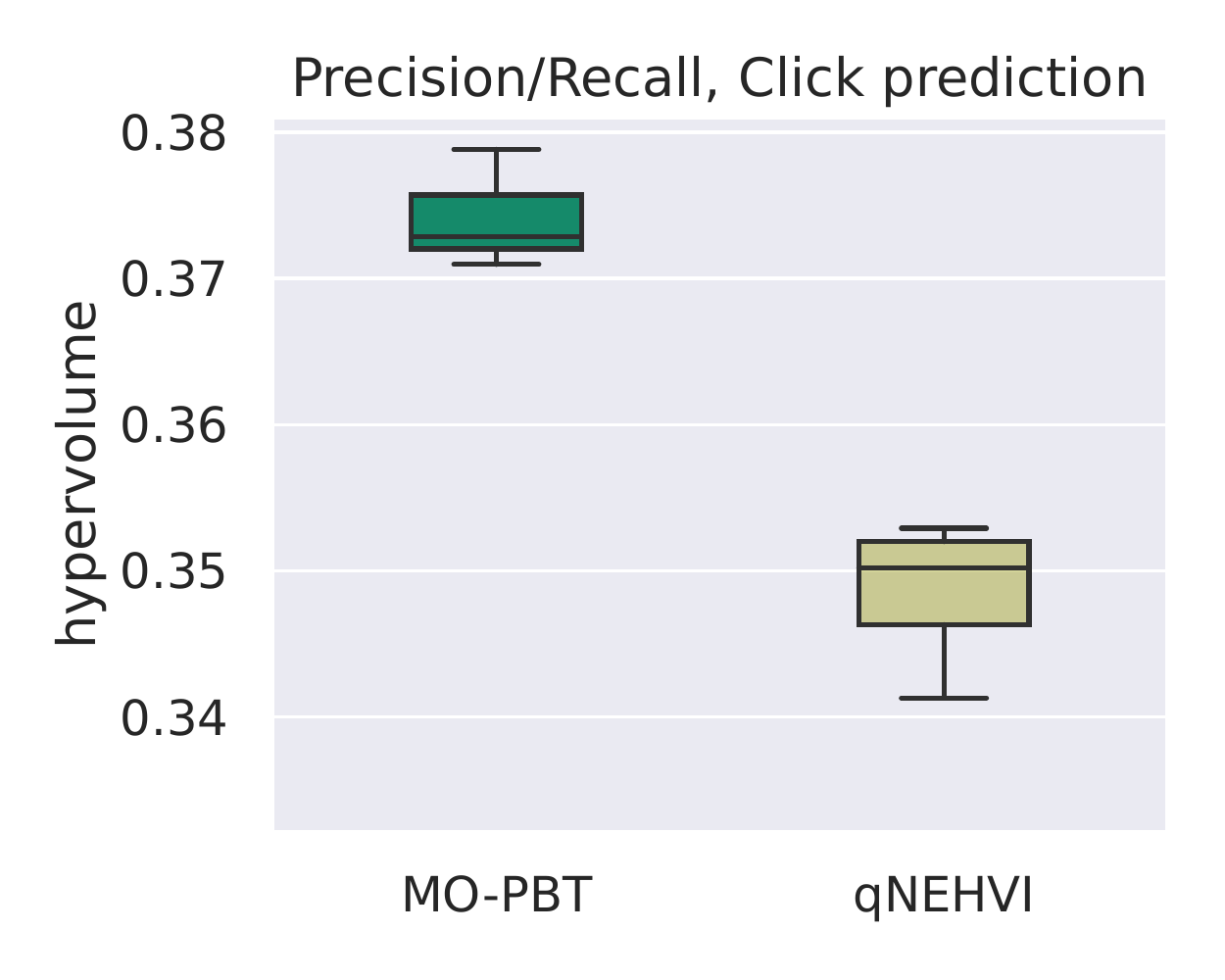} \newline
    \hspace{0.1cm}
    \includegraphics[width=0.2\textwidth]{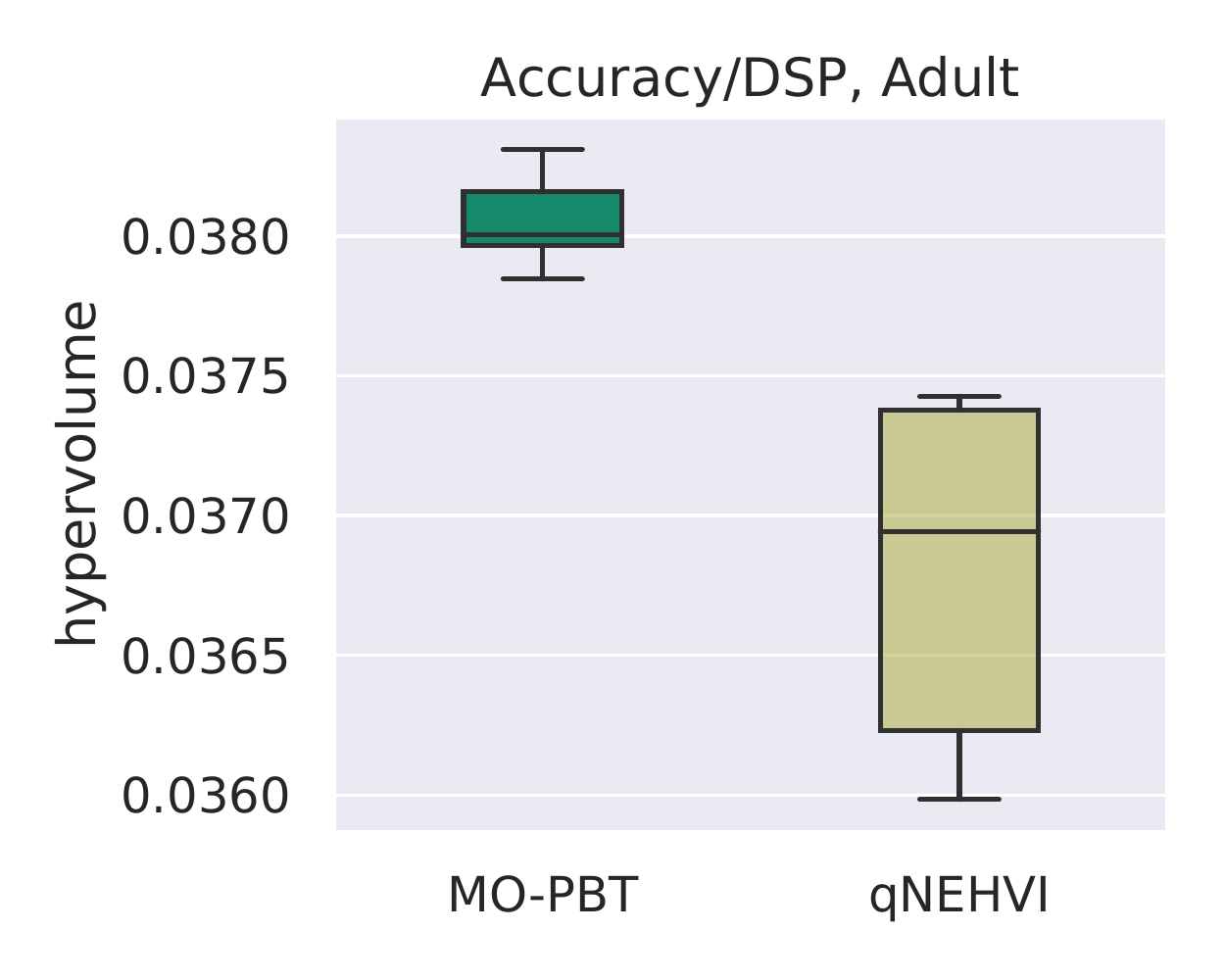}%
    \hspace{0.1cm}
    \includegraphics[width=0.2\textwidth]{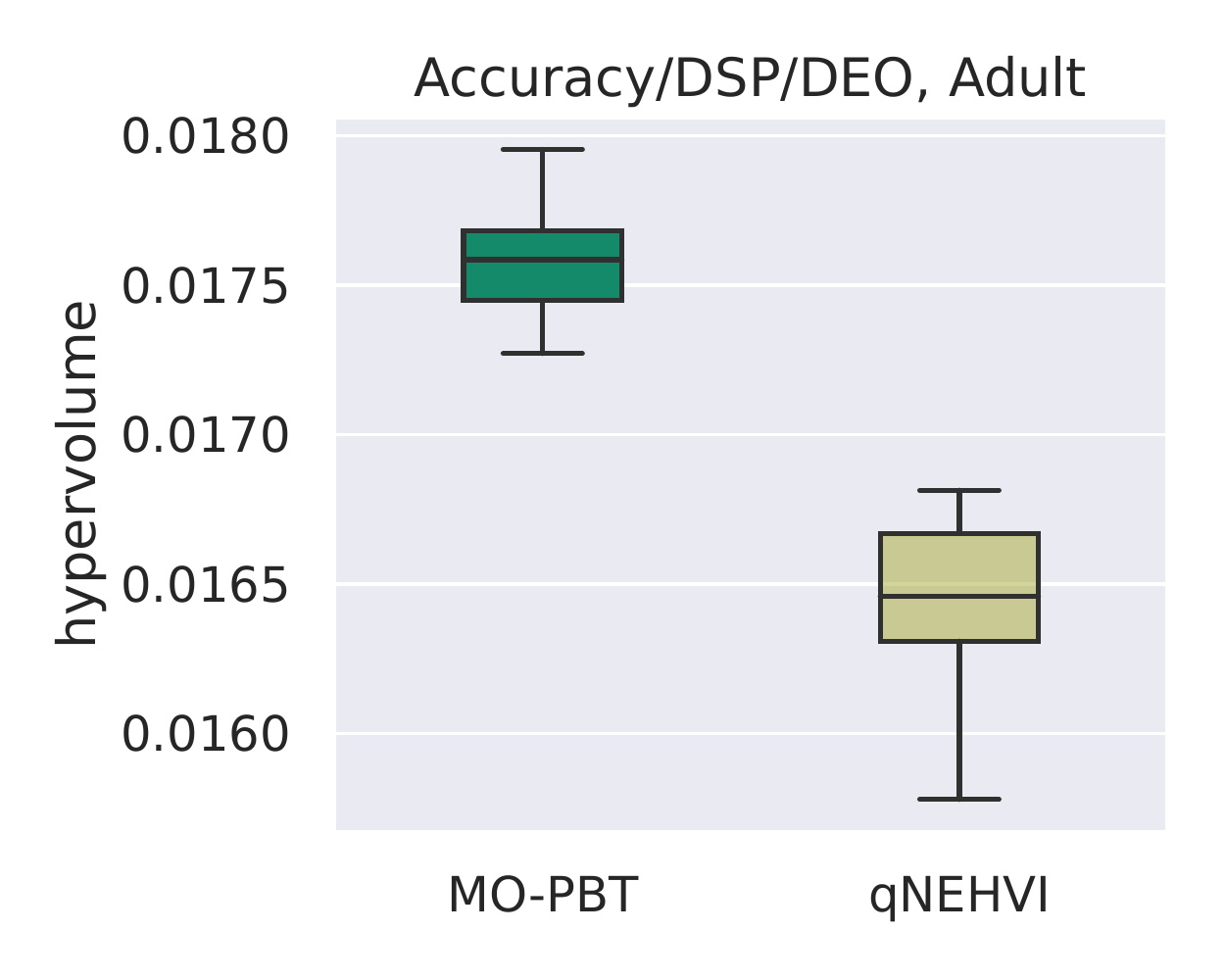}%
    \hspace{0.1cm}
    \includegraphics[width=0.2\textwidth]{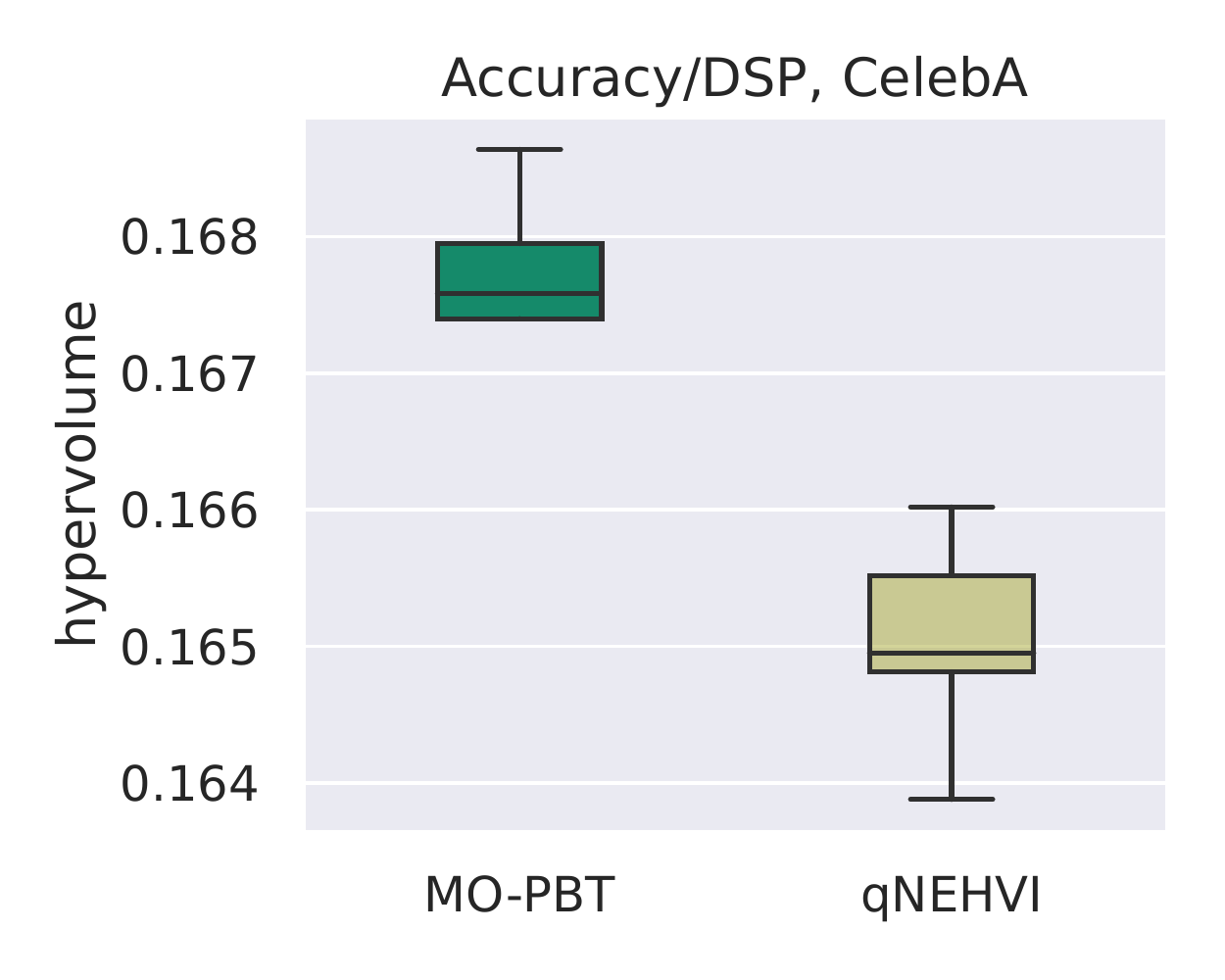}%
    \hspace{0.1cm}
    \includegraphics[width=0.2\textwidth]{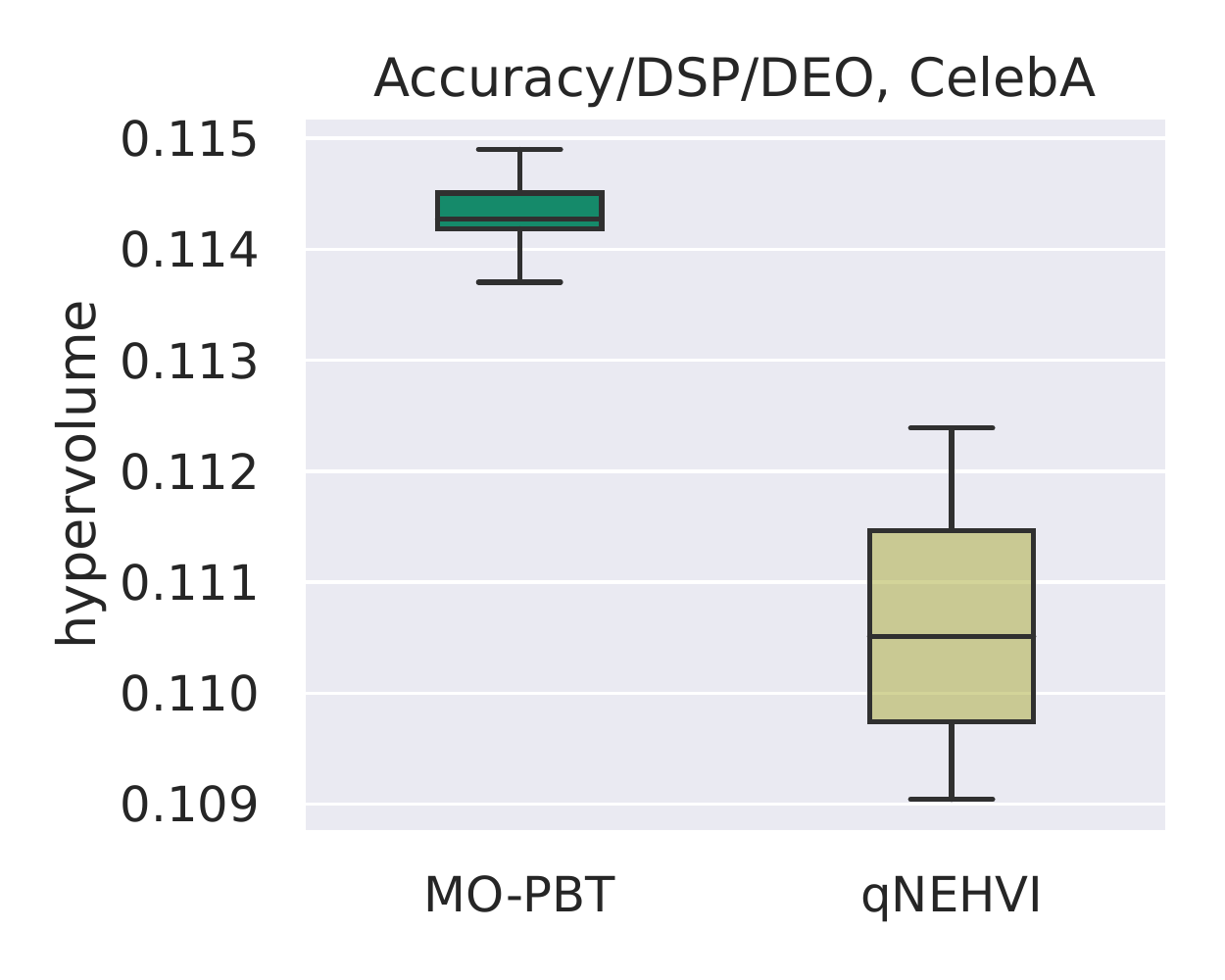} \newline
    \hspace{0.1cm}
    \includegraphics[width=0.2\textwidth]{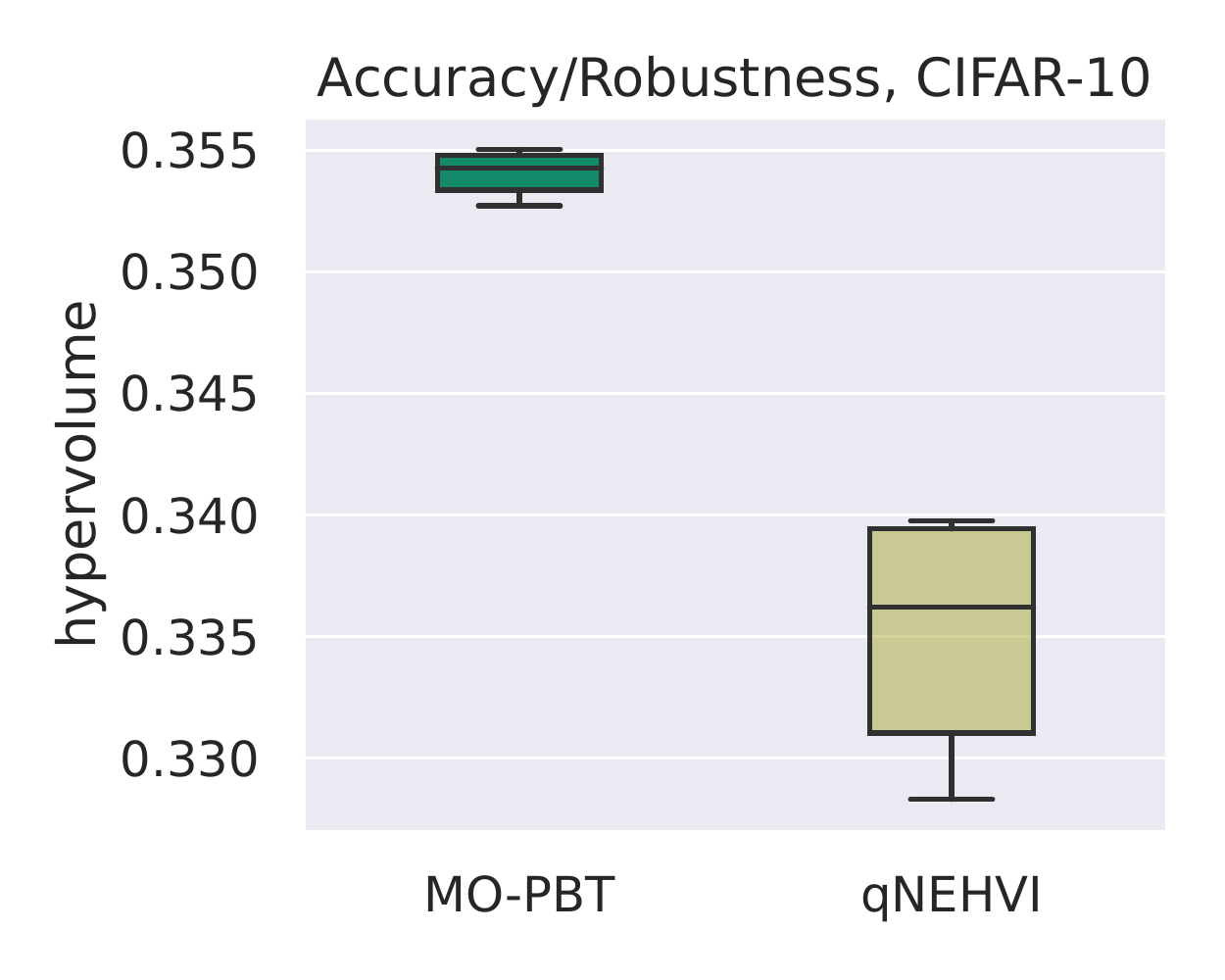}%
    \hspace{0.1cm}
    \includegraphics[width=0.2\textwidth]{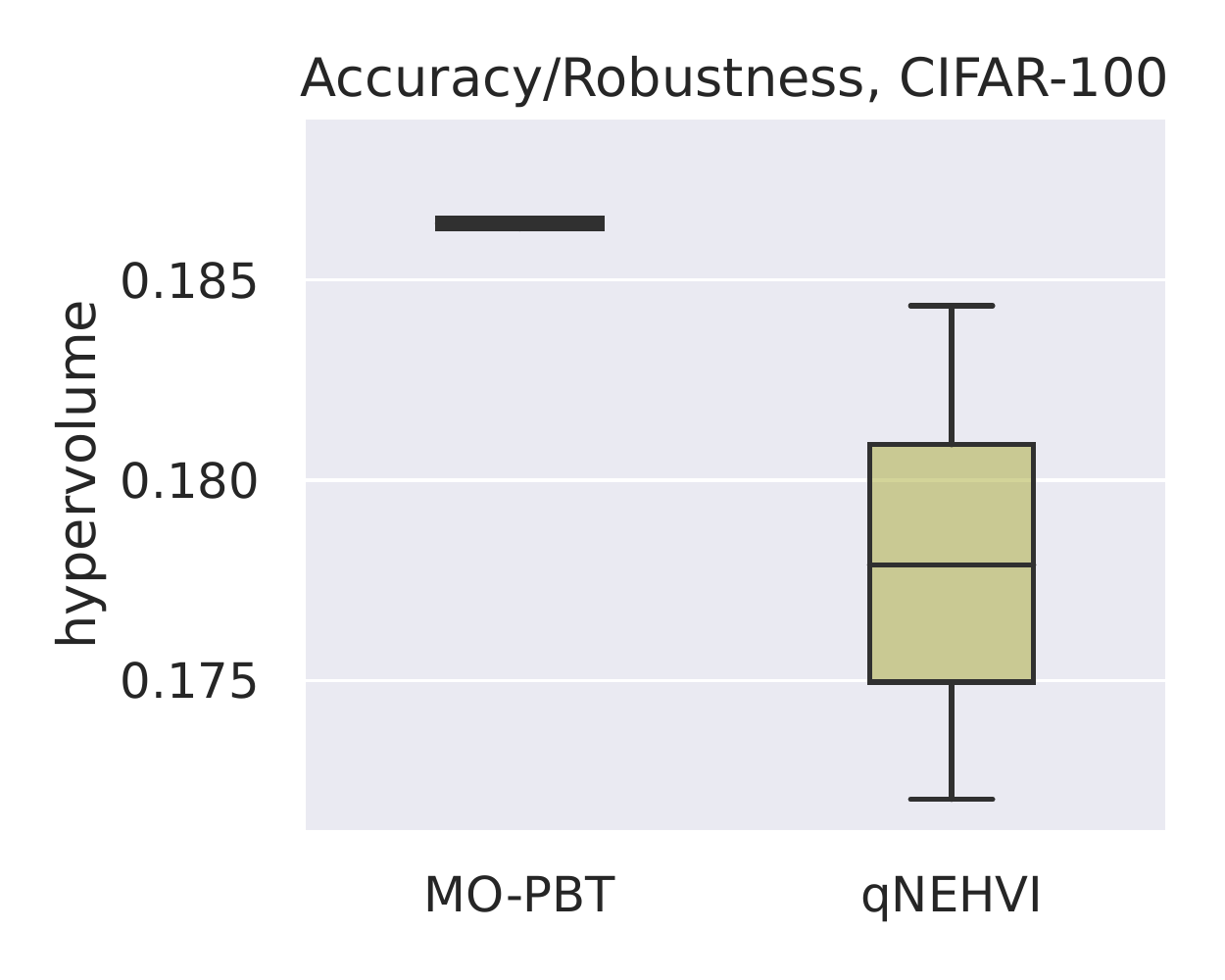}%

    \caption{Comparison of MO-PBT against the state-of-the-art BO algorithm, qNEHVI \cite{daulton2021parallel}. Similar to our main experimental setup described in Section~\ref{subsec:evaluation-setup}, time budgets of qNEHVI are equal to the longest run of MO-PBT in the corresponding task.}
    \label{fig:ablation_qnehvi}
\end{figure}

Here we compare MO-PBT to the state-of-the-art BO algorithm qNEHVI \cite{daulton2021parallel} which, in contrast to traditional BO algorithms, is capable of evaluating solutions in batches. The used batch sizes for qNEHVI are chosen according to our maximal available parallel capacity: 16 for tabular datasets and 12 for the image ones. These results are shown in Figure~\ref{fig:ablation_qnehvi}. They demonstrate that MO-PBT outperforms qNEHVI.

 We further note that the parallel capacity of MO-PBT is limited only by the available hardware: the whole population of $N$ networks can be potentially trained in approximately the same amount of time as one network if $N$ parallel workers are available and no bottlenecks appear in the system. This is not the case for, for instance, qNEHVI: first, multiple sequential training iterations (one batch comprises multiple networks) are required to achieve better than random performance; secondly, its performance is expected to deteriorate when the batch size is scaled up to large values \cite{daulton2021parallel}. 

\clearpage

\section{Search effectiveness of MO-PBT} \label{sec:appendix_search_eff}
\begin{figure}[ht!]
    \centering
    \includegraphics[width=0.3\textwidth]{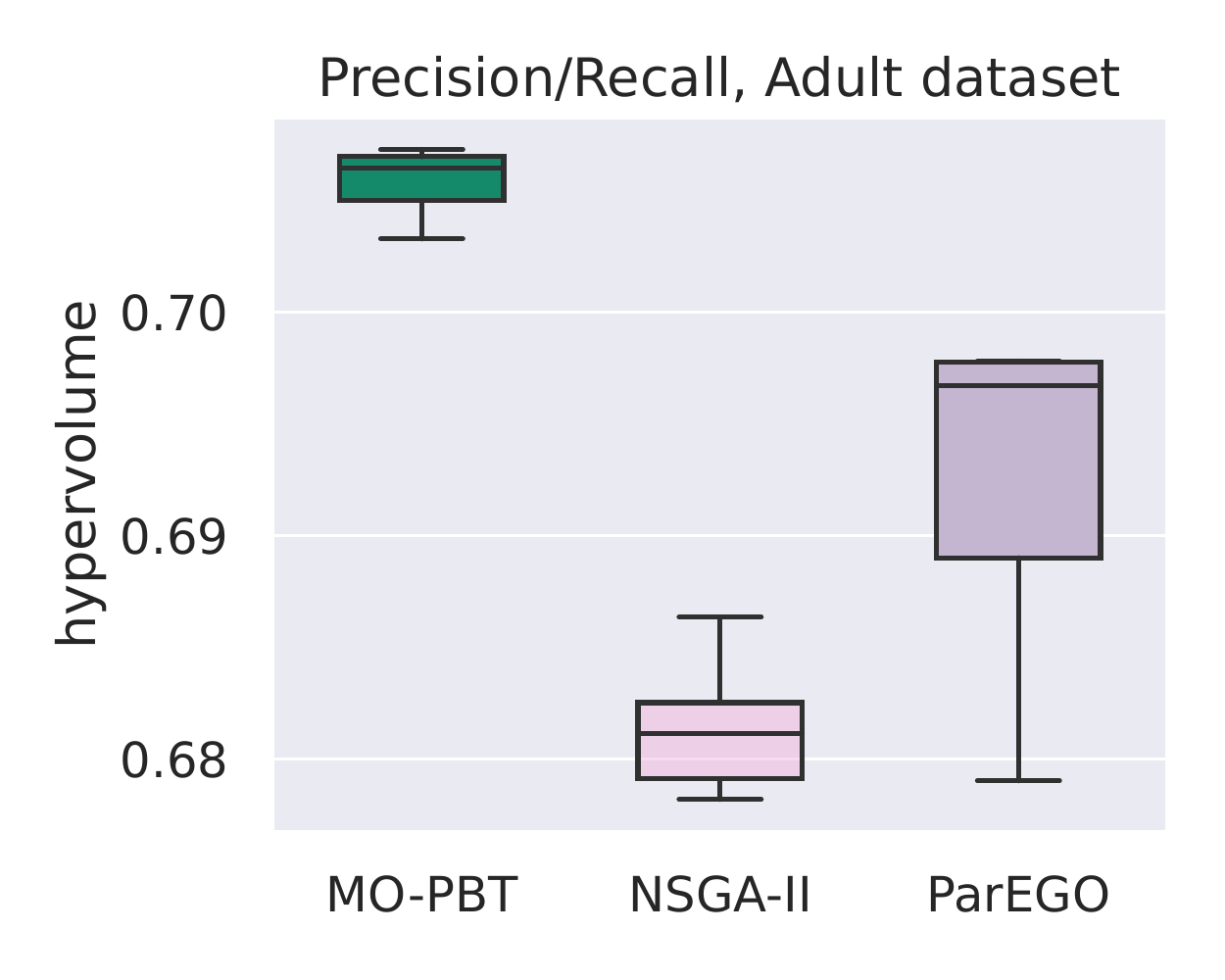}
    \hspace{1cm}
    \includegraphics[width=0.3\textwidth]{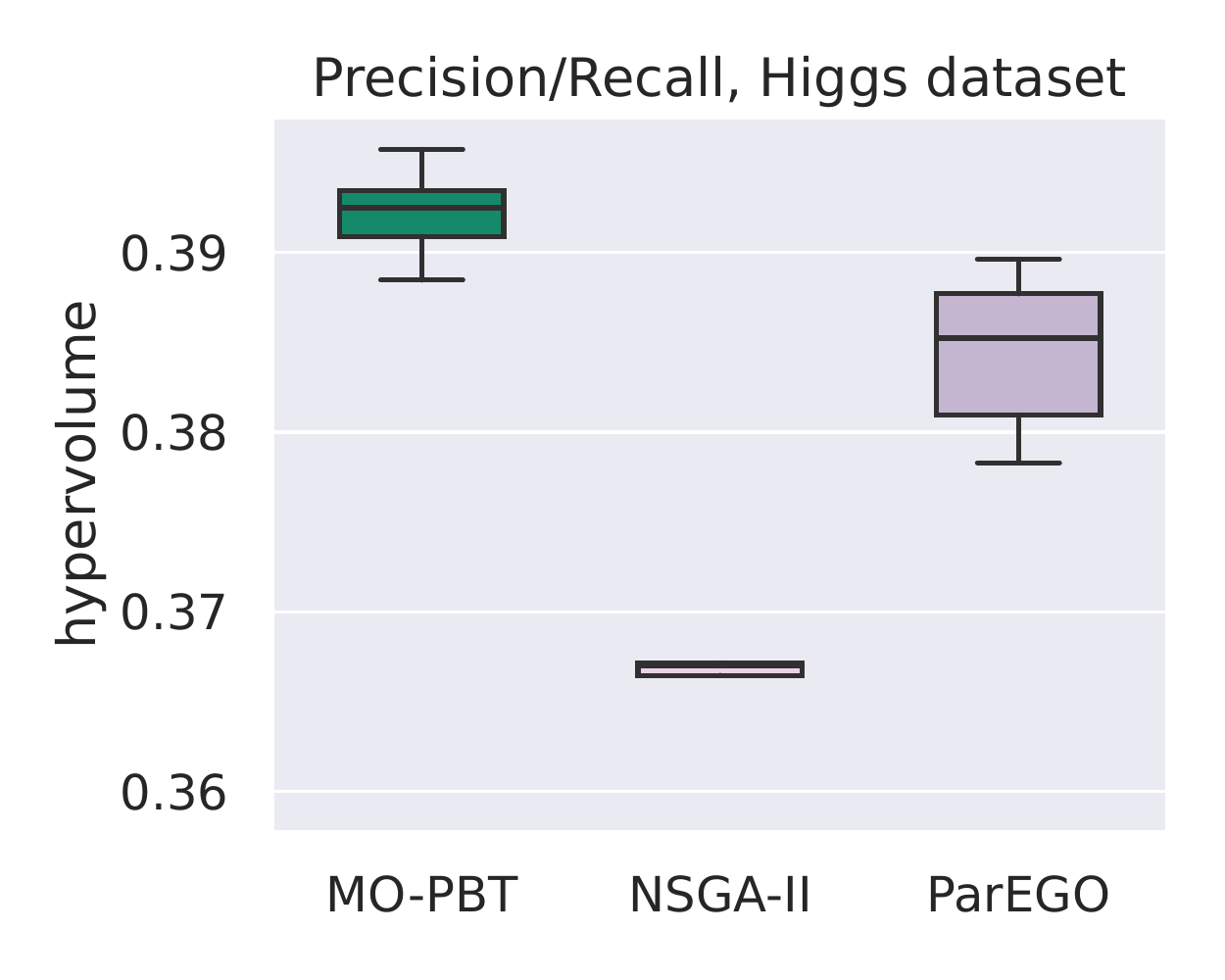}
    \caption{Comparison of MO-PBT against common MO optimization baselines NSGA-II and ParEGO. Here, all algorithms are allowed to fully train 32 networks (the population size in MO-PBT) and the consumed wall-clock time is not taken into account in the evaluation.}
    \label{fig:ablation_sequential}
\end{figure}

Our main experiments demonstrated the efficiency of MO-PBT for practical applications to MO-HPO tasks. Its highly parallel nature plays an important role in its efficiency. Here, we additionally test the search effectiveness of MO-PBT regardless of its parallelization capabilities. For this purpose, we test it against well-known MO baselines: NSGA-II \cite{deb2002fast} and ParEGO \cite{knowles2006parego}. We allow each algorithm to fully train 32 networks (not taking the required wall-clock time into account) and evaluate the performance of the algorithms based on the hypervolume value of obtained non-dominated fronts of solutions. The results are shown in Figure~\ref{fig:ablation_sequential}. We can conclude that MO-PBT outperforms the considered alternatives.

\section{Visualization of the used performance metrics} \label{sec:appendix_metrics}
\begin{figure}[hb!]
\begin{minipage}[t]{.47\textwidth}
    \centering
    \includegraphics[height=5cm]{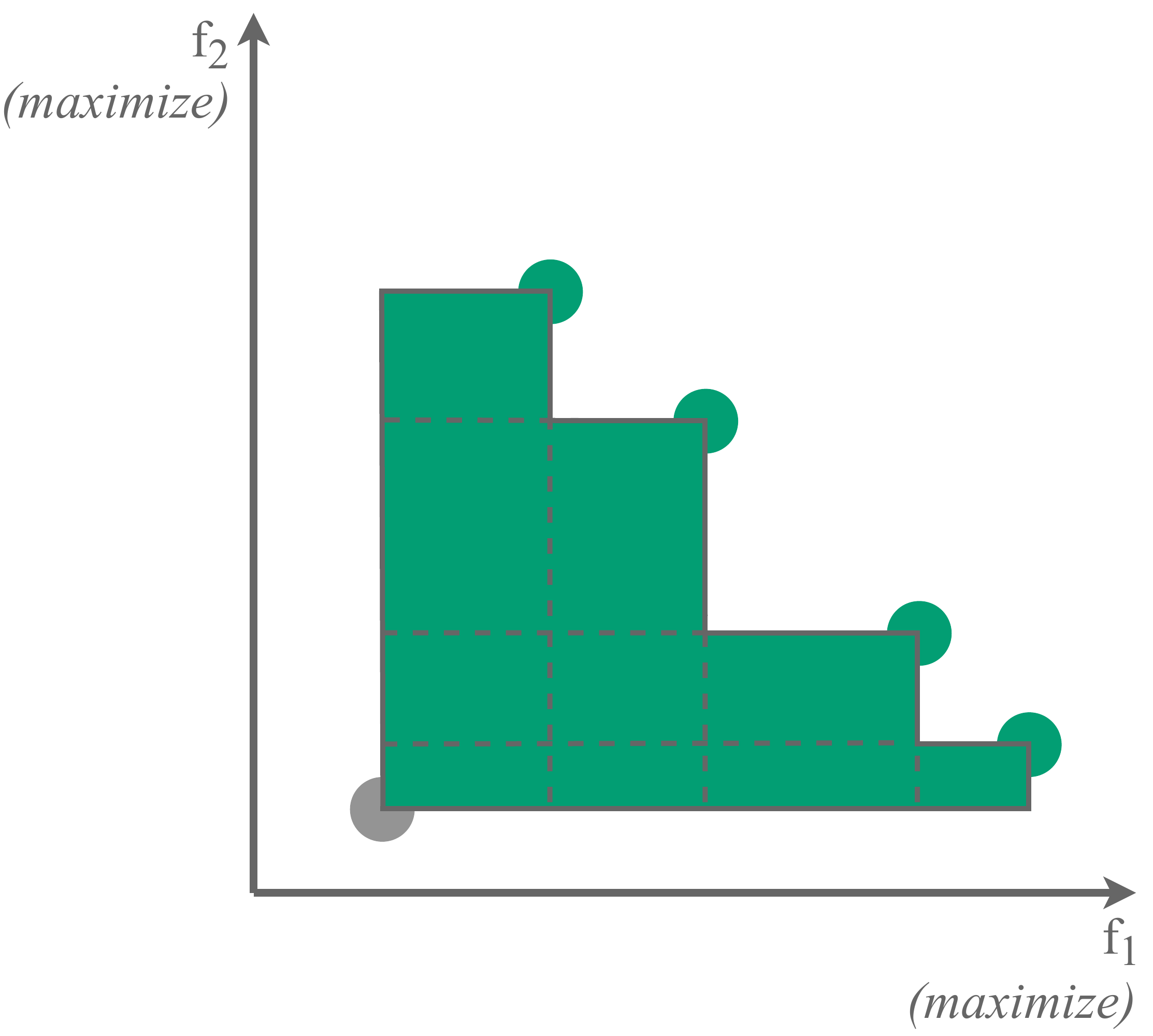}
    \caption{Calculation procedure of the hypervolume metric. Here the green points form the non-dominated front, the gray one is the reference point. The total hypervolume is calculated as the area of the union of rectangles, where each rectangle is formed by a point on the non-dominated front and the reference point.}    
\end{minipage}%
\hspace{1cm}
\begin{minipage}[t]{.47\textwidth}
    \centering
    \includegraphics[height=5cm]{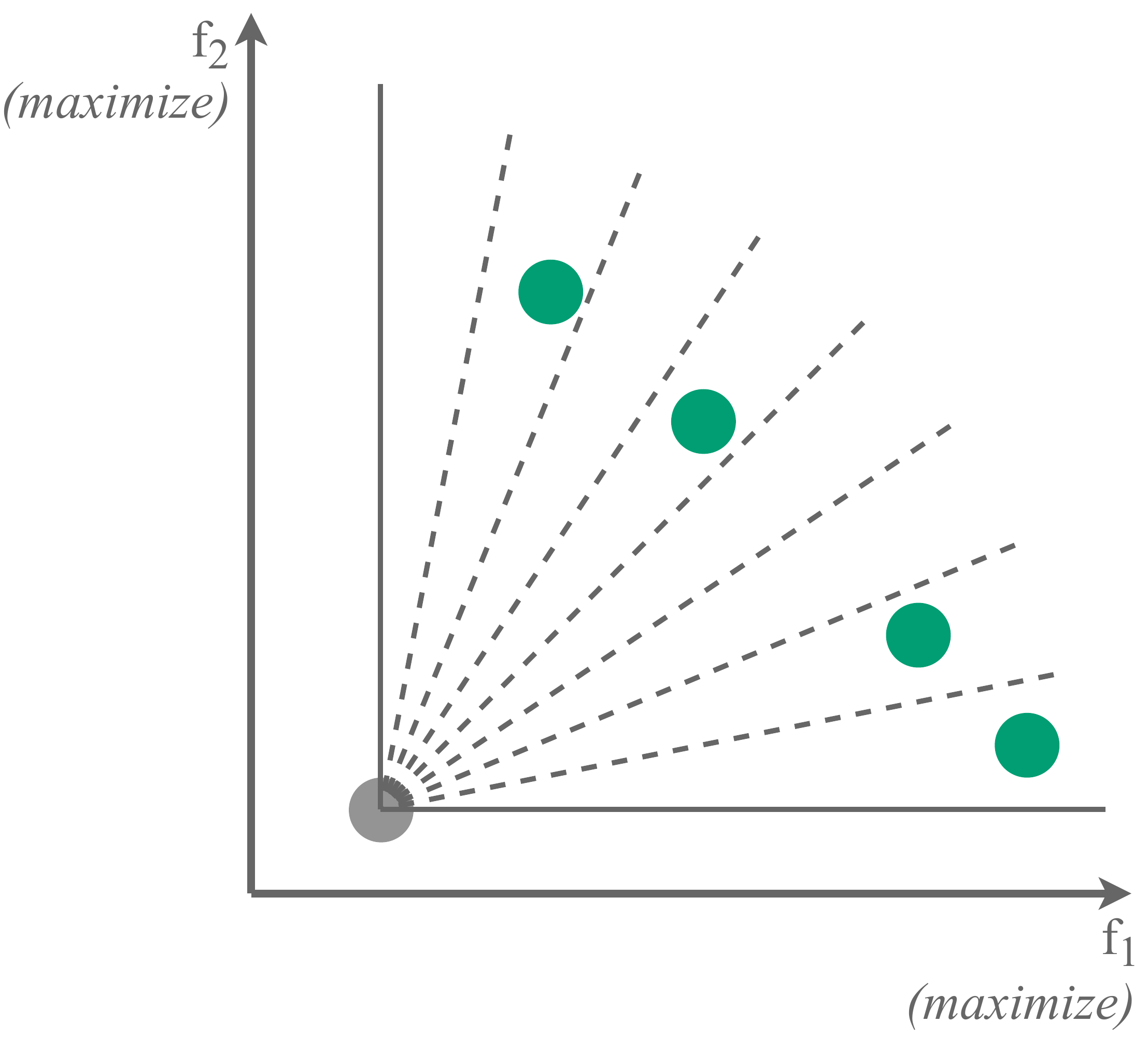}
    \caption{Calculation procedure of the coverage metric \cite{scriven2009dynamic}. Here the green points form the non-dominated front, the reference point is gray. The quadrant in objective space is divided in equal sectors by $M$ dashed lines. The metric value is calculated as the number of sectors with at least one point (here: 4) divided by the total number of sectors $M+1$ (here: 8). In our experiments, we use $M=360$ (larger $M$ means a more fine-grained metric calculation \cite{scriven2009dynamic}).
    }
\label{fig:metrics_calc}
\end{minipage}

\end{figure}
\clearpage

\normalsize

\section{Pseudocode} \label{sec:appendix_codes}
\scriptsize
\begin{algorithm}[h!]
   \caption{Procedure to sort solutions in MO-PBT (\texttt{sortPopulation})}
   \label{alg:epsnetwork}
\begin{algorithmic}
   \STATE \MYCOMMENT{For the sake of implementation simplicity, we sort all fronts regardless of their sizes (the overhead of this operation is negligible) even though not all of them might be needed to select the top and the bottom solutions of the population.}
   \STATE {\bfseries Input:} Population $\mathcal{P}$
   \STATE {\bfseries Output:} Sorted population $\mathcal{P^*}$ 
   \STATE $(F^1,\dots,F^R) \gets$ non-dominated sort of $P$ \COMMENT{$F^i$ is the $\mathit{i}^{th}$ non-dominated front}
   \STATE $\mathcal{P^*} \gets \{\argmax_{v \in F^1} f_1(v)\}$   \COMMENT{add the solution with the largest $f_1$ first}
    \STATE $F^1 \gets F^1 \setminus \{v\}$
   \FOR{$i=1,\dots,R$} 
       \WHILE{$F^i \neq \emptyset$ }
       \STATE $nextToAdd \gets \argmax_{v \in F^i} \argmin_{v' \in \mathcal{P^*}} D(f(v), f(v'))$ \COMMENT{$D$ is Euclidean distance in the objective space}
       \STATE $F^i \gets F^i \setminus \{nextToAdd\}$
       \STATE $\mathcal{P^*} \gets \mathcal{P^*} \cup \{nextToAdd\}$
       
       \ENDWHILE
    \ENDFOR
    \STATE \textbf{return~} $\mathcal{P^*}$
\end{algorithmic}
\end{algorithm}
\vspace{0.5cm}

\begin{algorithm}[h!]
   \caption{Exploit in MO-PBT (\texttt{exploit})}
   \label{alg:mopbt_exploit}
\begin{algorithmic}[1]
    \STATE
    
     {\bfseries Input:} population $\mathcal{P}$, truncation selection parameter $\tau$
    
     {\bfseries Output:} population $\mathcal{P}$ with changed weights and hyperparameters (in-place)
    
   \STATE $\mathcal{P} \gets$ \texttt{sortPopulation}($\mathcal{P})$ 
   \FOR{$p \in$ $\tau|\mathcal{P}|$ bottom solutions of $\mathcal{P}$}
        \STATE select a solution $d$ from $\tau|\mathcal{P}|$ top solutions
        \STATE $p_{\theta} \gets d_{\theta}$ \COMMENT{copy weights}
        \STATE $p_{h} \gets d_{h}$ \COMMENT{copy hyperparameters}
        \STATE $p_{h} \gets$ \texttt{explore}$(p_{h})$ \COMMENT{perturb hyperparameters}
   \ENDFOR
\STATE \textbf{return~} $\mathcal{P}$
\end{algorithmic}
\end{algorithm}
\vspace{0.5cm}

\begin{algorithm}[h!]
   \caption{Explore in MO-PBT (\texttt{explore})}
   \label{alg:mopbt_explore}
\begin{algorithmic}[1]
    \STATE
    
     {\bfseries Input:} hyperparameter value $h$, hyperparameter values domain $\mathcal{H}={v_1, \dots v_M}$, resample probability $p$
        
     {\bfseries Output:} perturbed hyperparameter value $h'$
   \STATE $x \gets$ sample from uniform distribution $U(0,1)$ 
   \IF{$x < p$} 
    \STATE $h' \gets $ uniformly sampled value from $\mathcal{H}$ \COMMENT{with small probability it is resampled}
    \ELSE 
        \STATE $shift \gets$ uniformly sampled value from $[0,1,2,3]$ 
        \STATE $shift \gets -$ \textit{shift} with probability 0.5
        \STATE $h' \gets h+$\textit{shift} \COMMENT{local perturbation}
   \ENDIF
   \STATE \textbf{return} $h'$
   
\end{algorithmic}
\end{algorithm}

\clearpage
\section{Search spaces} \label{sec:appendix_searchspaces}
\subsection{Precision/Recall}
\begin{table}[h!]
\footnotesize
    \centering
    \begin{tabular}{@{}lllll@{}}
    \textbf{Hyperparameter} & \textbf{Range of values} & \textbf{Scale} & \textbf{Number of values}\\ \midrule
        %\midrule
        Attention dropout & [0, 0.8] & linear & 10 \\
        FFN dropout & [0, 0.8] & linear & 10 \\
        Residual dropout & [0, 0.8] & linear & 10 \\
        Weight decay & [0, 0.1] & log & 10 \\
        Class weight in CE loss & [0.1, 0.9] & linear & 10 \\
        \bottomrule
    \end{tabular}
    \caption{Search space for the Precision/Recall task.}
    
\end{table}

\subsection{Accuracy/Fairness}
\begin{table}[h!]
\footnotesize
    \centering
    \begin{tabular}{@{}lllll@{}}
        \textbf{Hyperparameter} & \textbf{Range of values} & \textbf{Scale} & \textbf{Number of values}\\ \midrule
        %\midrule
        Attention dropout & [0, 0.8] & linear & 10 \\
        FFN dropout & [0, 0.8] & linear & 10 \\
        Residual dropout & [0, 0.8] & linear & 10 \\
        Weight decay & [0, 0.1] & log & 10 \\
        Class weight in fairness loss & [0, 10] & log & 10 \\
        \bottomrule
    \end{tabular}
    \caption{Search space for the Accuracy/Fairness tasks on the Adult dataset.}
    
\end{table}

\begin{table}[h!]
\footnotesize
    \centering
    \begin{tabular}{@{}lllll@{}}
    \textbf{Hyperparameter} & \textbf{Range of values} & \textbf{Scale} & \textbf{Number of values}\\ \midrule
        %\midrule
        RandAugment N & [0, 4] & linear & 5 \\
        RandAugment M & [0, 9] & linear & 10 \\
        CutOut probability & [0, 1] & linear & 10 \\
        CutOut magnitude & [0, 9] & linear & 10 \\
        Class weight in fairness loss & [0, 10] & log & 10 \\
        \bottomrule
    \end{tabular}
    \caption{Search space for the Accuracy/Fairness tasks on the CelebA dataset.}
    
\end{table}

\subsection{Accuracy/Adversarial robustness}
\begin{table}[h!]
\footnotesize
    \centering
    \begin{tabular}{@{}lllll@{}}
    \textbf{Hyperparameter} & \textbf{Range of values} & \textbf{Scale} & \textbf{Number of values}\\ \midrule
        %\midrule
        RandAugment N & [0, 4] & linear & 5 \\
        RandAugment M & [0, 9] & linear & 10 \\
        CutOut probability & [0, 1] & linear & 10 \\
        CutOut magnitude & [0, 9] & linear & 10 \\
        Coefficient in the TRADES loss & [0, 10] & log & 10 \\
        \bottomrule
    \end{tabular}
    \caption{Search space of the Accuracy/Robustness task.}
    
\end{table}

\normalsize

\clearpage

\section{Tabulated results} \label{sec:appendix_tables}
\begin{table}[h!]
\tiny
\setlength{\tabcolsep}{2.75pt}
\begin{tabular}{@{}ll|l|llc|ll|l|lll@{}}
\toprule
Problem & Dataset & random search & obj. 1  & obj. 2  & obj. 3  & rand. scalar.  & max. scalar. & MO-PBT & MO-ASHA & BO-MO-ASHA \\ \midrule
\multirow{3}{*}{Precision/Recall}  & Adult & $68.69 \pm 0.50$ & $69.61 \pm 0.27$ & $63.80 \pm 0.90$ & - & $68.13 \pm 0.97$ & $68.79 \pm 0.57$ & $\bm{70.59 \pm 0.13}$ & $67.89 \pm 0.52$ & $66.10 \pm 1.56$  \\ 
 & Higgs & $36.88 \pm 0.26$ & $38.14 \pm 0.36$ & $33.40 \pm 0.61$ & - & $36.50 \pm 0.43$ & $37.40 \pm 0.56$ & $\bm{39.22 \pm 0.22}$ & $37.46 \pm 0.32$ & $36.34 \pm 0.75$  \\ 
 & Click & $35.22 \pm 0.45$ & $34.65 \pm 0.50$ & $34.33 \pm 0.86$ & - & $35.01 \pm 0.39$ & $34.89 \pm 0.52$ & $\bm{37.39 \pm 0.24}$ & $35.69 \pm 0.43$ & $35.66 \pm 0.58$  \\ \midrule
\multirow{2}{*}{\shortstack{Fairness:\\ Acc./DSP}}  & CelebA & $16.51 \pm 0.09$ & $16.53 \pm 0.03$ & $15.17 \pm 0.12$ & - & $16.59 \pm 0.07$ & $16.51 \pm 0.02$ & $\bm{16.76 \pm 0.08}$ & $16.39 \pm 0.06$ & $16.40 \pm 0.05$  \\ 
 & Adult & $3.64 \pm 0.04$ & $3.69 \pm 0.02$ & $2.80 \pm 0.11$ & - & $3.55 \pm 0.11$ & $3.55 \pm 0.17$ & $\bm{3.81 \pm 0.02}$ & $3.74 \pm 0.01$ & $3.73 \pm 0.03$  \\ \midrule
\multirow{2}{*}{\shortstack{Fairness:\\ Acc./DSP/EOdd}}  & CelebA & $11.09 \pm 0.08$ & $11.11 \pm 0.05$ & $10.10 \pm 0.11$ & $10.13 \pm 0.13$ & $11.00 \pm 0.07$ & $11.20 \pm 0.10$ & $\bm{11.43 \pm 0.04}$ & $11.12 \pm 0.07$ & $11.15 \pm 0.10$  \\ 
 & Adult & $1.62 \pm 0.02$ & $1.64 \pm 0.02$ & $1.30 \pm 0.11$ & $1.32 \pm 0.09$ & $1.63 \pm 0.05$ & $1.47 \pm 0.02$ & $\bm{1.76 \pm 0.02}$ & $1.70 \pm 0.02$ & $1.68 \pm 0.02$  \\ \midrule
\multirow{2}{*}{Acc./Robustness}  & CIFAR-10 & $33.84 \pm 0.28$ & $24.65 \pm 0.81$ & $33.94 \pm 0.26$ & - & $34.74 \pm 0.48$ & $34.51 \pm 0.29$ & $\bm{35.40 \pm 0.09}$ & $33.62 \pm 0.59$ & $33.82 \pm 0.32$  \\ 
 & CIFAR-100 & $17.90 \pm 0.27$ & $12.06 \pm 0.15$ & $17.44 \pm 0.13$ & - & $17.77 \pm 0.25$ & $18.13 \pm 0.24$ & $\bm{18.65 \pm 0.11}$ & $16.98 \pm 0.53$ & $16.92 \pm 1.07$  \\
 \bottomrule
\end{tabular}
\caption{Obtained hypervolume (\emph{larger values are better}) data for all algorithms and all tasks. Average and standard deviation values of the best obtained hypervolume over multiple runs are provided. Obj.1, obj 2., and obj. 3 denote single-objective PBT applied to optimizing the corresponding objective of the task. Acc. denotes accuracy. For better readability, all values are multiplied by 100.}
\label{tab:hv_results}
\end{table}
\vspace{1cm}

\begin{table}[h!]
\tiny
\setlength{\tabcolsep}{2.75pt}
\begin{tabular}{@{}ll|l|llc|ll|l|lll@{}}
\toprule
Problem & Dataset & random search & obj. 1  & obj. 2  & obj. 3  & rand. scalar.  & max. scalar. & MO-PBT & MO-ASHA & BO-MO-ASHA \\ \midrule
\multirow{3}{*}{Precision/Recall}  & Adult & $68.57 \pm 0.25$ & $69.38 \pm 0.17$ & $64.94 \pm 0.70$ & - & $68.27 \pm 0.64$ & $68.63 \pm 0.52$ & $\bm{69.95 \pm 0.18}$ & $67.47 \pm 0.65$ & $65.92 \pm 1.41$  \\ 
 & Higgs & $36.01 \pm 0.53$ & $37.42 \pm 0.31$ & $32.74 \pm 0.80$ & - & $35.67 \pm 0.54$ & $36.49 \pm 0.51$ & $\bm{38.31 \pm 0.19}$ & $36.56 \pm 0.38$ & $35.21 \pm 0.95$  \\ 
 & Click & $34.06 \pm 0.53$ & $33.60 \pm 0.51$ & $33.60 \pm 0.74$ & - & $34.14 \pm 0.40$ & $34.16 \pm 0.56$ & $\bm{36.38 \pm 0.25}$ & $34.81 \pm 0.87$ & $34.95 \pm 0.63$  \\ \midrule
\multirow{2}{*}{\shortstack{Fairness:\\ Acc./DSP}}  & CelebA & $16.47 \pm 0.09$ & $16.50 \pm 0.05$ & $15.28 \pm 0.07$ & - & $16.57 \pm 0.09$ & $16.49 \pm 0.05$ & $\bm{16.79 \pm 0.04}$ & $16.34 \pm 0.10$ & $16.33 \pm 0.05$  \\ 
 & Adult & $3.69 \pm 0.02$ & $3.68 \pm 0.01$ & $3.04 \pm 0.12$ & - & $3.65 \pm 0.08$ & $3.65 \pm 0.10$ & $\bm{3.79 \pm 0.02}$ & $3.75 \pm 0.02$ & $3.73 \pm 0.02$  \\ \midrule
\multirow{2}{*}{\shortstack{Fairness:\\ Acc./DSP/EOdd}}  & CelebA & $10.97 \pm 0.09$ & $10.99 \pm 0.03$ & $10.09 \pm 0.07$ & $10.14 \pm 0.10$ & $10.94 \pm 0.05$ & $11.08 \pm 0.11$ & $\bm{11.36 \pm 0.04}$ & $11.04 \pm 0.07$ & $11.02 \pm 0.09$  \\ 
 & Adult & $1.69 \pm 0.01$ & $1.68 \pm 0.02$ & $1.42 \pm 0.11$ & $1.44 \pm 0.10$ & $1.72 \pm 0.04$ & $1.60 \pm 0.03$ & $\bm{1.80 \pm 0.01}$ & $1.75 \pm 0.01$ & $1.74 \pm 0.01$  \\ \midrule
\multirow{2}{*}{Acc./Robustness}  & CIFAR-10 & $33.40 \pm 0.40$ & $24.54 \pm 0.91$ & $33.45 \pm 0.24$ & - & $34.40 \pm 0.45$ & $34.13 \pm 0.27$ & $\bm{34.99 \pm 0.13}$ & $32.97 \pm 0.69$ & $33.30 \pm 0.31$  \\ 
 & CIFAR-100 & $18.30 \pm 0.29$ & $11.98 \pm 0.20$ & $17.69 \pm 0.15$ & - & $18.13 \pm 0.27$ & $18.50 \pm 0.26$ & $\bm{19.03 \pm 0.16}$ & $17.42 \pm 0.61$ & $17.25 \pm 1.28$  \\ 
 
 \bottomrule
\end{tabular}
\caption{Obtained hypervolume (\emph{larger values are better}) data for all algorithms and all tasks on \emph{test} data subsets. Average and standard deviation values of the best obtained hypervolume over multiple runs are provided. Obj.1, obj. 2, and obj. 3 denote single-objective PBT applied to optimizing the corresponding objective of the task. Acc. denotes accuracy. For better readability, all values are multiplied by 100.}
\label{tab:hv_results_test}
\end{table}
\vspace{1cm}

\begin{table}[h!]
\tiny
\setlength{\tabcolsep}{2.8pt}
\begin{tabular}{@{}ll|l|ll|ll|l|lll@{}}
\toprule
Problem & Dataset & random search & obj. 1  & obj. 2  & rand. scalar.  & max. scalar. & MO-PBT & MO-ASHA & BO-MO-ASHA \\ \midrule
\multirow{3}{*}{Precision/Recall}  & Adult & $29.36 \pm 1.63$ & $21.39 \pm 1.01$ & $23.68 \pm 2.02$ & $29.34 \pm 3.36$ & $26.12 \pm 1.41$ & $\bm{40.39 \pm 2.62}$ & $28.14 \pm 1.71$ & $25.65 \pm 2.40$  \\ 
 & Higgs & $27.87 \pm 2.50$ & $21.99 \pm 1.83$ & $14.38 \pm 1.71$ & $22.33 \pm 2.15$ & $17.76 \pm 1.21$ & $\bm{35.71 \pm 2.07}$ & $20.08 \pm 1.89$ & $19.22 \pm 2.41$  \\ 
 & Click & $30.11 \pm 2.79$ & $17.17 \pm 1.81$ & $20.39 \pm 2.38$ & $27.70 \pm 1.65$ & $18.39 \pm 1.92$ & $\bm{34.96 \pm 2.90}$ & $25.71 \pm 1.75$ & $26.29 \pm 2.19$  \\ \midrule
\multirow{2}{*}{\shortstack{Fairness:\\ Acc./DSP}}  & CelebA & \bm{$8.14 \pm 0.95$} & $7.20 \pm 0.63$ & $4.25 \pm 0.26$ & $6.26 \pm 0.89$ & $5.37 \pm 0.62$ & $7.20 \pm 0.70$ & $7.42 \pm 0.66$ & $7.87 \pm 1.22$  \\ 
 & Adult & $5.90 \pm 1.44$ & $6.93 \pm 1.26$ & $2.16 \pm 0.41$ & $4.60 \pm 1.44$ & $3.02 \pm 0.84$ & $6.76 \pm 1.26$ & \bm{$7.26 \pm 1.09$} & $5.57 \pm 1.71$  \\ \midrule
\multirow{2}{*}{Acc./Robustness}  & CIFAR-10 & $4.38 \pm 0.73$ & $4.80 \pm 0.13$ & $3.05 \pm 0.80$ & $4.60 \pm 0.57$ & $5.60 \pm 0.44$ & $\bm{9.58 \pm 0.51}$ & $2.99 \pm 0.54$ & $3.32 \pm 0.63$  \\ 
 & CIFAR-100 & $5.43 \pm 1.22$ & $5.45 \pm 0.47$ & $3.71 \pm 0.48$ & $4.99 \pm 0.72$ & $6.70 \pm 0.27$ & $\bm{10.75 \pm 1.37}$ & $3.49 \pm 1.06$ & $4.38 \pm 0.98$  \\ 
 \bottomrule
\end{tabular}
\caption{Obtained coverage metric introduced in \cite{scriven2009dynamic} and illustrated in Figure~\ref{fig:metrics_calc} (\emph{larger values are better}) data for all algorithms and all bi-objective tasks. Average and standard deviation values of the best obtained coverage value over multiple runs are provided. Obj.1, obj 2. denote single-objective PBT applied to optimizing the corresponding objective of the task. Acc. denotes accuracy. For better readability, all values are multiplied by 100.}
\label{tab:coverage_results}
\end{table}
\normalsize
\clearpage

\section{Effects of the constraints on the performance}
\begin{figure}[h!]
\captionsetup[subfigure]{labelformat=empty}
    \centering
    \subfigure{\includegraphics[height=3.4cm]{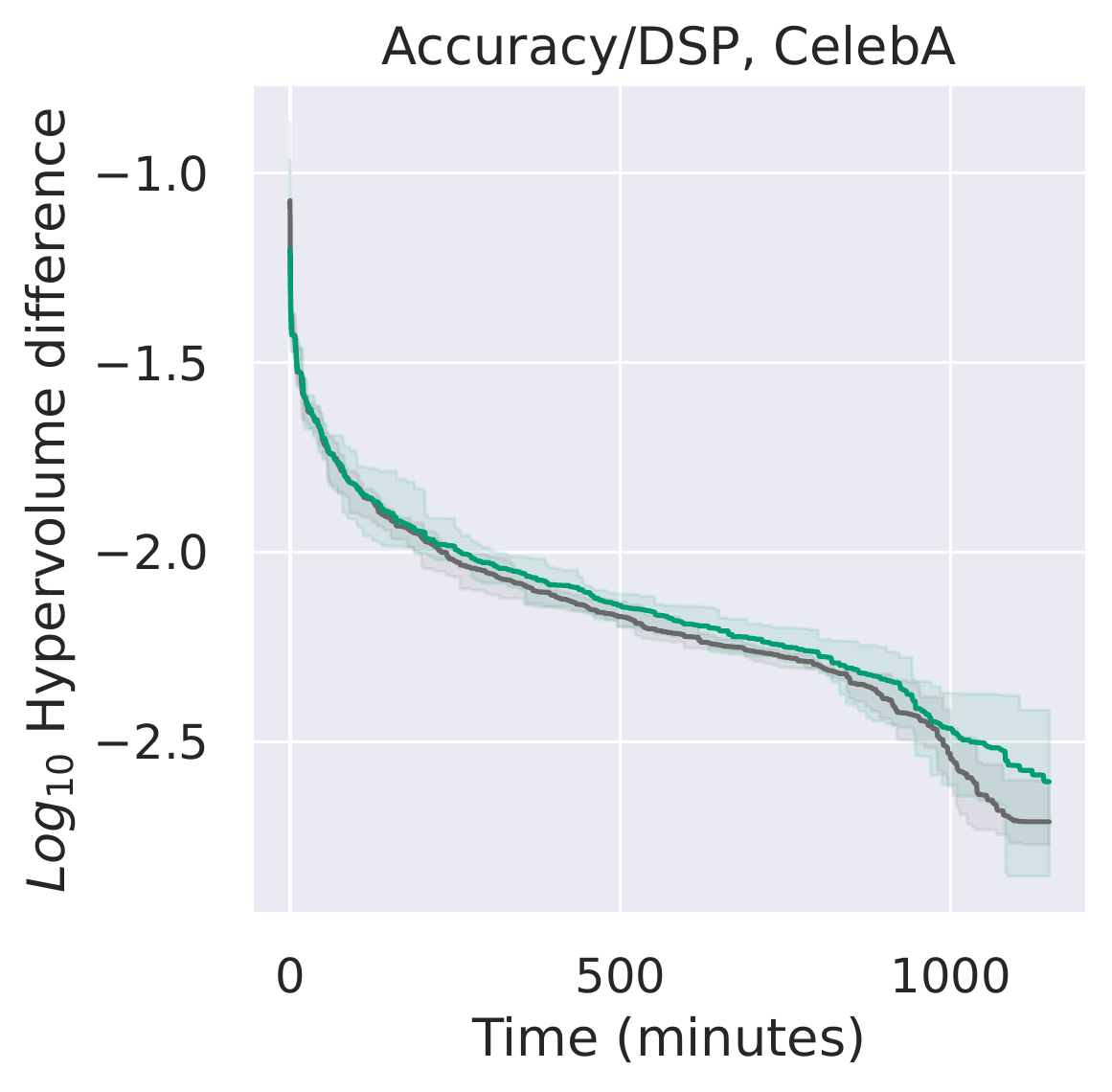}}%
    \hspace{1.5cm}
    \subfigure{\includegraphics[height=3.6cm]{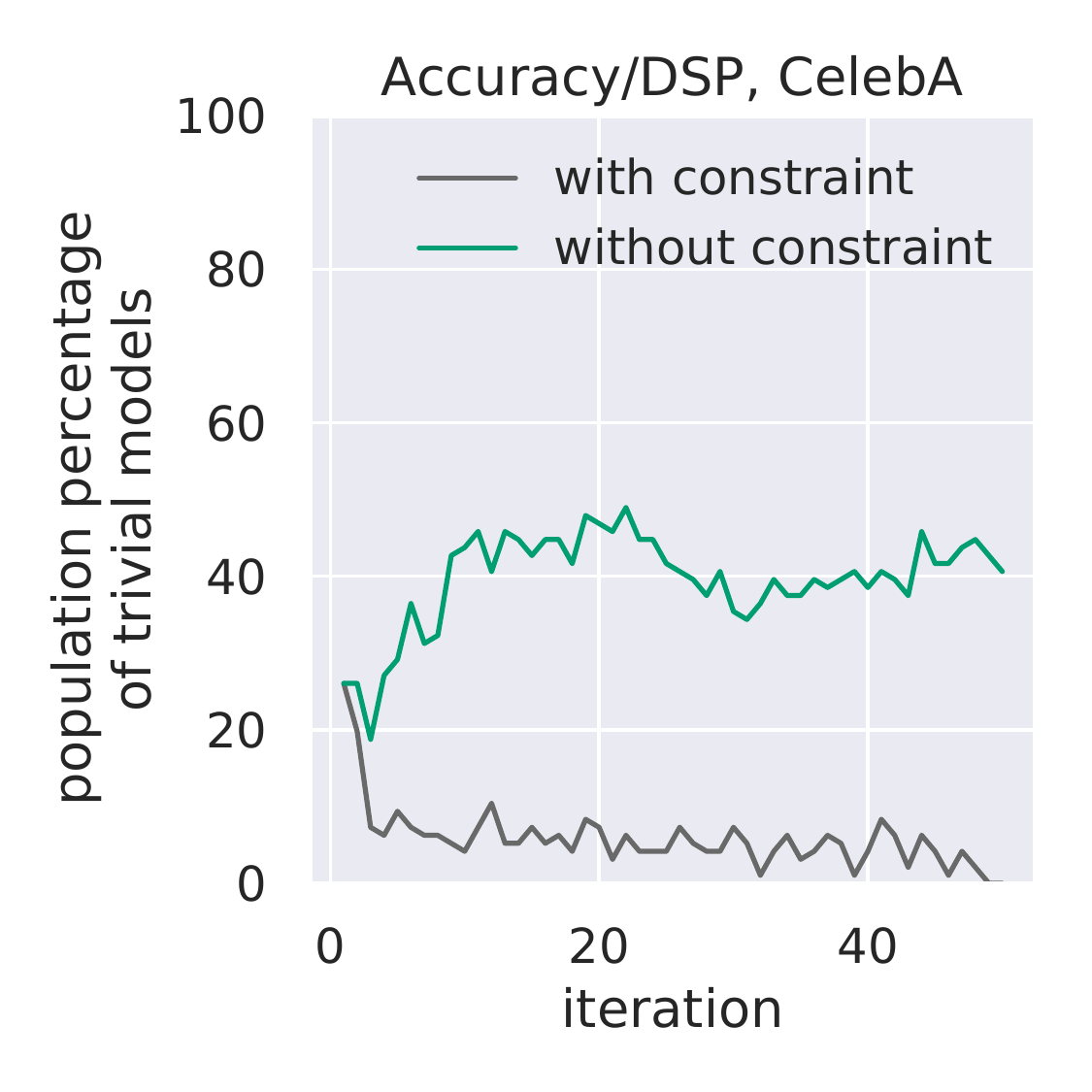}}%

    \vspace{-0.5cm}
    
    \renewcommand{\thesubfigure}{(a)}
    \subfigure[Performance over time]{\includegraphics[height=3.4cm]{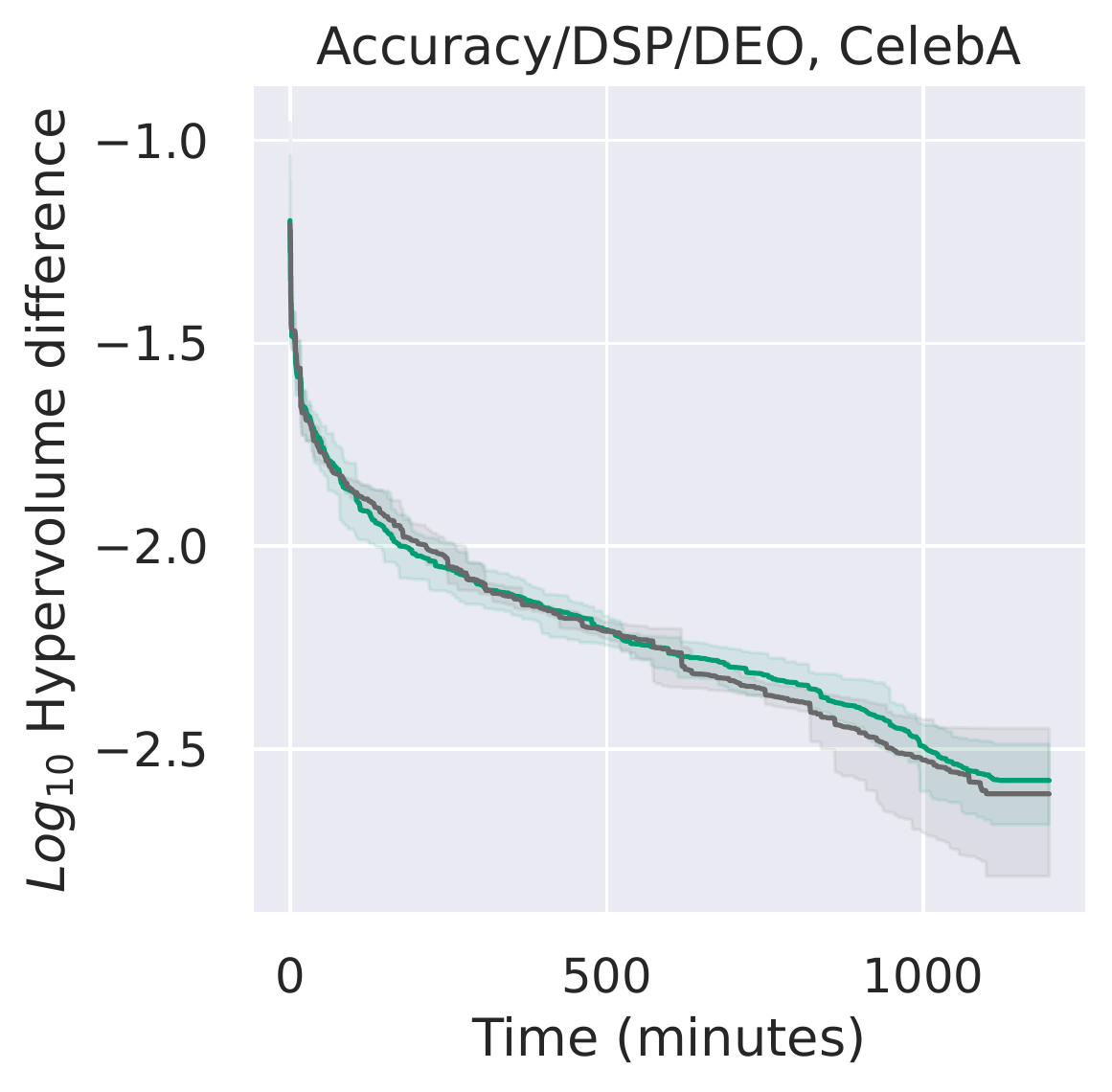}}%
    \renewcommand{\thesubfigure}{(b)}
     \hspace{1.5cm}
    \subfigure[Percentage of trivial classifiers in the population]{\includegraphics[height=3.6cm]{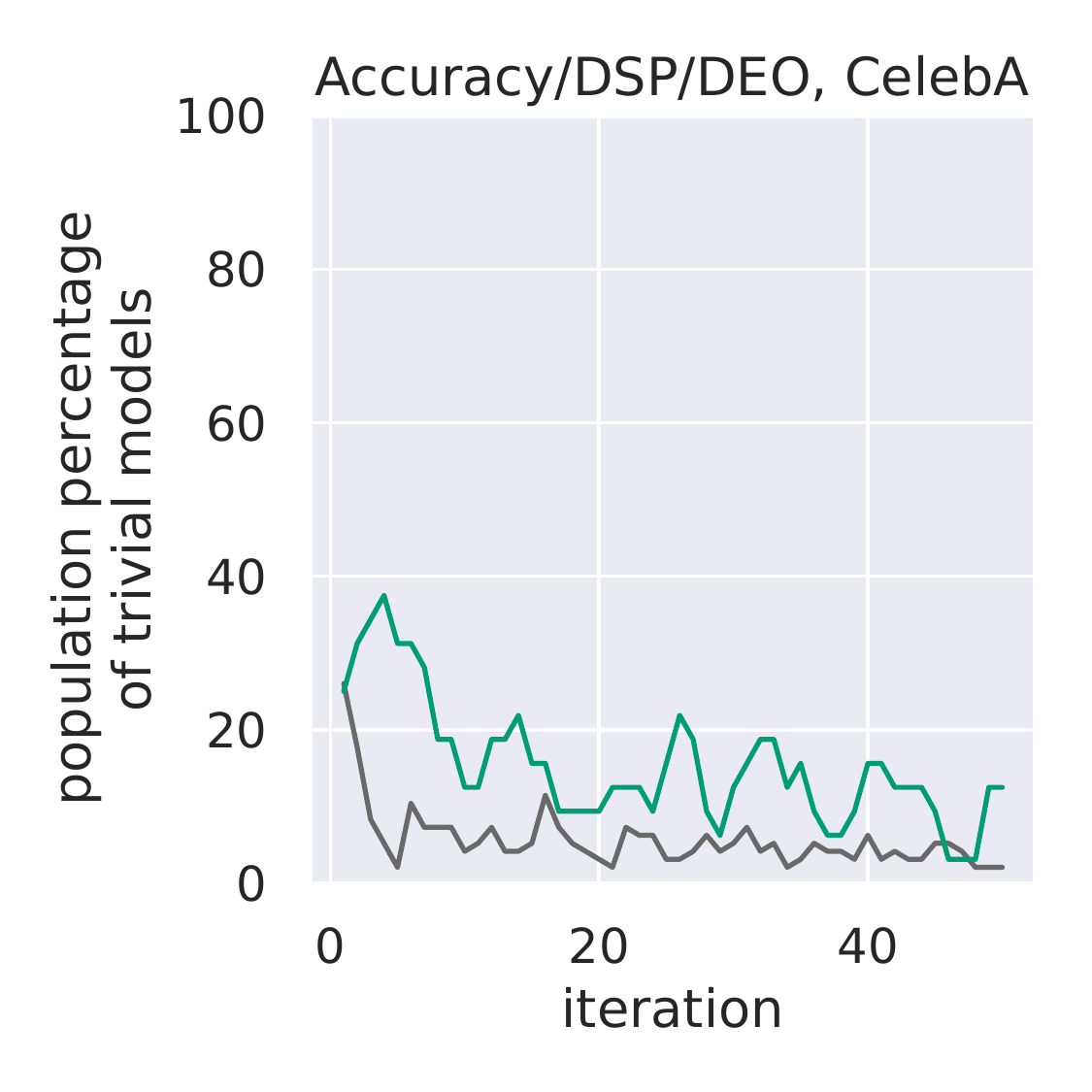}}%

    \caption{Comparison of MO-PBT performance on the CelebA Accuracy/DSP and Accuracy/DSP/DEO tasks with and without constraint that demands the models to have better accuracy than trivial classifiers. }
    \label{fig:CelebA_constrained}
\end{figure}

We noticed that on the CelebA Accuracy/Fairness task, many trivial classifiers (they always predict one of the classes, which leads to a perfect fairness score) appear in the population. However, we are not interested in obtaining trivial classifiers, as such a classifier is the most straightforward way to obtain a perfectly fair model, but it is not particularly useful. Therefore, we can consider a solution to be feasible if its accuracy is better than the accuracy of a trivial classifier. If constraints are imposed, the domination criterion can be extended to \emph{constraint domination} \cite{deb2002fast} (the solutions that violate constraints are considered to be dominated by the solutions that do not).
We hypothesized that with such constraints, the algorithm can become more effective in finding non-trivial solutions. The imposed constraint on the accuracy to be higher than the accuracy of a trivial classifier reduces the number of trivial classifiers in the population, but the hypervolume-measured performance improvement is subtle, as shown in Figure~\ref{fig:CelebA_constrained}. Therefore, we did not use this technique in the main experiments.
\vspace{-0.4cm}

\end{document}